\def\EgoForesightArxiv{1}
\documentclass{egofore_arxiv}

\usepackage{amsmath,amsfonts,bm}

\def\eqref#1{equation~\ref{#1}}

\def\1{\bm{1}}

\DeclareMathAlphabet{\mathsfit}{\encodingdefault}{\sfdefault}{m}{sl}
\SetMathAlphabet{\mathsfit}{bold}{\encodingdefault}{\sfdefault}{bx}{n}

\renewcommand{\eqref}[1]{\textup{(\ref{#1})}}

\usepackage{amsmath}
\usepackage{amssymb}
\usepackage{booktabs}
\usepackage{array}
\usepackage{graphicx}
\usepackage{capt-of}
\usepackage{afterpage}
\usepackage{hyperref}
\usepackage{url}
\usepackage{xcolor}
\usepackage{tikz}
\definecolor{ffInk}{HTML}{514D50}
\definecolor{ffMuted}{HTML}{827C80}
\definecolor{ffArrow}{HTML}{C3C5C2}
\definecolor{ffBlue}{HTML}{718A9B}
\definecolor{ffBluel}{HTML}{D5DFE5}
\definecolor{ffBluebg}{HTML}{EEF2F4}
\definecolor{ffSage}{HTML}{7E9585}
\definecolor{ffSagel}{HTML}{CAD8CD}
\definecolor{ffClay}{HTML}{AB8D7C}
\definecolor{ffClayl}{HTML}{DDCDC0}
\definecolor{ffLav}{HTML}{92849F}
\definecolor{ffLavl}{HTML}{D9D1E0}
\definecolor{ffLavbg}{HTML}{F4F1F7}
\definecolor{ffData}{HTML}{F4F2F3}
\definecolor{ffNoise}{HTML}{F6F0EC}
\definecolor{ffJoint}{HTML}{EFF4F2}
\definecolor{ffPaper}{HTML}{FFFFFF}
\definecolor{ffBlueDark}{HTML}{5C7485}
\definecolor{ffSageDark}{HTML}{607D6B}
\definecolor{ffClayDark}{HTML}{98705B}
\definecolor{ffLavDark}{HTML}{766683}
\definecolor{ffGrid}{HTML}{DAD6DA}
\definecolor{ffFloor}{HTML}{ECE8EC}

\ExplSyntaxOn
\prop_new:N \l_ablation_values_prop
\prop_new:N \l_ablation_colors_prop
\prop_new:N \l_ablation_second_colors_prop
\prop_new:N \l_ablation_maxima_prop
\prop_new:N \l_ablation_seconds_prop
\fp_new:N \l_ablation_maximum_fp
\fp_new:N \l_ablation_second_fp
\bool_new:N \l_ablation_complete_bool
\tl_new:N \l_ablation_data_tl
\tl_new:N \l_ablation_value_tl
\tl_new:N \l_ablation_color_tl
\tl_new:N \l_ablation_second_color_tl
\tl_new:N \l_ablation_maximum_tl
\tl_new:N \l_ablation_second_tl

\cs_new_protected:Npn \ablation_set_column:nnn #1#2#3
  {
    \prop_put:Nnn \l_ablation_values_prop {#1/#2} {#3}
    \fp_set:Nn \l_ablation_maximum_fp {-1}
    \bool_set_true:N \l_ablation_complete_bool
    \clist_map_inline:nn {#3}
      {
        \tl_if_eq:nnTF {##1} {---}
          {\bool_set_false:N \l_ablation_complete_bool}
          {\fp_set:Nn \l_ablation_maximum_fp {max(\l_ablation_maximum_fp, ##1)}}
      }
    \bool_if:NF \l_ablation_complete_bool
      {\fp_set:Nn \l_ablation_maximum_fp {-1}}
    \prop_put:Nnx \l_ablation_maxima_prop {#1/#2} {\fp_use:N \l_ablation_maximum_fp}
    \fp_set:Nn \l_ablation_second_fp {-1}
    \bool_if:NT \l_ablation_complete_bool
      {
        \clist_map_inline:nn {#3}
          {
            \fp_compare:nT {##1 < \l_ablation_maximum_fp}
              {\fp_set:Nn \l_ablation_second_fp {max(\l_ablation_second_fp, ##1)}}
          }
      }
    \prop_put:Nnx \l_ablation_seconds_prop {#1/#2} {\fp_use:N \l_ablation_second_fp}
  }

\NewDocumentCommand{\ablationValues}{O{}mmmmm}
  {
    \prop_put:Nnn \l_ablation_colors_prop {#2} {#3}
    \prop_put:Nnn \l_ablation_second_colors_prop {#2} {#1}
    \ablation_set_column:nnn {#2} {pnp} {#4}
    \ablation_set_column:nnn {#2} {art} {#5}
    \ablation_set_column:nnn {#2} {avg} {#6}
  }

\cs_new_protected:Npn \ablation_cell:nnn #1#2#3
  {
    \group_begin:
    \prop_get:NnN \l_ablation_values_prop {#1/#2} \l_ablation_data_tl
    \tl_set:Nx \l_ablation_value_tl
      {\exp_args:NV \clist_item:nn \l_ablation_data_tl {#3}}
    \prop_get:NnN \l_ablation_colors_prop {#1} \l_ablation_color_tl
    \prop_get:NnN \l_ablation_second_colors_prop {#1} \l_ablation_second_color_tl
    \prop_get:NnN \l_ablation_maxima_prop {#1/#2} \l_ablation_maximum_tl
    \prop_get:NnN \l_ablation_seconds_prop {#1/#2} \l_ablation_second_tl
    \setlength{\fboxsep}{1.5pt}
    \makebox[2.9em][c]
      {
        \tl_if_eq:NnTF \l_ablation_value_tl {---}
          {---}
          {
            \fp_compare:nTF {\l_ablation_value_tl = \l_ablation_maximum_tl}
              {\colorbox{\l_ablation_color_tl}{\textcolor{white}{\textbf{\l_ablation_value_tl}}}}
              {
                \tl_if_empty:NTF \l_ablation_second_color_tl
                  {\l_ablation_value_tl}
                  {
                    \fp_compare:nTF {\l_ablation_value_tl = \l_ablation_second_tl}
                      {\colorbox{\l_ablation_second_color_tl}{\textcolor{ffInk}{\l_ablation_value_tl}}}
                      {\l_ablation_value_tl}
                  }
              }
          }
      }
    \group_end:
  }

\NewDocumentCommand{\ablationResults}{mm}
  {
    \ablation_cell:nnn {#1} {pnp} {#2} &
    \ablation_cell:nnn {#1} {art} {#2} &
    \ablation_cell:nnn {#1} {avg} {#2}
  }
\ExplSyntaxOff

\title{\raggedright EgoForesight: Effectively Scaling Policy Learning with Noisy Egocentric Actions}

\title{EgoForesight: Effectively Scaling Policy Learning with Noisy Egocentric Actions}
\author{%
  Chubin Zhang\textsuperscript{1,2}\quad
  Changyuan Wang\textsuperscript{1}\quad
  Yirui Wang\textsuperscript{1,5}\quad
  Ao Li\textsuperscript{1,2}\quad
  Shilin Ma\textsuperscript{1}\\[3pt]
  Yue Su\textsuperscript{3}\quad
  Cai Zhou\textsuperscript{4}\quad
  Xiuwei Xu\textsuperscript{5}\quad
  Tianle Zhang\textsuperscript{2}\quad
  Yansong Tang\textsuperscript{1,\ensuremath{\dagger}}\\[8pt]
  {\mdseries\fontsize{10}{12}\selectfont
    \textsuperscript{1}Tsinghua University\quad
    \textsuperscript{2}JD Future Academy\\[2pt]
    \textsuperscript{3}University of Hong Kong\quad
    \textsuperscript{4}Massachusetts Institute of Technology\quad
    \textsuperscript{5}Loopeva}\\[6pt]
  {\normalfont\fontsize{10}{12}\selectfont
    \textsuperscript{\ensuremath{\dagger}}Corresponding author}%
}

\newcommand{\arxivresourcelink}[3]{%
  \noindent\raisebox{-0.2em}{%
    #1}%
  \hspace{0.5em}\textsf{\textbf{#2:}}\hspace{0.45em}\url{#3}\par
}
\renewcommand{\arxivresources}{%
  \par\vspace{3mm}%
  {\raggedright\fontsize{9}{13}\selectfont
    \arxivresourcelink{\includegraphics[width=0.025\linewidth]{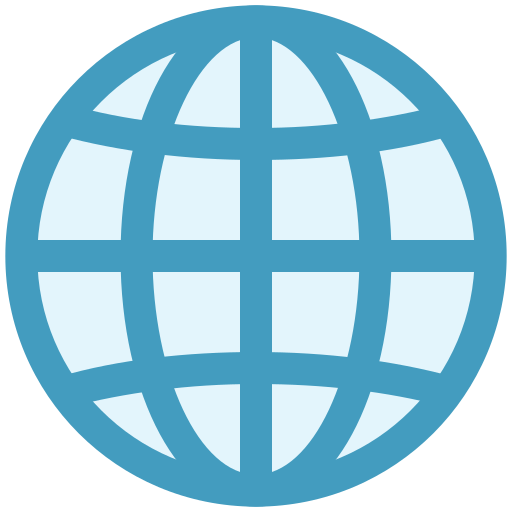}}{Project Page}{https://lin-shan.com/EgoForesight}%
    \vspace{3pt}%
    \arxivresourcelink{\includegraphics[width=0.025\linewidth]{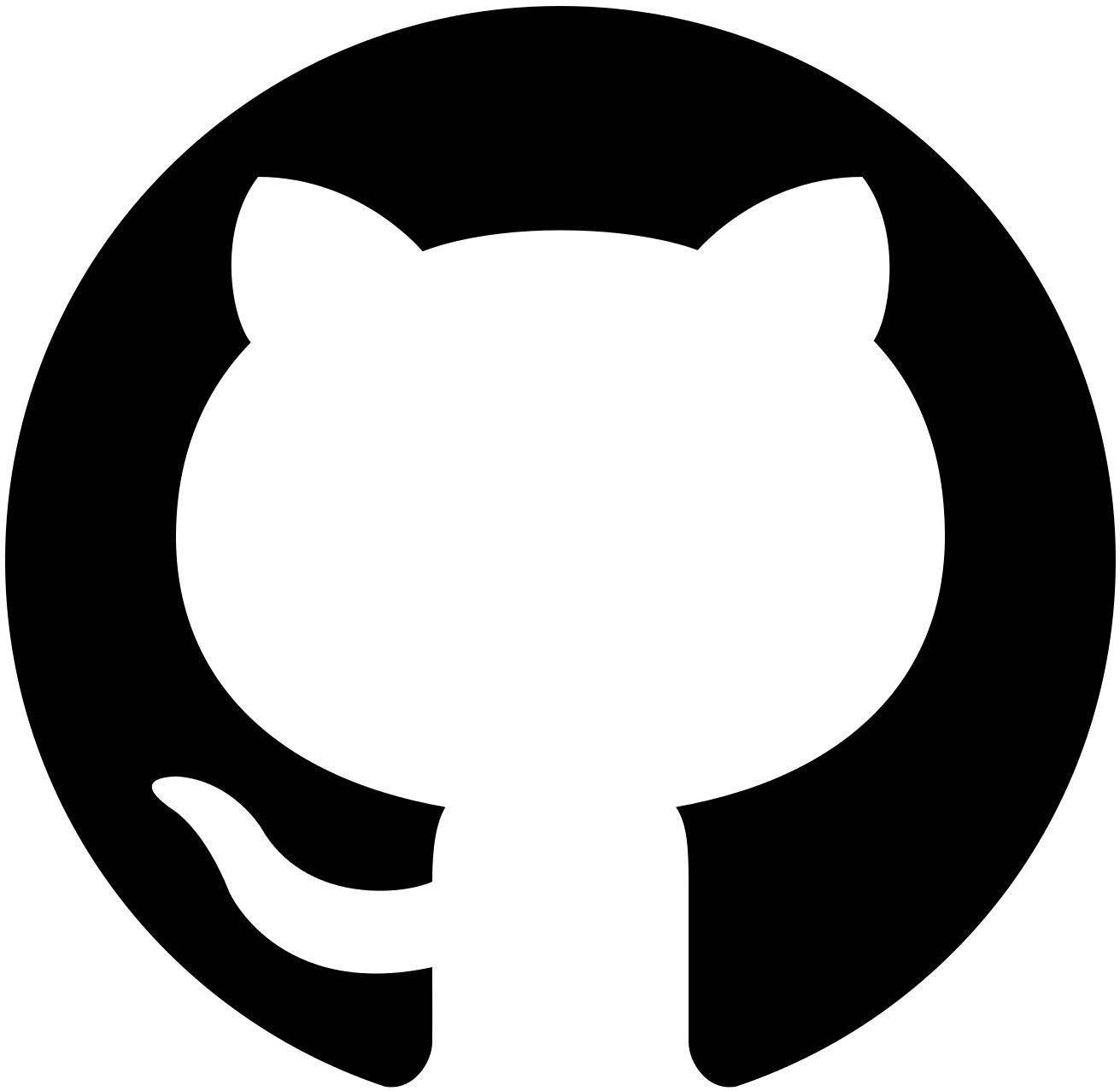}}{GitHub}{https://github.com/LinShan-Bin/EgoForesight}%
    \vspace{3pt}%
    \arxivresourcelink{\includegraphics[width=0.025\linewidth]{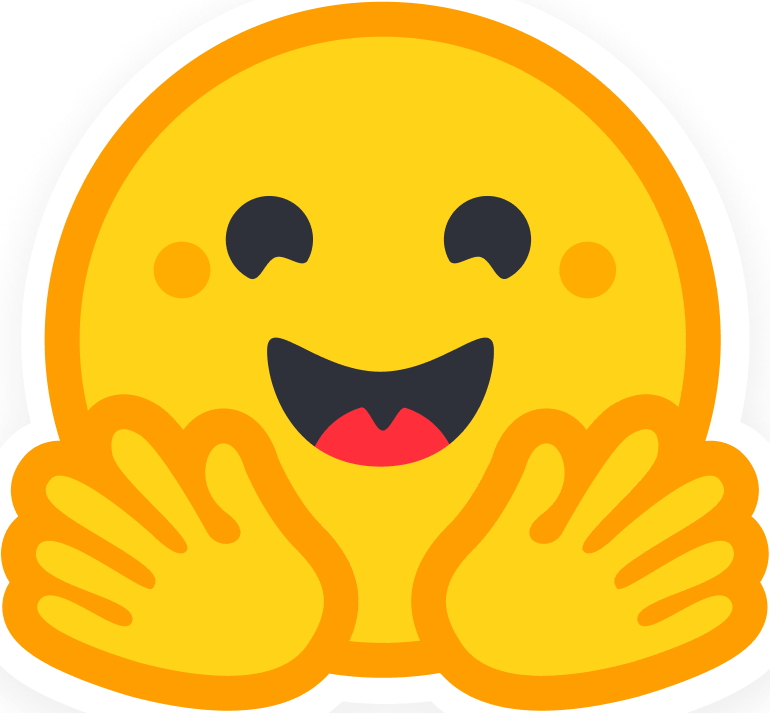}}{HuggingFace}{https://huggingface.co/collections/LinShan/egoforesight}%
  }%
}

\hypersetup{%
  pdftitle={EgoForesight: Effectively Scaling Policy Learning with Noisy Egocentric Actions},
  pdfauthor={Chubin Zhang, Changyuan Wang, Yirui Wang, Ao Li, Shilin Ma, Yue Su, Cai Zhou, Xiuwei Xu, Tianle Zhang, Yansong Tang}%
}

\newcommand{\arxivfinalwords}[1]{%
  \ifdefined\EgoForesightArxiv\mbox{#1}\else#1\fi}

\newcommand{\methodname}{EgoForesight}
\newcommand{\context}{c}
\newcommand{\futuretarget}{y_v}
\newcommand{\actiontarget}{y_a}
\newcommand{\visiontime}{\tau_v}
\newcommand{\actiontime}{\tau_a}

\begin{document}

\maketitle

\begin{abstract}
Scaling policy learning requires broad experience and reliable supervision. Robot demonstrations provide high-quality trajectory supervision but are costly to collect. Egocentric videos broaden task and environment coverage, yet motion reconstruction and retargeting introduce noisy action labels. To address this challenge, we introduce \methodname, a vision-language-action framework that learns jointly from both sources by reducing the influence of unreliable action labels and exploiting demonstrated outcomes as complementary supervision. A joint vision-action flow couples future visual representation prediction with action generation, unifying reliability-aware action learning with outcome-informed representation transfer. Specifically, \emph{reliability-conditioned corruption} allocates action supervision across noise levels according to source reliability, while reliability-weighted regression regulates each source's contribution. Complementing this treatment of noisy labels, \emph{privileged-future representation distillation} draws supervision from demonstrated outcomes. An exponential-moving-average teacher observes the current scene and clean future to encode action-relevant structure from the resulting scene change. Its deeper representations supervise an earlier student layer receiving a corrupted future, providing outcome-informed structure beyond noisy action annotations. Together, these mechanisms support effective scaling with noisy egocentric actions. Over the tested range, \methodname\ exhibits steeper scaling slopes than the baseline, reducing real-world open-loop action MSE by 20.5\% and improving RoboCasa few-shot success by 8.1 percentage points at 30,000 hours of pretraining data. After post-training, it also achieves strong absolute performance in both simulated and real-world dexterous manipulation.

\end{abstract}

\afterpage{\begin{figure}[t]
    \centering
    \includegraphics[width=\linewidth]{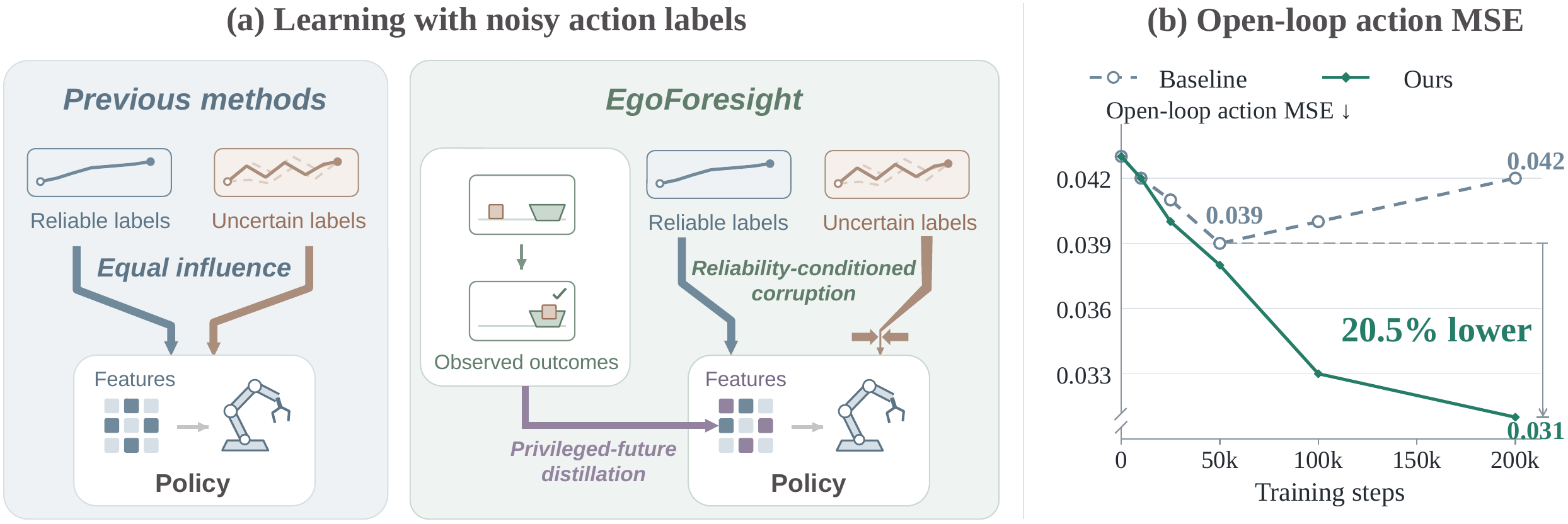}
    \vspace{-15pt}
    \caption{\textbf{Learning with noisy action labels.} \textbf{(a)} Unlike methods that treat action labels equally, \methodname\ accounts for label reliability and enriches action representations through privileged-future distillation. \textbf{(b)} Expansion training on 20\,kh of egocentric data. At each checkpoint, both methods are fine-tuned on real-robot data and evaluated by test-set open-loop action MSE. See Figure~\ref{fig:pretraining_scaling} for a more detailed scaling analysis.}
    \label{fig:teaser}
    \vspace{-10pt}
\end{figure}
}

\raggedbottom
\section{Introduction}
\label{sec:introduction}

Vision-language-action (VLA) models connect broad visual and linguistic knowledge to robot control~\citep{brohan2023rt2,black2024pi0}. We organize their supervision into an embodied data pyramid: vision--language data offer semantic coverage, egocentric demonstrations capture diverse interactions, and robot trajectories provide executable controls but require dedicated hardware and collection effort~\citep{hoque2025egodex,oxe2023,khazatsky2024droid}. EgoMimic~\citep{kareer2024egomimic} and EgoScale~\citep{zheng2026egoscale} demonstrate the value of egocentric data through human--robot co-training and large-scale policy pretraining.

However, recovering actions from egocentric videos introduces annotation noise. Camera tracking and hand-pose reconstruction can produce errors, particularly under occlusion and rapid motion~\citep{zhang2025hawor,hoque2025egodex}. Retargeting adds a geometric mismatch: differences in palm size, finger proportions, and joint structure mean that accurately tracked human poses need not admit an equivalent robot configuration~\citep{wang2024dexcap,xin2026retargeting}. Appearance and dynamics further complicate transfer. Effective learning from egocentric data therefore requires more than increasing its scale: \emph{policies must limit the influence of unreliable action labels while drawing on supervision beyond those labels, enabling additional ego-data to provide effective supervision.}

To meet these requirements, we introduce \methodname, a VLA framework that reduces the influence of unreliable actions while learning from demonstrated outcomes. Inspired by prior unified video--action and dynamics--action models~\citep{zhu2025uwm,lyu2026lda}, our design couples experts for future visual representations and action chunks and trains them with modality-specific flow-matching paths. Separate corruption times control the information available in the vision and action streams, allowing us to regulate noisy action supervision and distill outcome-informed representations within the same model.

First, \emph{reliability-conditioned corruption} allocates action supervision across noise levels according to source reliability. Motivated by ADP's finding that stronger diffusion noise can contract target and suboptimal distributions~\citep{wei2026ambientpolicy}, our continuous source-reliability schedule shifts less reliable actions toward stronger corruption. Reliability-weighted action regression complements this schedule by controlling how strongly each source contributes to learning. Together, they enable learning from diverse egocentric data without assuming uniformly reliable labels.

Second, \emph{privileged-future representation distillation} draws on demonstrated outcomes for action-relevant evidence beyond noisy annotations. An exponential-moving-average (EMA) teacher receives the current observation and observed future at the clean endpoint, acting as a conditional inverse dynamics model (IDM). Initialized from higher-quality demonstrations, the teacher provides action representations informed by the observed scene change. Adapting Self-Flow's cleaner, noise-shifted EMA teacher and cross-layer alignment~\citep{chefer2026selfflow}, we apply this mechanism to robot action tokens: a deeper teacher feature supervises an earlier student layer, which enriches early policy features with outcome-informed structure. In our setting, observed-future conditioning gives the teacher its inverse-dynamics role, and the student learns to anticipate the resulting outcome-informed representation while receiving a corrupted future.

For model training, we first establish visual--action correspondences on higher-quality robot demonstrations and curated egocentric trajectories. This checkpoint initializes the student and teacher before training expands to a larger action-annotated egocentric-data mixture with reliability-conditioned corruption and privileged-future distillation enabled. Our experiments examine scaling as a test of how effectively policies use noisy egocentric supervision. Under matched data and optimization budgets, \methodname\ translates added pretraining data into stronger gains in action prediction and closed-loop control. Ablations support complementary roles for reliability conditioning and future distillation, and show the value of establishing action correspondences on higher-quality data before expanding coverage.

\section{Related Work}
\label{sec:related}

\paragraph{Embodied Foundation Models.}
\citet{levine2016visuomotor} demonstrated that robot policies could be learned end to end from images by jointly training perception and control. Subsequent work expanded this scope toward sharing knowledge across tasks and embodiments, with RT-1 and Octo scaling transformer policies into generalist robot models~\citep{brohan2022rt1,octo2024}. At the action interface, tokenized prediction~\citep{brohan2023rt2,kim2024openvla,pertsch2025fast} and continuous diffusion or flow policies~\citep{chi2023diffusionpolicy,black2024pi0,wu2026lingbotvla} offer different ways to connect perception to control, with flow-based variants grounded in flow matching and rectified flow~\citep{lipman2023flowmatching,liu2023rectifiedflow}. For models built on pretrained VLMs, effective transfer benefits from retaining pretrained semantic competence while grounding spatial and task knowledge in robot actions~\citep{pi2025pi05,driess2025ki,chen2025internvlam1}. This emphasis extends to the information used to condition a policy: \mbox{$\pi_{0.7}$~\citep{pi2026pi07}} combines language, visual subgoals, and execution metadata to exploit heterogeneous data and recombine skills across tasks. Future prediction offers another source of guidance by relating actions to their anticipated effects. This idea developed from video-based planning~\citep{ebert2018foresight} into joint future/action prediction~\citep{wu2023gr1} and a broader family of unified future/action models, many of which use diffusion~\citep{zhu2025uwm,bi2026motus,ye2026dreamzero,cai2026internvlaa1,li2026lingbotva}. Within joint diffusion, separate modality noise times support flexible conditioning~\citep{bao2023unidiffuser} and allow generation schedules such as X-WAM's action-before-video denoising~\citep{guo2026xwam}. The choice of prediction target matters as well: compact future representations can guide or be jointly coupled to action generation~\citep{su2026wog,tang2026lumo2,ma2026internvlaa15}. LDA-1B~\citep{lyu2026lda} uses future DINO features and allocates policy, dynamics, and forecasting objectives according to the quality of the available data. Appendix~\ref{sec:additional_related} discusses the data pyramid, representation learning for generation, and learning \arxivfinalwords{from noisy data.}

\section{\methodname}
\label{sec:method}

\begin{figure}[t]
    \centering
    \includegraphics[width=\linewidth]{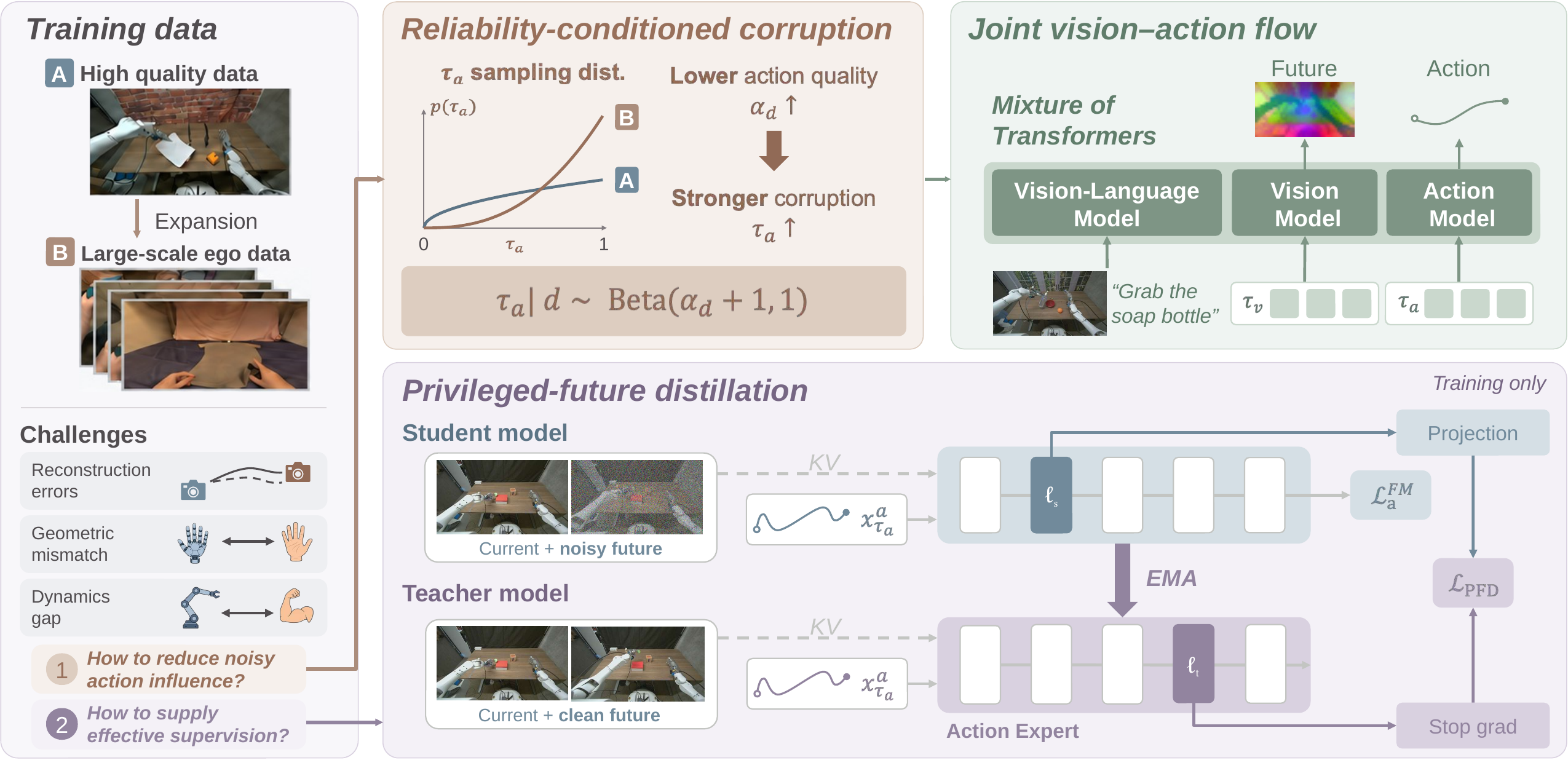}
    \vspace{-15pt}
    \caption{\textbf{Overview of \methodname.} Large-scale human videos expand training data but have less reliable action labels. Reliability-conditioned corruption samples stronger action corruption as reliability falls. The vision--action flow jointly predicts future visual representations and action chunks; during training, an observed-future teacher distills features into the student.}
    \label{fig:foresight_flow_overview}
    \vspace{-10pt}
\end{figure}

\methodname\ addresses two complementary questions: how to reduce the influence of noisy action labels in egocentric data, and how to supply effective supervision beyond those annotations. Both designs operate on the flow timesteps of a joint future--action model. \emph{Reliability-conditioned corruption} assigns less reliable actions greater corruption while keeping future vision comparatively informative. \emph{Privileged-future representation distillation} places a teacher at the observed-future endpoint and transfers its action representations to a student with a corrupted future. Figure~\ref{fig:foresight_flow_overview} summarizes how these two uses of flow time connect data reliability to policy learning.

\subsection{Problem Formulation}
\label{sec:formulation}

We consider datasets \(\mathcal{D}=\bigcup_d\mathcal{D}_d\) of robot and human manipulation demonstrations, all paired with action labels. At step \(t\), the context \(\context\) comprises the current observation \(o_t\), a language instruction, available current state measurements, and embodiment information. For a horizon of \(K\) action steps, a \arxivfinalwords{training example supplies}
\begin{equation}
    \futuretarget=D(o_{t+K})\in\mathbb{R}^{N_v\times d_v},
    \qquad
    \actiontarget=(\widetilde a_t,\ldots,\widetilde a_{t+K-1})
    \in\mathbb{R}^{K\times d_a},
    \label{eq:training_targets}
\end{equation}
where \(D\) is a frozen visual encoder and \(\widetilde a\) is a normalized action label. Robot demonstrations provide recorded controls; human demonstrations provide tracked or reconstructed hand motions and, where applicable, retargeted robot actions. A binary mask \(M_a\in\{0,1\}^{K\times d_a}\) excludes padded action coordinates and timesteps.

Tracking and retargeting errors make these human action labels less reliable (Appendix~\ref{sec:egocentric_action_errors}). We therefore regulate action supervision by source reliability and use observed futures as complementary outcome evidence.

We model future visual features and action chunks jointly as \(p_{\psi,\theta}(\futuretarget,\actiontarget\mid\context)\), with the action marginal defining the policy for the specified embodiment. At deployment, the policy conditions only on \(\context\); the observed future is available solely for training.

\paragraph{Mixture-of-Transformers (MoT).}
\label{sec:architecture}
Our model comprises a pretrained VLM expert and two coupled Transformer experts for future vision and action. VLM features, current visual features, and available state and embodiment inputs form the conditioning inputs \(C_\psi(\context)\). The vision and action experts use modality-specific projections and feed-forward layers with joint bidirectional attention, and both read the context. The VLM retains causal attention and cannot read either generated stream, so future observations never enter \(C_\psi(\context)\). Student objectives nevertheless backpropagate through the conditioning features into the VLM. We denote the coupled experts and their conditioning projections by \(\theta\), and the VLM parameters by \(\psi\).

\paragraph{Joint flow parameterization.}
\label{sec:joint_flow}
Using the flow-matching framework~\citep{lipman2023flowmatching,liu2023rectifiedflow}, we define our own separate interpolation paths for future vision and action:
\begin{equation}
    x_{\visiontime}^{v}=(1-\visiontime)\futuretarget+\visiontime\epsilon_v,
    \qquad
    x_{\actiontime}^{a}=(1-\actiontime)\actiontarget+\actiontime\epsilon_a,
    \label{eq:joint_interpolants}
\end{equation}
where \(\epsilon_v\) and \(\epsilon_a\) are independent standard Gaussian tensors, and \(\visiontime,\actiontime\in[0,1]\) control corruption from the data endpoint at zero to Gaussian noise at one. Action expressions are restricted to valid coordinates. The coupled experts predict
\begin{equation}
    (\widehat u_v,\widehat u_a)
    =F_\theta\bigl(x_{\visiontime}^{v},x_{\actiontime}^{a},
    C_\psi(\context),\visiontime,\actiontime;M_a\bigr),
    \label{eq:coupled_velocity}
\end{equation}
with velocity targets \(u_v^\star=\epsilon_v-\futuretarget\) and \(u_a^\star=\epsilon_a-\actiontarget\). Separate flow times control the information available in each stream, while joint attention lets the action prediction draw on the partially \mbox{resolved future}.

\subsection{Reliability-Conditioned Corruption}
\label{sec:reliability}

We use flow time to allocate action supervision across noise levels. Ambient Diffusion Policy (ADP)~\citep{wei2026ambientpolicy} motivates this noise-dependent use of data: its high-noise region exploits distribution contraction, while its low-noise region exploits shared local motion. In our human--robot mixture, fine human motions need not transfer across hand morphologies, and robot demonstrations already provide substantial local-control supervision. We therefore favor stronger corruption for unreliable actions without introducing ADP's additional low-noise admission window. Each source \(d\) receives a reliability \(q_d\in[0,1]\), where larger values indicate greater trust in its action labels. This training prior determines the corruption parameter through
\begin{equation}
    \alpha_d=\alpha_{\min}+(1-q_d)(\alpha_{\max}-\alpha_{\min}),
    \qquad 0<\alpha_{\min}<\alpha_{\max}.
    \label{eq:reliability_alpha}
\end{equation}
We use \(\alpha_{\min}=0.5\) and \(\alpha_{\max}=3\), so lower reliability gives a larger \(\alpha_d\). The prior expresses relative trust in a source without requiring a calibrated error estimate or a known \mbox{annotation-noise distribution}.

We define a continuous joint corruption law
\begin{equation}
    (\visiontime,\actiontime)\sim\mathcal{S}_{\mathrm{reg}}(q_d),
    \qquad 0\leq\visiontime\leq\actiontime\leq1,
    \label{eq:reliability_schedule}
\end{equation}
with marginals
\begin{equation}
    \visiontime\sim\mathcal{U}(0,1),
    \qquad
    \actiontime\mid d\sim\operatorname{Beta}(\alpha_d+1,1).
    \label{eq:corruption_marginals}
\end{equation}
The sampler preserves both marginals while keeping future vision less corrupted than action. All sources share the full action-time support \((0,1)\); reliability changes the density instead of imposing two hard admission thresholds. This fits mixed-domain VLA co-training by varying supervision frequency smoothly across noise levels and avoiding entire intervals supervised by only a subset of domains. The effect on action \arxivfinalwords{inputs is explicit:}
\begin{equation}
    \mathbb{E}_{\mathrm{reg}}[1-\actiontime\mid d]
    =\frac{1}{\alpha_d+2},
    \qquad
    \mathbb{E}_{\mathrm{reg}}[\actiontime-\visiontime\mid d]
    =\frac{\alpha_d}{2(\alpha_d+2)}.
    \label{eq:reliability_moments}
\end{equation}
Lower reliability reduces the expected contribution of the action label to the input and increases the vision--action corruption gap, leaving the vision marginal unchanged. This encourages the action expert to use the current context and the comparatively more informative future. Appendix~\ref{sec:reliability_derivation} provides the \arxivfinalwords{sampler and derivation.}

Flow matching learns a conditional average of velocity targets, so the contribution of annotation noise is governed by the resulting conditional field. Appendix~\ref{sec:population_interpretation} analyzes how stronger corruption brings the annotation and reference fields closer under a simple noise model. We additionally multiply the direct action loss by \(q_d\) to regulate each source's contribution to learning. Timestep sampling and loss weighting thus determine where and how strongly each source supervises the joint model.

\subsection{Privileged-Future Representation Distillation}
\label{sec:pfd}

The observed future provides supervision beyond action labels. At \(\visiontime'=0\), the action expert sees both the current observation and the demonstrated outcome, acting as a conditional inverse dynamics model (IDM). Trained on higher-quality demonstrations, its action estimates can be more reliable than labels affected by reconstruction or retargeting errors. An exponential-moving-average (EMA) teacher in this regime transfers outcome-informed action features to a student with a \mbox{corrupted future}.

The teacher maintains an EMA of the coupled experts,
\begin{equation}
    \bar\theta\leftarrow\mu\bar\theta+(1-\mu)\theta,
    \qquad 0\leq\mu<1,
    \label{eq:teacher_ema}
\end{equation}
while reusing the student's contextual features with gradients stopped. Given the student input, we construct a teacher input with the observed future and the corrupted action:
\begin{equation}
    \visiontime'=0,\qquad \actiontime'=\actiontime,
    \qquad
    x_{\actiontime'}^{a}=(1-\actiontime')\actiontarget+\actiontime'\epsilon_a.
    \label{eq:teacher_input}
\end{equation}
The teacher receives \(\futuretarget\), \(x_{\actiontime'}^a\), and the detached context \(\operatorname{sg}(C_\psi(\context))\), where \(\operatorname{sg}\) denotes stop-gradient. It shares the student's action mask, target, and Gaussian noise, keeping both action inputs on the same interpolation path. The EMA covers the coupled experts and conditioning projections; VLM features are reused from the student. To calibrate this clean-future regime, we set \(\visiontime=0\) for a small fraction \(p_{\mathrm{edge}}\) of student examples, retaining action flow and representation losses but applying \emph{no vision flow loss}. Appendix~\ref{sec:reliability_derivation} gives \arxivfinalwords{the complete sampler.}

Adapting Self-Flow's projection-based cosine alignment between an earlier student layer and a deeper EMA-teacher layer~\citep{chefer2026selfflow}, we apply the mechanism to action tokens whose teacher representations are informed by the observed future. Let \(h_{S,k}^{\ell_s}\) and \(h_{T,k}^{\ell_t}\) denote the student and teacher action features at token \(k\), with \(\ell_s<\ell_t\). A trainable projection \(P_\phi\) matches the features through a cosine loss:
\begin{equation}
    \ell_{\mathrm{PFD},k}
    =1-\cos\!\left(P_\phi(h_{S,k}^{\ell_s}),
      \operatorname{sg}(h_{T,k}^{\ell_t})\right).
    \label{eq:pfd}
\end{equation}
In our construction, the flow-time gap trains early student features to anticipate deeper representations informed by the observed future.

\paragraph{Two-stage data selection.}
To ground distillation in reliable supervision, we first learn visual--action correspondences from higher-quality robot demonstrations and curated human trajectories through joint flow matching and VLM co-training, without distillation. The resulting checkpoint initializes both the student and its EMA teacher. Stage two retains these sources and adds a much larger collection of action-annotated human videos, enabling both proposed components. Appendix~\ref{sec:implementation} details the datasets and their \arxivfinalwords{allocation across stages.}

\suppressfloats[t]
\begin{figure}[t]
    \centering
    \setlength{\abovecaptionskip}{5pt}
    \includegraphics[width=\linewidth]{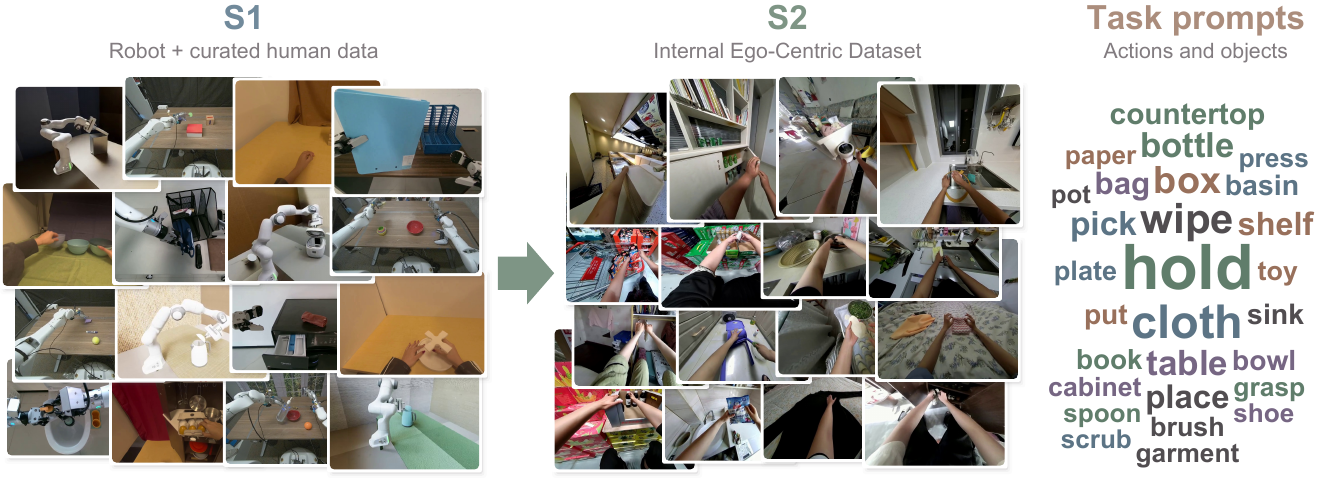}
    \caption{\textbf{Two-stage pretraining data.} Left: examples from AgiBot World, EgoDex, InternData-A1, and T-Rex used in S1. Center: examples from the Internal Ego-Centric Dataset, which adds 20,000 hours to S2 alongside all S1 sources. Right: task terms sized by frequency in 8,192 sampled instructions from this dataset, excluding stopwords and generic descriptors.}
    \label{fig:pretraining_data}
\end{figure}

\subsection{Training Objective}
\label{sec:objective}

Action labels are mapped to a canonical action space and normalized; the space and dataset-specific mappings are detailed in Appendix~\ref{sec:canonical_action}. We train the joint flow model with mean \mbox{squared error (MSE)}:
\begin{equation}
    \mathcal L_v^{\mathrm{FM}}
    =\mathbb E\bigl[\operatorname{MSE}(\widehat u_v,u_v^\star)\bigr],
    \qquad
    \mathcal L_a^{\mathrm{FM}}
    =\mathbb E\bigl[q_d\operatorname{MSE}(\widehat u_a,u_a^\star)\bigr].
    \label{eq:expected_losses}
\end{equation}
MSE averages over valid coordinates, and the vision loss excludes clean-future boundary examples. The distillation loss \(\mathcal L_{\mathrm{PFD}}\) averages Equation~\eqref{eq:pfd} over valid action tokens without reliability weighting, preserving feature supervision for less reliable sources. Appendix~\ref{sec:loss_details} gives the exact \mbox{batch reductions}.

During continual adaptation to embodied tasks, we retain the autoregressive cross-entropy \(\mathcal L_{\mathrm{AR}}\) on interleaved vision--language data to preserve general visual--language knowledge and reduce forgetting. Combining this objective with the flow and distillation losses gives
\begin{equation}
    \min_{\psi,\theta,\phi}\;
    \lambda_{\mathrm{AR}}\mathcal{L}_{\mathrm{AR}}
    +\lambda_v\mathcal{L}_{v}^{\mathrm{FM}}
    +\lambda_a\mathcal{L}_{a}^{\mathrm{FM}}
    +\lambda_T\mathcal{L}_{\mathrm{PFD}},
    \label{eq:full_objective}
\end{equation}
with nonnegative loss weights. Stage one uses unit action weights and \(\lambda_T=0\); stage two enables both proposed components. The teacher is updated only by EMA.

\flushbottom
\section{Experiments}
\label{sec:experiments}

\raggedbottom

Our experiments address three questions: (i) does reliability-conditioned corruption improve learning from noisy action labels (Section~\ref{sec:reliability}), (ii) does privileged-future representation distillation provide effective policy supervision (Section~\ref{sec:pfd}), and (iii) how does the complete \methodname\ compare with other policies on downstream tasks? We evaluate overall performance on RoboCasa and real-world tasks. Our ablations examine the two proposed mechanisms, pretraining-data scaling, the allocation of data across training stages, and the use of future information during generation.

\subsection{Model Pretraining}
\label{sec:model_pretraining}

We pretrain in two stages (Figure~\ref{fig:pretraining_data}). S1 uses AgiBot World, EgoDex, InternData-A1, and T-Rex; its human videos undergo clip-level screening for visual quality, hand visibility, and global motion. S2 retains these sources and adds 20,000 hours of action-annotated egocentric human data, enabling reliability-conditioned corruption and privileged-future representation distillation after the S1 initialization. All manipulation sources provide action labels.

Our implementation selects a 4B backbone from the Qwen3.5 family~\citep{qwen2026qwen35} and pairs it with our coupled vision and action experts; we use a DINOv3 encoder~\citep{simeoni2025dinov3} as a frozen visual-target extractor. The action expert predicts 30-step chunks of normalized actions in a shared 130-dimensional canonical space, with unavailable coordinates masked. We use Muon~\citep{jordan2024muon} for selected hidden-layer weight matrices and auxiliary AdamW for the remaining parameters. We train S1 for 100k updates and S2 for 200k updates, using a global batch size of 2,048 and 5k learning-rate warmup steps in both stages. Appendix~\ref{sec:implementation} provides the dataset table, filtering rules, architecture, action layout, \arxivfinalwords{and training settings.}

\subsection{Downstream Evaluation}
\label{sec:downstream}

\paragraph{RoboCasa.}
We use RoboCasa's simulator and household-task assets~\citep{nasiriany2024robocasa} to evaluate dexterous manipulation with a humanoid robot. Following DIAL's GR1 Tabletop suite~\citep{chen2026dial}, we consider 24 tasks: 18 pick-and-place (PnP) tasks and 6 articulated-object tasks that combine placement with closing a cabinet, drawer, or microwave. We post-train on 100 demonstrations per task in the few-shot setting and 1,000 per task in the full-data setting, and evaluate each task over 50 episodes. Privileged-future distillation remains enabled in both settings. Appendix~\ref{sec:robocasa_protocol} gives the training and evaluation details.

\paragraph{Real world.}
We evaluate four tasks on a Galaxea R1 Pro with Sharpa Wave hands, as shown in Figure~\ref{fig:real_world_tasks}. We control only the two arms and hands, totaling 58 DoF, using relative end-effector commands for the arms and joint commands for the hands. For \methodname, $\pi_{0.5}$~\citep{pi2025pi05}, and GR00T N1.7~\citep{nvidia2026gr00tn17}, we jointly post-train one model per method on 100 teleoperated demonstrations per task and evaluate full-task success over 20 trials per task. Appendix~\ref{sec:real_world_protocol} provides protocol details.

\subsection{Overall Performance}
\label{sec:overall_performance}

\begin{table}[!t]
\caption{\textbf{RoboCasa evaluation.} Success rates (\%) over 18 pick-and-place (PnP), 6 articulated-object (Art.), and all 24 tasks (Avg.). Blue and green denote full-data and few-shot training; dark and light shading mark the best and second-best score per column within each setting. Baseline variants and sources are detailed in Appendix~\ref{sec:robocasa_protocol}.}
\label{tab:robocasa_results}
\vspace{3pt}
\centering
\definecolor{ffInk}{HTML}{514D50}
\definecolor{ffMuted}{HTML}{827C80}
\definecolor{ffArrow}{HTML}{C3C5C2}
\definecolor{ffBlue}{HTML}{718A9B}
\definecolor{ffBluel}{HTML}{D5DFE5}
\definecolor{ffBluebg}{HTML}{EEF2F4}
\definecolor{ffSage}{HTML}{7E9585}
\definecolor{ffSagel}{HTML}{CAD8CD}
\definecolor{ffClay}{HTML}{AB8D7C}
\definecolor{ffClayl}{HTML}{DDCDC0}
\definecolor{ffLav}{HTML}{92849F}
\definecolor{ffLavl}{HTML}{D9D1E0}
\definecolor{ffLavbg}{HTML}{F4F1F7}
\definecolor{ffData}{HTML}{F4F2F3}
\definecolor{ffNoise}{HTML}{F6F0EC}
\definecolor{ffJoint}{HTML}{EFF4F2}
\definecolor{ffPaper}{HTML}{FFFFFF}
\definecolor{ffBlueDark}{HTML}{5C7485}
\definecolor{ffSageDark}{HTML}{607D6B}
\definecolor{ffClayDark}{HTML}{98705B}
\definecolor{ffLavDark}{HTML}{766683}
\definecolor{ffGrid}{HTML}{DAD6DA}
\definecolor{ffFloor}{HTML}{ECE8EC}

\footnotesize
\setlength{\tabcolsep}{3pt}
\setlength{\fboxsep}{2pt}
\renewcommand{\arraystretch}{1.17}
\newcommand{\rcbest}[2]{\colorbox{#1}{\makebox[2.3em][c]{\textcolor{white}{\textbf{#2}}}}}
\newcommand{\rcsecond}[2]{\colorbox{#1}{\makebox[2.3em][c]{\textcolor{ffInk}{#2}}}}
\begin{tabular*}{\linewidth}{@{\extracolsep{\fill}}lcc c@{\hspace{12pt}}lcc c@{}}
\toprule
\multicolumn{4}{c}{\textcolor{ffBlueDark}{\textit{Full data (1,000 demonstrations / task)}}} &
\multicolumn{4}{c}{\textcolor{ffSageDark}{\textit{Few-shot (100 demonstrations / task)}}} \\
\cmidrule(r{6pt}){1-4}\cmidrule(l{6pt}){5-8}
Method & PnP & Art. & Avg. & Method & PnP & Art. & Avg. \\
\midrule
GR00T N1.6 & 55.4 & 24.2 & 47.6 & GR00T-Qwen2.5 & 17.1 & 36.0 & 21.8 \\
LDA-1B & 55.1 & 56.3 & 55.4 & GR00T-Qwen2.5-FT & 26.9 & 41.7 & 30.6 \\
RLDX-1 & 59.0 & 57.7 & 58.7 & \quad + FLARE & 47.7 & 64.7 & 51.9 \\
JoyAI-RA & 60.3 & 71.7 & 63.2 & DIAL (robot-only) & 56.0 & \rcsecond{ffSagel}{65.3} & 58.3 \\
DIAL (robot-only) & \rcsecond{ffBluel}{68.9} & \rcsecond{ffBluel}{74.3} & \rcsecond{ffBluel}{70.2} & DIAL + EgoDex & \rcsecond{ffSagel}{60.8} & 62.0 & \rcsecond{ffSagel}{61.1} \\
\textbf{\methodname\ (Ours)} & \rcbest{ffBlueDark}{71.9} & \rcbest{ffBlueDark}{80.0} & \rcbest{ffBlueDark}{73.9} & \textbf{\methodname\ (Ours)} & \rcbest{ffSageDark}{64.8} & \rcbest{ffSageDark}{69.7} & \rcbest{ffSageDark}{66.0} \\
\bottomrule
\end{tabular*}
\vspace{-10pt}
\end{table}

\paragraph{Dexterous manipulation on RoboCasa.}
Table~\ref{tab:robocasa_results} shows that \methodname\ reaches 66.0\% success with only 100 robot demonstrations per task, improving on robot-only DIAL by \textbf{7.7 percentage points} and on DIAL with EgoDex by \textbf{4.9 points}. With full data, our success rate rises to 73.9\%, exceeding DIAL by 3.7 points. Reducing the robot demonstrations by 90\% lowers our success rate by only 7.9 points, compared with 11.9 points for robot-only DIAL. These results highlight effective adaptation to dexterous tasks when robot supervision is limited. The comparison uses published baseline results with different pretraining data and optimization budgets, as detailed in Appendix~\ref{sec:robocasa_protocol}.

\begin{figure}[!t]
\centering
\begingroup
\input{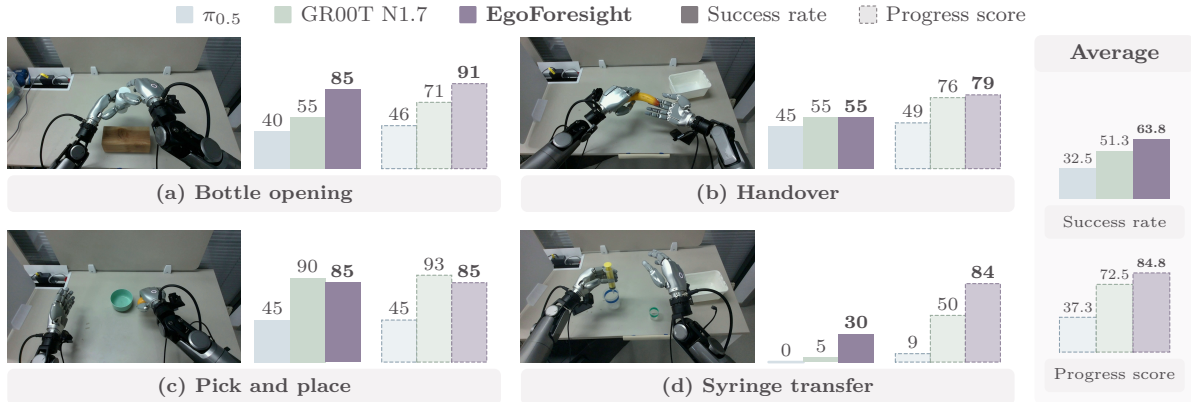}
{\fontsize{8}{9}\selectfont\color{ffInk}%
    \RWLegend{fill=ffBluel}{$\pi_{0.5}$}\hspace{12pt}%
    \RWLegend{fill=ffSagel}{GR00T N1.7}\hspace{12pt}%
    \RWLegend{fill=ffLav}{\textbf{\methodname}}\hspace{18pt}%
    \RWLegend{fill=ffMuted}{Success rate}\hspace{12pt}%
    \RWLegend{fill=ffMuted!18!white,draw=ffMuted,line width=0.4pt,
        dash pattern=on 1.25pt off 0.9pt}{Progress score}\par}
\vspace{3pt}
\RWFigure\par
\setlength{\abovecaptionskip}{5pt}
\caption{\textbf{Real-world dexterous manipulation.} Each task pairs a head-camera image with success rates (\%, solid bars) and progress scores (0--100, dashed borders) on the R1 Pro with Sharpa Wave hands. The right column reports the equal-weight mean across the four tasks. The evaluation protocol uses 100 demonstrations and 20 trials per task.}
\label{fig:real_world_tasks}
\endgroup
\end{figure}

\paragraph{Real-world dexterous manipulation.}
\methodname\ demonstrates strong overall dexterous manipulation performance in Figure~\ref{fig:real_world_tasks}, achieving the highest mean success rate of 63.8\% and progress score of 84.8. GR00T N1.7 was pretrained on R1 Pro--Sharpa, whereas ours saw only limited Sharpa-hand data on other robots, indicating strong cross-embodiment adaptation under the same downstream demonstration budget. Our advantage is most pronounced on bottle opening and syringe transfer, which demand precise manipulation, with the latter additionally requiring sustained coordination over a long sequence of tool-use stages. For $\pi_{0.5}$, the pretrained checkpoint supports only 32 action dimensions, insufficient for our 58-DoF embodiment, so we replace its input and output projection layers to accommodate the larger action space, as detailed in Appendix~\ref{sec:real_world_protocol}. This change alters the pretrained action interface and may affect performance on difficult tasks, underscoring that transferring from grippers to dexterous hands is nontrivial.

\subsection{Ablation}
\label{sec:ablations}

\subsubsection{Pretraining}
\label{sec:pretrain_ablations}

We evaluate pretraining components and data allocation through RoboCasa few-shot adaptation at one tenth of the pretraining data. We additionally vary the pretraining scale and measure real-world open-loop error and RoboCasa closed-loop success. Detailed settings are in Appendix~\ref{sec:ablation_protocol}.

\begin{table}[!ht]
\centering
\begin{minipage}[t]{0.38\linewidth}
\vspace{0pt}
\setlength{\belowcaptionskip}{6pt}
\caption[Pretraining components]{\raggedright \textbf{Pretraining components.} RoboCasa few-shot success (\%) at $1/10$ pretraining scale. Ours combines both components; shading marks column maxima.}
\label{tab:ablation_plan}
\centering
\small
\setlength{\tabcolsep}{1.5pt}
\ifdefined\EgoForesightArxiv
\renewcommand{\arraystretch}{1.45}
\else
\renewcommand{\arraystretch}{1.18}
\fi
\ablationValues{preReduced}{ffSageDark}
  {53.4, 53.1, 55.8, 57.0}
  {47.7, 58.0, 57.7, 63.3}
  {52.0, 54.3, 56.3, 58.6}
\begin{tabular*}{\linewidth}{@{\extracolsep{\fill}}l|rr|r@{}}
\toprule
\textbf{Variant} & \textbf{PnP} & \textbf{Art.} & \textbf{Avg.} \\
\midrule
Baseline & \ablationResults{preReduced}{1} \\
w/ reliability & \ablationResults{preReduced}{2} \\
w/ teacher & \ablationResults{preReduced}{3} \\
\midrule
\textbf{Ours} & \ablationResults{preReduced}{4} \\
\bottomrule
\end{tabular*}
\end{minipage}\hfill
\begin{minipage}[t]{0.60\linewidth}
\setlength{\parskip}{0pt}
\ifdefined\EgoForesightArxiv
\vspace{3.6pt}
\else
\vspace{8.87pt}
\fi
\centering
\includegraphics[width=\linewidth]{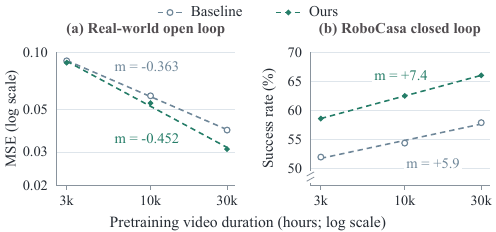}\par
\nointerlineskip
\setlength{\abovecaptionskip}{3pt}
\captionof{figure}[Pretraining-data scaling]{\raggedright \textbf{Pretraining-data scaling.} Dashed lines show fits in the plotted coordinates; $m$ denotes the slope.}
\label{fig:pretraining_scaling}
\end{minipage}
\end{table}

\paragraph{Reliability and privileged teacher.}
\label{sec:reliability_ablation}
\label{sec:teacher_ablation}
Table~\ref{tab:ablation_plan} shows that reliability at $q_d=0.2$ and the privileged teacher are complementary: individually, they raise reduced-scale average success from 52.0\% to 54.3\% and 56.3\%; together, they reach 58.6\%, exceeding either component alone. PnP primarily probes generalization across objects, where the complete method improves success by 3.6 percentage points at reduced scale. Art. additionally demands accurate action generation to manipulate articulated fixtures. Here, reliability alone yields a large gain from 47.7\% to 58.0\%, despite nearly unchanged PnP performance, and adding the teacher further raises success to 63.3\%. This pattern supports assigning different trust to higher-quality robot action labels and noisier human labels, while using privileged future evidence for complementary supervision.

Average success is relatively insensitive to the precise reliability value within the tested range: human $q_d\in\{0.2,0.5,0.8\}$ gives 54.0--54.3\%, consistently above 52.0\% at $q_d=1$, as shown in Table~\ref{tab:ablation_reliability} of Appendix~\ref{sec:ablation_protocol}. We therefore use simple source-level reliability rather than introducing task- or sample-specific estimates, which would add complexity.

\paragraph{Pretraining-data scaling.}
\label{sec:scaling_ablation}
We subsample each pretraining source at $1/10$, $1/3$, and full scale, yielding nested 3k-, 10k-, and 30k-hour subsets with matched source proportions. Each scale follows the same two-stage recipe, retaining S1 replay in S2. Baseline and \methodname\ use identical subsets and optimization budgets at each scale; downstream data and adaptation budgets are fixed. The~baseline disables reliability conditioning and privileged-future distillation; video hours count unique data and exclude repeated sampling or S1 replay. In Figure~\ref{fig:pretraining_scaling}, we fit $\log_{10}(\mathrm{MSE})$ and success percentage against $\log_{10}(h)$ by ordinary least squares over the three nonzero scales. The real-world open-loop MSE slope is $-0.452$ for ours versus $-0.363$ for the baseline; RoboCasa closed-loop success increases by 7.4 versus 5.9 percentage points per decade of data. \textbf{\emph{Both slope magnitudes are about 25\% larger}}, indicating stronger returns from added data within the tested range. At 30k hours, MSE is 20.5\% lower and success is 8.1 points higher. The MSE comparison uses the baseline's best checkpoint rather than its overfitted endpoint.

\paragraph{Data allocation.}
\label{sec:allocation_ablation}
Table~\ref{tab:ablation_allocation}(a) compares S2$\to$S1 ordering, joint S1+S2 training, and S1$\to$S2. Our S1$\to$S2 schedule replays S1 data throughout S2 alongside the additional egocentric data, rather than switching exclusively to the latter. We call the first ordering EgoScale-style~\citep{zheng2026egoscale} only in the sense that broad human data precede higher-quality aligned data; our stages, datasets, and objectives differ. S1$\to$S2 achieves the highest average success of 58.6\%, exceeding reverse ordering by 1.0 point and joint training by 4.3 points. It also yields the best PnP score, although reverse ordering is slightly better on Art. Ending with higher-quality S1 data emphasizes action precision, but may partially overwrite the broader object and scene knowledge acquired during S2, consistent with this trade-off. These results suggest a promising scaling strategy: establish a strong action model with higher-quality data, then expand its coverage using \arxivfinalwords{broad human data.}

\subsubsection{Post-training}
\label{sec:posttrain_ablations}
\label{sec:trajectory_ablation}
Table~\ref{tab:ablation_posttraining}(b) shows that privileged-future distillation improves average success during post-training from the same full-scale pretrained checkpoint. This supports its ability to learn better action representations during downstream adaptation. Using $\visiontime=\actiontime^{1.5}$ at inference then raises Art. success to 80.0\% while retaining 71.9\% on PnP, yielding the best average of 73.9\%. This supports a vision-leading denoising path consistent with the training-time ordering $\visiontime<\actiontime$. The last two variants share policy weights and integration budget.

\begin{table}[!ht]
\setlength{\belowcaptionskip}{2pt}
\caption{\textbf{Ablation of data allocation and post-training.} RoboCasa success rates (\%): \textbf{(a)} pretraining order in the few-shot setting; \textbf{(b)} teacher distillation and denoising path in the full setting. Blue shading marks the best score per column within each panel.}
\label{tab:ablation_allocation}
\label{tab:ablation_posttraining}
\centering
\small
\setlength{\tabcolsep}{2.5pt}
\renewcommand{\arraystretch}{1.20}
\ablationValues{allocation}{ffBlueDark}
  {55.2, 53.1, 57.0}
  {64.7, 58.0, 63.3}
  {57.6, 54.3, 58.6}
\begin{minipage}[t]{0.46\linewidth}
\vspace{0pt}
\centering
\textbf{(a) Data allocation}\par\vspace{5pt}
\begin{tabular*}{\linewidth}{@{\extracolsep{\fill}}l|rr|r@{}}
\toprule
\shortstack[l]{\textbf{Pretraining}\\\textbf{order}} & \textbf{PnP} & \textbf{Art.} & \textbf{Avg.} \\
\midrule
S2 $\to$ S1 & \ablationResults{allocation}{1} \\
S1 + S2 (joint) & \ablationResults{allocation}{2} \\
S1 $\to$ S2 & \ablationResults{allocation}{3} \\
\bottomrule
\end{tabular*}
\end{minipage}\hfill
\begin{minipage}[t]{0.51\linewidth}
\vspace{0pt}
\centering
\textbf{(b) Post-training}\par\vspace{5pt}
\ablationValues{posttraining}{ffBlueDark}
  {67.0, 72.0, 71.9}
  {81.7, 75.0, 80.0}
  {70.7, 72.8, 73.9}
\begin{tabular*}{\linewidth}{@{\extracolsep{\fill}}cc|rr|r@{}}
\toprule
\shortstack{\textbf{Teacher}\\\textbf{distillation}} & \shortstack{\textbf{Denoising}\\\textbf{path}} & \textbf{PnP} & \textbf{Art.} & \textbf{Avg.} \\
\midrule
Off & $\visiontime=\actiontime$ & \ablationResults{posttraining}{1} \\
On & $\visiontime=\actiontime$ & \ablationResults{posttraining}{2} \\
On & $\visiontime=\actiontime^{1.5}$ & \ablationResults{posttraining}{3} \\
\bottomrule
\end{tabular*}

\end{minipage}
\end{table}

\flushbottom

\section{Conclusion}
\label{sec:conclusion}

We presented \methodname, which makes noisy egocentric experience more useful for policy learning through reliability-aware action supervision and privileged-future representation distillation. Within a joint visual--action flow model, these mechanisms regulate the influence of unreliable action labels while learning complementary structure from demonstrated outcomes. Scaling experiments and component ablations support this combination, with improved returns from added pretraining data and strong downstream performance in simulation and on real hardware. Together, these results suggest that effective scaling depends on supervision quality as well as data volume.

These findings motivate extending outcome-based supervision to videos without action annotations, using self-supervised objectives to learn interaction structure without reconstructed controls. At larger scales, a key question is whether broader experience enables compositional generalization to unseen combinations of objects, skills, and scenes. Studying how supervision quality and data diversity affect this capability could clarify when scaling produces reusable manipulation knowledge rather than better coverage of familiar combinations.

\clearpage

\bibliography{reference}
\bibliographystyle{iclr2027_conference}

\clearpage
\appendix
\raggedbottom
\section{Reliability-conditioned sampling}
\label{sec:reliability_derivation}

This appendix specifies the corruption distribution \(\mathcal S(q_d)\) used in Sections~\ref{sec:reliability}--\ref{sec:pfd}. It combines ordinary examples from \(\mathcal S_{\mathrm{reg}}(q_d)\), with a strictly positive vision--action time gap, and clean-future boundary examples.

\paragraph{Ordinary examples.}
For source reliability \(q_d\in[0,1]\), use the mapping in Equation~\eqref{eq:reliability_alpha}, with \(\alpha_{\min}=0.5\) and \(\alpha_{\max}=3\). Independently sample
\begin{equation}
    \visiontime\sim\mathcal{U}(0,1),\qquad
    g\sim\operatorname{Beta}(\alpha_d,2),\qquad
    \actiontime=\visiontime+(1-\visiontime)g.
    \label{eq:reliability_sampling}
\end{equation}
Since \(0<g<1\) almost surely, this rule gives \(0<\visiontime<\actiontime<1\) almost surely and assigns no point mass to \(\visiontime=\actiontime\). At fixed \(\visiontime\), increasing \(\alpha_d\) shifts the action time upward in first-order stochastic order. To see this, for \(\alpha_2>\alpha_1>0\), the ratio of the two gap densities is proportional to \(g^{\alpha_2-\alpha_1}\), which is increasing on \((0,1)\). The affine map from \(g\) to \(\actiontime\) preserves this ordering. Hence decreasing reliability increases action corruption while leaving the vision-time distribution unchanged.

\paragraph{Action-time marginal.}
Write \(\alpha=\alpha_d\). The density of \(g\) is \(\alpha(\alpha+1)g^{\alpha-1}(1-g)\). Changing variables from \(g\) to \(\actiontime\) gives the joint density
\begin{equation}
    f_{\visiontime,\actiontime}(v,a\mid d)
    =\frac{\alpha(\alpha+1)(a-v)^{\alpha-1}(1-a)}{(1-v)^{\alpha+1}},
    \qquad 0<v<a<1.
\end{equation}
For \(0<a<1\), integrate over \(v\) and substitute \(r=(a-v)/(1-v)\):
\begin{equation}
\begin{aligned}
    f_{\actiontime}(a\mid d)
    &=\int_0^a
    \frac{\alpha(\alpha+1)(a-v)^{\alpha-1}(1-a)}{(1-v)^{\alpha+1}}\,dv\\
    &=\alpha(\alpha+1)\int_0^a r^{\alpha-1}\,dr
    =(\alpha+1)a^\alpha.
\end{aligned}
\end{equation}
Thus, for ordinary examples,
\begin{equation}
\begin{aligned}
    \actiontime\mid d&\sim\operatorname{Beta}(\alpha_d+1,1),\\
    \mathbb{E}[\actiontime\mid d]&=\frac{\alpha_d+1}{\alpha_d+2},
    \qquad
    \mathbb{E}[\actiontime-\visiontime\mid d]=\frac{\alpha_d}{2(\alpha_d+2)}.
\end{aligned}
    \label{eq:action_time_marginal}
\end{equation}
The last identity also follows directly from independence of \(g\) and \(\visiontime\). Both the mean action time and the expected gap increase as reliability decreases.

\paragraph{Boundary examples and the complete distribution.}
Independently select a boundary indicator \(e\sim\operatorname{Bernoulli}(p_{\mathrm{edge}})\). When \(e=0\), use Equation~\eqref{eq:reliability_sampling}; when \(e=1\), sample
\begin{equation}
    \visiontime=0,\qquad u\sim\mathcal{U}(0,1),\qquad
    \actiontime=u^{1/\alpha_d}\sim\operatorname{Beta}(\alpha_d,1).
    \label{eq:boundary_sampling}
\end{equation}
Consequently, the full training distribution has marginals
\begin{align}
    \visiontime\mid d
    &\sim(1-p_{\mathrm{edge}})\mathcal{U}(0,1)
    +p_{\mathrm{edge}}\delta_0,\nonumber\\
    \actiontime\mid d
    &\sim(1-p_{\mathrm{edge}})\operatorname{Beta}(\alpha_d+1,1)
    +p_{\mathrm{edge}}\operatorname{Beta}(\alpha_d,1),
    \label{eq:complete_time_marginals}
\end{align}
where \(\delta_0\) is a point mass at zero. The Beta marginal in Equation~\eqref{eq:action_time_marginal} therefore applies only to ordinary examples. The shared \(p_{\mathrm{edge}}\) preserves the source-independent vision marginal, and the action marginal remains stochastically increasing as reliability decreases. The boundary action distribution has positive density throughout \((0,1)\), covering the action times used by the privileged teacher without requiring an identical density. Boundary examples retain valid-action flow and representation supervision, but their vision flow loss is masked out. This calibrates action prediction conditioned on the observed future.

\subsection{Population interpretation of action corruption}
\label{sec:population_interpretation}

We analyze the action flow loss at fixed flow times and conditioning \(z=(\context,x_{\visiontime}^{v},\visiontime,M_a)\), including the partially corrupted future available to the action expert. For this calculation, write \(\actiontarget=y_a^0+\eta\), with \(y_a^0\mid z\sim\mathcal N(m,\Sigma_p)\) and \(\eta\mid z\sim\mathcal N(0,\Sigma_\eta)\) independent conditional on \(z\). The reference and annotation laws, denoted by \(p\) and \(r\), share the mean \(m\), with \(\Sigma_r=\Sigma_p+\Sigma_\eta\). All expressions restrict to valid action coordinates, and the interpolation noise \(\epsilon_a\) is independent of both \(y_a^0\) and \(\eta\).

\paragraph{Conditional flow field.}
Squared regression has the conditional expectation as its population minimizer. Using \(\epsilon_a-\actiontarget=(x_{\actiontime}^a-\actiontarget)/\actiontime\), the optimal action field under \(r\) is
\begin{equation}
    v_r^\star(x,\actiontime;z)
    =\mathbb E_r[\epsilon_a-\actiontarget\mid x_{\actiontime}^a=x,z]
    =\frac{x-\mathbb E_r[\actiontarget\mid x_{\actiontime}^a=x,z]}{\actiontime},
    \label{eq:population_action_field}
\end{equation}
for \(\actiontime>0\). The corresponding expression holds for \(p\). Thus, the field comparison depends on the posterior action means under the two laws.

\paragraph{Noise-dependent agreement.}
Write \(a=1-\actiontime\), \(w=x-am\), and \(C_s=a^2\Sigma_s+\actiontime^2I\) for \(s\in\{p,r\}\). Gaussian conditioning gives the posterior action mean
\begin{equation}
    m_s(x,\actiontime;z)=m+a\Sigma_s C_s^{-1}w.
    \label{eq:conditional_action_mean}
\end{equation}
Substituting into Equation~\eqref{eq:population_action_field} and applying the inverse-matrix identity gives
\begin{equation}
\begin{aligned}
    v_r^\star-v_p^\star
    &=\frac{m_p-m_r}{\actiontime}
      =-\actiontime a\,C_r^{-1}\Sigma_\eta C_p^{-1}w,\\
    \|v_r^\star-v_p^\star\|_2
    &\leq\frac{1-\actiontime}{\actiontime^3}
      \|\Sigma_\eta\|_{\mathrm{op}}\|w\|_2.
\end{aligned}
    \label{eq:matched_mean_field_bound}
\end{equation}
The bound follows from \(C_s\succeq\actiontime^2I\), with \(\|\cdot\|_{\mathrm{op}}\) denoting the operator norm. For fixed \(x,z\), the field discrepancy vanishes as \(\actiontime\to1\), even though individual velocity targets retain annotation variation. This manuscript-specific bound gives a conditional-flow interpretation for our setting. Its high-noise contraction is consistent with ADP's noise-dependent use of suboptimal data~\citep{wei2026ambientpolicy} and motivates our choice to assign less reliable sources stronger action corruption. The weight \(q_d\) complements this allocation by controlling each source's contribution to the shared training objective.

\section{Masked objectives and teacher implementation}
\label{sec:loss_details}

\paragraph{Action validity.}
All manipulation examples have action labels. Let \(M_{a,i,kj}\in\{0,1\}\) identify a valid action coordinate \(j\) at step \(k\) of example \(i\), excluding padding. Padded coordinates remain zero after corruption. The full action expressions underlying Equation~\eqref{eq:joint_interpolants} are
\begin{equation}
    x_{\actiontime,i}^{a}
    =M_{a,i}\odot\bigl[(1-\actiontime)y_{a,i}+\actiontime\epsilon_{a,i}\bigr],
    \qquad
    u_{a,i}^\star=M_{a,i}\odot(\epsilon_{a,i}-y_{a,i}).
\end{equation}
An action token is valid when at least one of its coordinates is annotated:
\(b_{i,k}=\mathbf{1}\{\sum_jM_{a,i,kj}>0\}\).
Tokens with \(b_{i,k}=0\) are padding and are excluded from action attention and representation alignment.

\paragraph{Loss reductions.}
The MSE notation in Equation~\eqref{eq:expected_losses} suppresses padding and batch reductions: flow losses average over valid coordinates, and representation loss averages over valid action tokens. With source reliability applied inside the action average, the sampled losses for a manipulation batch \(\mathcal B\) are
\begin{align}
    \widehat{\mathcal{L}}_v^{\mathrm{FM}}
    &=\frac{\sum_{i\in\mathcal{B}}(1-e_i)
      \|\widehat u_{v,i}-u_{v,i}^\star\|_F^2}
      {N_vd_v\max(1,\sum_{i\in\mathcal{B}}(1-e_i))},\\
    \widehat{\mathcal{L}}_a^{\mathrm{FM}}
    &=\frac{\sum_{i,k,j}q_{d_i}M_{a,i,kj}
      (\widehat u_{a,i,kj}-u_{a,i,kj}^\star)^2}
      {\sum_{i,k,j}M_{a,i,kj}},\\
    \widehat{\mathcal{L}}_{\mathrm{PFD}}
    &=\frac{\sum_{i,k}b_{i,k}\ell_{\mathrm{PFD},i,k}}
      {\sum_{i,k}b_{i,k}}.
    \label{eq:masked_losses}
\end{align}
Here \(e_i\) marks boundary examples. Vision loss averages only over ordinary examples and vanishes for an all-boundary batch. The \(\mathcal{L}\)'s in Equation~\eqref{eq:full_objective} average over batches, Gaussian noise, and corruption times. Direct action loss normalizes by valid coordinates, retaining the effect of \(q_d\) in single-source batches. Distillation averages over valid tokens without reliability weighting, including boundary examples. The autoregressive cross-entropy \(\mathcal L_{\mathrm{AR}}\) uses separately interleaved vision--language batches.

\paragraph{Teacher construction.}
The EMA covers the coupled vision--action module and its conditioning projections, excluding the VLM. The teacher reuses detached contextual, current visual, and state features, runs without gradients, and maintains FP32 parameters. For each example, we reconstruct the action input with the student's original \(\epsilon_a\), keeping the input consistent.

The action projection \(P_\phi\) is a trainable MLP. Cosine similarity in Equation~\eqref{eq:pfd} is evaluated with norm stabilization,
\begin{equation}
    \cos_\varepsilon(u,v)
    =\frac{u^\top v}{\max(\|u\|_2,\varepsilon)\max(\|v\|_2,\varepsilon)},
    \qquad \varepsilon=10^{-8}.
\end{equation}
The current configuration uses \(\mu=0.999\), \(p_{\mathrm{edge}}=0.02\), student layer \(\ell_s=5\), teacher layer \(\ell_t=11\), and a projection hidden width of \(3072\). Stage one uses synchronized times \(\visiontime=\actiontime\sim\operatorname{Beta}(1.5,1)\), unit action-loss weights, and no boundary examples or teacher. Its coupled-expert weights initialize both the stage-two student and teacher.

\section{Model pretraining details}
\label{sec:implementation}

\subsection{Data and quality filtering}
\label{sec:pretraining_data_details}

Table~\ref{tab:pretraining_data} lists the manipulation and VLM sources. S1 combines real and simulated robot demonstrations with curated human trajectories. S2 retains this mixture and adds the larger internal egocentric collection. Human samples pair observed motion with reconstructed wrist/fingertip trajectories or retargeted hand commands. Task and subtask annotations form the language instruction, together with available control-mode information. The VLM data comprise LLaVA-OneVision and the RoboPoint and SAT subsets released through Molmo2-ER; separate batches from these sources provide \(\mathcal L_{\mathrm{AR}}\) in both stages.

\begin{table}[!h]
\caption{\textbf{Pretraining sources and data allocation.} Hours denote the retained manipulation data after filtering.}
\label{tab:pretraining_data}
\centering
\small
\setlength{\tabcolsep}{4pt}
\renewcommand{\arraystretch}{1.12}
\begin{tabular*}{\linewidth}{@{\extracolsep{\fill}}>{\raggedright\arraybackslash}p{0.30\linewidth}r>{\raggedright\arraybackslash}p{0.57\linewidth}@{}}
\toprule
\textbf{Dataset} & \shortstack{\textbf{Retained}\\\textbf{hours}} & \textbf{Supervision used in this work} \\
\midrule
\multicolumn{3}{@{}l}{\textit{Stage 1: Higher-quality initialization}} \\[2pt]
AgiBot World (2025)\textsuperscript{a} & 2,392 & EEF, arm joints, gripper, head/waist, selected base commands \\
EgoDex\textsuperscript{b} & 715 & Wrist poses, fingertips, and retargeted gripper/hand targets \\
InternData-A1\textsuperscript{c} & 7,330 & Simulated single-/dual-arm EEF, joint, and gripper targets \\
T-Rex\textsuperscript{d} & 50 & Bimanual and dexterous-hand joints \\
\midrule
\multicolumn{3}{@{}l}{\textit{Stage 2: Human-video expansion}} \\[2pt]
All Stage 1 sources & 10,487 & Same action annotations as S1 \\
Internal Ego-Centric Dataset & 20,000 & Bimanual EEF and fingertip targets \\
\midrule
\multicolumn{3}{@{}l}{\textit{VLM data: Shared across S1 and S2}} \\[2pt]
LLaVA-OneVision\textsuperscript{e} & --- & Visual question answering and instruction following \\
Molmo2-ER-RoboPoint\textsuperscript{f} & --- & Spatial affordance prediction and visual grounding \\
Molmo2-ER-SAT\textsuperscript{g} & --- & Spatial reasoning question answering \\
\bottomrule
\end{tabular*}
\par\vspace{3pt}
\begin{minipage}{\linewidth}
\footnotesize\raggedright
\textsuperscript{a}\citet{agibot2025};
\textsuperscript{b}\citet{hoque2025egodex};
\textsuperscript{c}\citet{tian2025interndata};
\textsuperscript{d}\citet{niu2026trex}.
\par
\textsuperscript{e}\citet{li2024llavaonevision};
\textsuperscript{f}\citet{yuan2024robopoint};
\textsuperscript{g}\citet{ray2025sat}.
\par\smallskip
Hours and selected action fields are outputs of our filtering and preprocessing rather than totals or schemas reported by the cited sources; unavailable coordinates are masked. VLM data supervise autoregressive text prediction through \(\mathcal L_{\mathrm{AR}}\); hours are not applicable. The Molmo2-ER row names refer to our selected re-releases of RoboPoint and SAT subsets, while the citations establish only the original datasets.
\end{minipage}
\end{table}

\begin{figure}[h]
    \centering
    \begingroup
    \input{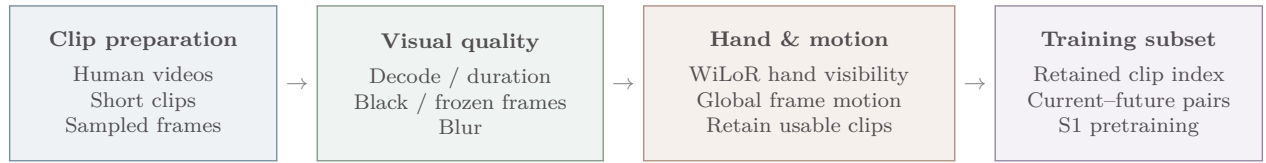}
    \color{ffInk}
    \footnotesize
    \setlength{\fboxsep}{4pt}
    \setlength{\fboxrule}{0.5pt}
    \fcolorbox{ffBlue}{ffBluebg}{%
        \begin{minipage}[c][1.75cm][c]{0.195\linewidth}
        \centering\textbf{Clip preparation}\\[4pt]
        Human videos\\Short clips\\Sampled frames
        \end{minipage}}\hfill
    \textcolor{ffMuted}{\(\rightarrow\)}\hfill
    \fcolorbox{ffSage}{ffJoint}{%
        \begin{minipage}[c][1.75cm][c]{0.21\linewidth}
        \centering\textbf{Visual quality}\\[4pt]
        Decode / duration\\Black / frozen frames\\Blur
        \end{minipage}}\hfill
    \textcolor{ffMuted}{\(\rightarrow\)}\hfill
    \fcolorbox{ffClay}{ffNoise}{%
        \begin{minipage}[c][1.75cm][c]{0.23\linewidth}
        \centering\textbf{Hand \& motion}\\[4pt]
        WiLoR hand visibility\\Global frame motion\\Retain usable clips
        \end{minipage}}\hfill
    \textcolor{ffMuted}{\(\rightarrow\)}\hfill
    \fcolorbox{ffLav}{ffLavbg}{%
        \begin{minipage}[c][1.75cm][c]{0.195\linewidth}
        \centering\textbf{Training subset}\\[4pt]
        Retained clip index\\Current--future pairs\\S1 pretraining
        \end{minipage}}%
    \endgroup
    \caption{\textbf{Human-video filtering for S1.} Clips that pass visual-quality, hand-visibility, and global-motion checks define the retained time intervals from which training pairs are constructed.}
    \label{fig:data_filtering}
\end{figure}

We screen the human videos used for S1 with a lightweight clip-level pipeline (Figure~\ref{fig:data_filtering}). Videos are divided into short clips and sparsely sampled to reject decoding failures, insufficient duration, excessive black or frozen frames, and severe blur. WiLoR hand detections measure the fraction of frames with visible hands, while frame-to-frame motion estimates identify excessive global movement. We retain clips with sufficient hand visibility and acceptable visual quality and motion, and export their time intervals as a dataset index. Training samples use current and future frames from these retained regions. This screening improves visual usability while leaving pose-estimation and retargeting noise to be handled during training.

\begin{figure}[tbp]
    \centering
    \includegraphics[width=\linewidth]{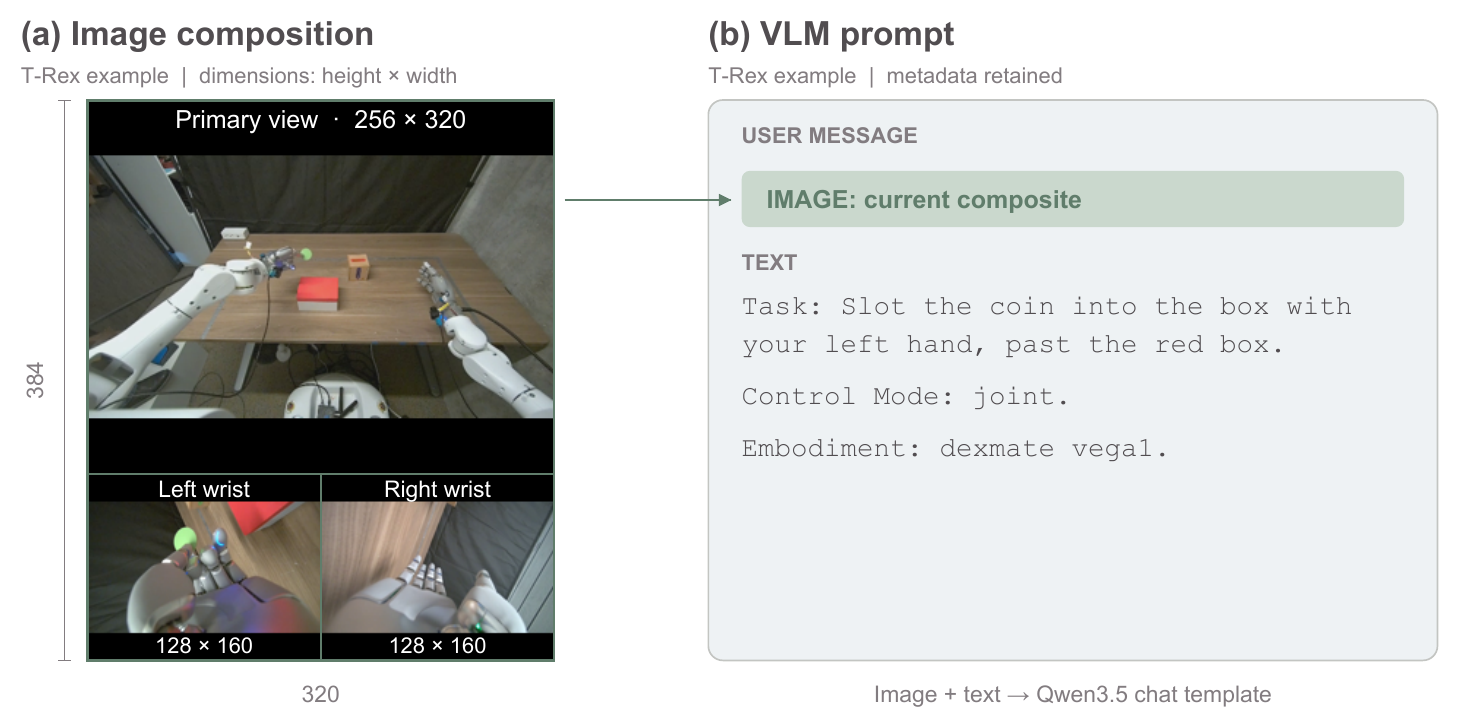}
    \caption{\textbf{Shared input format.} Input format across data sources, illustrated with a T-Rex example. Left: the primary view occupies the upper tile, with left and right wrist views below; each view is resized with its aspect ratio preserved and centered on black padding. Dimensions are height $\times$ width. Right: the VLM user message contains the composite image followed by task instructions and available metadata. Image labels and borders are explanatory overlays; prompt line breaks are added for readability.}
    \label{fig:input_format}
\end{figure}

Available action annotations are prepared before pretraining: for EgoDex, we construct wrist, fingertip, and retargeted hand targets, with the dexterous retargeting procedure detailed in Appendix~\ref{sec:retargeting}; robot sources provide EEF, joint, and gripper targets as applicable. Each source populates its corresponding canonical coordinates throughout training.

\subsection{Model architecture}
\label{sec:pretraining_architecture}

In our architecture, a 4B checkpoint from the Qwen3.5 family~\citep{qwen2026qwen35} provides visual--language context to our 16-block coupled vision--action module with hidden width 1,536 and 24 attention heads. The two streams have separate projections and feed-forward networks, exchange information through joint attention, and use modality-specific flow-time conditioning.

Images are resized or composed from available camera views at \(384\times320\) resolution. We use a frozen DINOv3 ViT-B/16 encoder~\citep{simeoni2025dinov3}; in our preprocessing, its 768-dimensional patch features are pooled to 120 tokens per image. The vision expert predicts future features and the action expert predicts 30-step action chunks. Conditioning uses the current observation, instruction, state, and source/control metadata, without observation history. The teacher follows Appendix~\ref{sec:loss_details}.

\subsection{Input format}
\label{sec:input_format}

We use a shared input format across data sources, illustrated in Figure~\ref{fig:input_format} with a T-Rex example. A \(256\times320\) primary-view tile (e.g., a head or egocentric view) sits above two \(128\times160\) wrist-view tiles, yielding a \(384\times320\) composite (height \(\times\) width). Views preserve their aspect ratios through centered black padding; unavailable wrist tiles remain black. The current composite precedes the instruction text in a user message processed by the Qwen3.5 chat template. The prompt combines task/subtask instructions with available control-mode, embodiment, and reference-frame metadata; the fields and their values depend on the source. Numeric state values enter the coupled module through a separate state encoder.

\subsection{Canonical action space}
\label{sec:canonical_action}

All sources share a 130-dimensional interface for normalized continuous actions. Table~\ref{tab:canonical_action} lists the used coordinates; unavailable coordinates and padded timesteps are masked. The current state uses the same layout with its own validity mask.

\begin{table}[htbp]
\caption{\textbf{Canonical action space.} Used coordinates in the 130-dimensional canonical action space. Indices are zero-based and ranges are inclusive. Availability is source-specific; unused slots and unused capacity within a slot are omitted.}
\label{tab:canonical_action}
\centering
\small
\setlength{\tabcolsep}{4pt}
\renewcommand{\arraystretch}{1.12}
\begin{tabular*}{\linewidth}{@{\extracolsep{\fill}}>{\raggedright\arraybackslash}p{0.22\linewidth}>{\raggedright\arraybackslash}p{0.22\linewidth}>{\raggedright\arraybackslash}p{0.48\linewidth}@{}}
\toprule
\textbf{Field} & \textbf{Indices} & \textbf{Target convention} \\
\midrule
Arm joints & L: 0--6; R: 29--35 & Up to 7 absolute joint positions per arm \\
EEF motion & L: 7--12; R: 36--41 & 3D translation and 3D rotation vector, relative to the current pose in camera coordinates \\
Gripper & L: 13; R: 42 & One opening/control scalar per gripper \\
Fingertips & L: 14--28; R: 43--57 & Five 3D positions per hand: thumb, index, middle, ring, little finger \\
Base velocity & 58, 63 & Forward velocity and yaw rate for selected AgiBot tasks \\
Head / waist & 64--65 / 68--69 & Two absolute coordinates each, in source order \\
Dexterous joints & L: 80--101; R: 105--126 & 22 absolute Sharpa joint angles per hand; shared order for human retargeting and T-Rex \\
\bottomrule
\end{tabular*}
\end{table}

EEF targets express motion relative to the current end-effector state in the current observation-camera frame. Let \({}^{A}T_B\) map coordinates from frame \(B\) to frame \(A\), and let \(C_t\) be the camera frame at the current observation. We convert camera trajectories to a shared camera-to-world convention \({}^{W}T_{C_s}\). For an EEF target at time \(s\) stored in its contemporaneous camera frame, camera-motion compensation transports the target into \(C_t\):
\begin{equation}
    \bar T_s
    =\bigl({}^{W}T_{C_t}\bigr)^{-1}
      {}^{W}T_{C_s}\,{}^{C_s}T_{E_s}
    =\begin{bmatrix}\bar R_s&\bar p_s\\0&1\end{bmatrix}.
    \label{eq:camera_motion_compensation}
\end{equation}
For robot targets already expressed in a common world or robot-base frame \(W\), we instead use \(\bar T_s=({}^{W}T_{C_t})^{-1}{}^{W}T_{E_s}\). Let \((R_t^{\mathrm{ref}},p_t^{\mathrm{ref}})\) denote the current EEF state in \(C_t\). The six-dimensional motion target is
\begin{equation}
    \Delta p_{t,s}=\bar p_s-p_t^{\mathrm{ref}},
    \qquad
    \Delta r_{t,s}=\operatorname{Log}\!\left(
        \bar R_s(R_t^{\mathrm{ref}})^{\top}
    \right)^{\vee},
    \label{eq:camera_eef_target}
\end{equation}
where \(\operatorname{Log}(\cdot)^{\vee}\) yields an axis--angle rotation vector. All targets in a chunk share the current EEF reference and camera axes. With consistent pose estimates, this removes apparent EEF motion caused solely by camera movement; for a static camera, the camera-to-camera transform in Equation~\eqref{eq:camera_motion_compensation} is the identity. Future camera poses are used only to construct training labels.

Joint, gripper, and fingertip targets retain their absolute conventions before normalization. We align images and actions in time, construct the EEF increments in physical units, and then normalize actions per dataset using percentile or bounded-range statistics, mapping valid coordinates to \([-1,1]\).

\subsection{Tracking and retargeting errors}
\label{sec:egocentric_action_errors}

Egocentric action labels depend on both camera tracking and hand reconstruction. For monocular video, moving hands and objects can interfere with camera-motion estimation, and recovering metric scale can introduce additional errors; camera-pose errors then propagate into hand trajectories expressed in a common frame~\citep{zhang2025hawor}. Hand tracking also degrades under occlusion and rapid motion. Calibrated multi-camera capture, as in EgoDex, provides multi-view geometric information, but EgoDex still reports imperfect hand annotations under heavy occlusion and rapid motion~\citep{hoque2025egodex}. The resulting annotation noise therefore depends on the capture and annotation pipeline.

\begin{figure}[t]
    \centering
    \includegraphics[width=\linewidth]{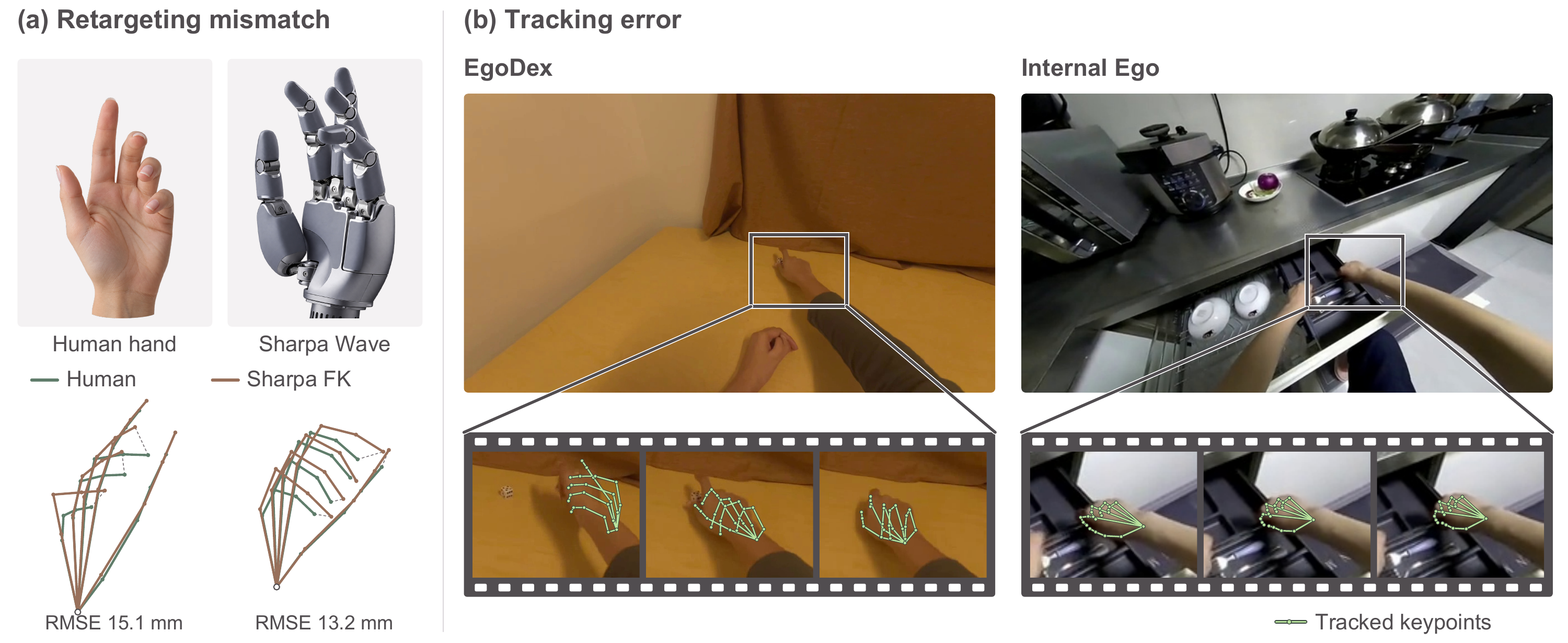}
\caption{\textbf{Retargeting mismatch and tracking errors.} \textbf{(a)} Human--robot morphological differences lead to residual geometric mismatch when retargeting human motions to Sharpa Wave. \textbf{(b)} Tracking annotations can deviate from visible hand poses in both EgoDex and Internal Ego. Both effects introduce noise into egocentric action supervision.}
    \label{fig:human_action_errors}
\end{figure}

Retargeting introduces a further geometric mismatch even when the human pose is accurately known. Human and robot hands differ in palm size, finger-length proportions, and joint structure, so the same wrist pose, fingertip positions, and finger orientations need not be jointly realizable by the robot~\citep{wang2024dexcap,xin2026retargeting}. Retargeting must compromise among these objectives; matching fingertips alone need not preserve the hand's contact geometry. This morphology-induced error is separate from upstream tracking errors and from the remaining visual and dynamical domain gap. Figure~\ref{fig:human_action_errors} illustrates retargeting mismatch and visible tracking errors.

\subsection{Retargeting implementation}
\label{sec:retargeting}

Starting from EgoDex's tracked wrist and fingertip data~\citep{hoque2025egodex}, we retarget each hand to 22 absolute Sharpa Wave joint angles. This paper-specific conversion follows the joint-space representation used in EgoScale~\citep{zheng2026egoscale}; our implementation uses the adaptive optimizer in AnyDexRetarget~\citep{qiu2025anydexretarget}, with its AVP-to-Sharpa configuration at revision \texttt{a76a0e1}. The two hands are processed independently, and optimizer and filter states are reset at every episode boundary.

\paragraph{Keypoints and coordinate alignment.}
From the 25 tracked transforms per hand, we select 21 positions: the wrist and four points along each finger, ordered thumb, index, middle, ring, and little finger. The four non-thumb metacarpal entries are omitted. AnyDexRetarget subtracts the wrist position, estimates a local frame from the wrist and the index and middle finger bases, and applies its handedness-specific axis convention. We then apply an additional rotation about the local \(x\)-axis of \(+7^\circ\) for the left hand and \(-7^\circ\) for the right hand. Finger fitting uses these positions and the distal segment directions derived from them. Tracked wrist poses are retained separately for the camera-frame EEF construction in Appendix~\ref{sec:canonical_action}.

\paragraph{Adaptive geometric objective.}
Let \(\theta\in\mathbb R^{22}\) be the robot joint angles, and let \(v_{f,k}(\theta)\) denote the forward-kinematic vector from the wrist to point \(k\) on finger \(f\). We fit the two intermediate points and fingertip, indexed by \(k\in\{\mathrm P,\mathrm D,\mathrm T\}\). Human targets \(\widehat v_{f,k}\) are constructed by accumulating scaled finger segments from the wrist; the wrist-to-finger-base vector is left unscaled. In thumb-to-little-finger order, the three successive segment multipliers are
\begin{equation}
    S=\begin{bmatrix}
    1.251&1.259&1.214\\
    1.108&1.148&1.160\\
    1.101&1.115&1.128\\
    1.117&1.141&1.155\\
    1.212&1.290&1.305
    \end{bmatrix}.
    \label{eq:retarget_scaling}
\end{equation}
These are the fixed scales in the upstream AVP-to-Sharpa configuration. Let \(d_f\) be the human thumb-to-fingertip distance in centimeters. For each non-thumb finger, the blend weight is
\begin{equation}
    \beta_f=\operatorname{clip}\!\left(\frac{4-d_f}{2},0,0.7\right),
    \qquad \beta_{\mathrm{th}}=\max_{f\ne\mathrm{th}}\beta_f.
    \label{eq:retarget_pinch}
\end{equation}
Let \(u_f(\theta)\) and \(\widehat u_f\) be unit distal-segment directions. Using the Huber penalty \(h_\delta(r)=r^2/2\) for \(r\leq\delta\) and \(h_\delta(r)=\delta(r-\delta/2)\) otherwise, define
\begin{align}
    L_f^{\mathrm{hand}}(\theta)
    &=\frac{1}{3}\sum_{k\in\{\mathrm P,\mathrm D,\mathrm T\}}
      h_2\!\left(\|v_{f,k}(\theta)-\widehat v_{f,k}\|_2\right),\\
    L_f^{\mathrm{tip}}(\theta)
    &=5h_2\!\left(\|v_{f,\mathrm T}(\theta)-\widehat v_{f,\mathrm T}\|_2\right)
      +h_{0.5}\!\left(\|u_f(\theta)-\widehat u_f\|_2\right).
\end{align}
All position residuals are evaluated in centimeters; direction residuals are dimensionless. With \(f^\star=\arg\max_{f\ne\mathrm{th}}\beta_f\), each frame solves
\begin{equation}
\begin{aligned}
    \theta_t^\star=\arg\min_{\theta_{\min}\leq\theta\leq\theta_{\max}}\;&
      \sum_f\bigl[(1-\beta_f)L_f^{\mathrm{hand}}(\theta)
                    +\beta_f L_f^{\mathrm{tip}}(\theta)\bigr]\\
      &+6\beta_{f^\star}\|v_{\mathrm{th},\mathrm T}(\theta)-v_{f^\star,\mathrm T}(\theta)\|_2^2
       +0.04\|\theta-\theta_{\mathrm{prev}}^\star\|_2^2.
\end{aligned}
    \label{eq:retarget_objective}
\end{equation}
The bounds come from the robot URDF. The contact term brings the thumb toward the most active pinch partner, while the cap \(\beta_f\leq0.7\) retains a posture penalty during pinching. The configuration uses \texttt{pinch\_scaling}=1, so the tip target remains the segment-scaled target above. We use NLopt SLSQP with analytical gradients, at most 50 objective evaluations per frame, and absolute objective tolerance \(10^{-4}\). The previous optimizer solution supplies both the initialization and temporal reference; the first solve starts from the neutral pose and omits the temporal term.

\paragraph{Temporal filtering and invalid frames.}
Accepted solutions pass through an exponential filter with new-sample weight \(0.2\). Its output is then clipped to a maximum change of \(12/30=0.4\) radians per joint per frame at 30\,Hz and to the URDF joint limits. Before optimization, we check that all selected transforms are finite and geometrically valid. Wrist-to-finger-base distances must lie in \((0.005,0.20)\)\,m, successive finger segments in \((0.002,0.10)\)\,m, wrist-to-middle-tip distance in \((0.06,0.35)\)\,m, and index-to-little-finger base distance in \((0.02,0.18)\)\,m. Consecutive wrist displacements must be below \(0.75\)\,m. When confidence annotations are present, all scores must be finite, their median must be at least \(0.05\), and wrist confidence must be at least \(0.10\). Invalid tracking, a negative solver status, or non-finite joint output causes the exported joints to hold their previous values, using the neutral pose if no preceding output exists.

\paragraph{Action alignment and diagnostics.}
For a source sequence with \(T\) frames, the training export contains \(T-1\) rows: row \(t\) stores the filtered hand state at \(t\) and the absolute joint target at \(t+1\). It preserves the separately constructed wrist targets. The 22 joint values follow the URDF joint-name order and populate coordinates 80--101 and 105--126 in Table~\ref{tab:canonical_action}. We retain tracking-validity and solver-failure flags, fitting costs, pinch weights, and keypoint errors as diagnostics; these fields do not themselves define the training action-validity mask. The diagnostic RMSE compares 21 forward-kinematic points from the exported joints against the aligned, unscaled human points in meters, as illustrated in Figure~\ref{fig:human_action_errors}(a). It therefore measures residual geometric mismatch rather than the weighted objective in Equation~\eqref{eq:retarget_objective}. This kinematic fit enforces joint limits but does not impose object-contact or collision constraints.

\subsection{Training details}
\label{sec:pretraining_optimization}

Table~\ref{tab:pretraining_settings} summarizes the recipe. S1 trains for 100k updates with synchronized flow times and no privileged teacher. S2 trains for 200k updates on the expanded mixture with both proposed mechanisms enabled. Both stages use a global batch size of 2,048 and 5k learning-rate warmup steps, with cosine learning-rate decay and BF16 mixed precision. The two-stage pretraining uses 22k NVIDIA H200 GPU-hours in total. Vision--language co-training provides the auxiliary language objective.

Following the Kimi/Moonlight learning-rate scaling~\citep{liu2025muonscalable}, the orthogonalized Muon update for an \(m\times n\) matrix is scaled by \(0.2\sqrt{\max(m,n)}\) to approximately match AdamW's update RMS. The learning rates in Table~\ref{tab:pretraining_settings} are the base rates before this matrix-dependent scaling.

\begin{table}[htbp]
\caption{\textbf{Pretraining settings.} Settings for S1 initialization and S2 expansion. Global batch sizes count manipulation samples.}
\label{tab:pretraining_settings}
\centering
\small
\setlength{\tabcolsep}{4pt}
\renewcommand{\arraystretch}{1.12}
\begin{tabular*}{\linewidth}{@{\extracolsep{\fill}}lcc@{}}
\toprule
\textbf{Setting} & \textbf{S1} & \textbf{S2} \\
\midrule
Optimizer updates & 100k & 200k \\
Global batch size & 2,048 & 2,048 \\
Learning-rate warmup & 5k & 5k \\
Peak base LR: VLM / coupled module & \(10^{-5}\) / \(5\times10^{-5}\) & \(10^{-5}\) / \(5\times10^{-5}\) \\
Flow-time sampling & Synchronized Beta(1.5, 1) & Reliability-conditioned \\
Action reliability \(q_d\) & 1 for all sources & Robot: 1; human: 0.2 \\
Vision / action loss coefficient & 0.1 / 1 & 0.1 / 1 \\
Distillation / AR loss coefficient & 0 / 0.05 & 0.05 / 0.05 \\
\bottomrule
\end{tabular*}
\end{table}

\section{Additional experimental protocol}
\label{sec:additional_experiments}

\subsection{RoboCasa GR1 Tabletop}
\label{sec:robocasa_protocol}

\paragraph{Tasks and observations.}
We follow DIAL's 24-task protocol~\citep{chen2026dial}. The training trajectories come from the humanoid tabletop subset of PhysicalAI-Robotics-GR00T-X-Embodiment-Sim. PnP tasks require moving objects between supports or containers; articulated tasks additionally require closing the destination fixture after placement. DIAL uses language, a $224\times224$ egocentric image, and proprioception, with 16-step action chunks and a 47-dimensional learning representation that includes 18 end-effector-pose dimensions and 29 commanded joint dimensions. Our preprocessing instead forms the $384\times320$ composite used during pretraining, placing the head-camera view above two blank wrist-camera regions. The GR1 controller executes the same 29 joint commands: 14 for the arms, 12 for the hands, and 3 for the waist.

\paragraph{Post-training.}
We adapt the pretrained model jointly to all 24 tasks using the settings in Table~\ref{tab:robocasa_training}. Few-shot training uses one tenth of the full robot demonstrations. Both settings retain the privileged-future distillation objective from Section~\ref{sec:pfd}; privileged future observations are used only during training.

\begin{table}[htbp]
\centering
\caption{\textbf{RoboCasa post-training settings.} Settings for \methodname\ on RoboCasa GR1.}
\label{tab:robocasa_training}
\small
\setlength{\tabcolsep}{8pt}
\renewcommand{\arraystretch}{1.10}
\begin{tabular*}{0.86\linewidth}{@{\extracolsep{\fill}}lrr@{}}
\toprule
Setting & Few-shot & Full data \\
\midrule
Demonstrations per task & 100 & 1,000 \\
Total robot trajectories & 2,400 & 24,000 \\
Global batch size & 128 & 512 \\
Training steps & 70,000 & 200,000 \\
Feature distillation & Enabled & Enabled \\
Evaluation episodes per task & 50 & 50 \\
\bottomrule
\end{tabular*}
\end{table}

\begin{table}[t]
\caption{\textbf{Per-task results on RoboCasa.} Success rates (\%) of \methodname\ on RoboCasa GR1. Each task is evaluated over 50 episodes in each setting.}
\label{tab:robocasa_per_task}
\centering
\small
\setlength{\tabcolsep}{8pt}
\renewcommand{\arraystretch}{1.12}
\begin{tabular*}{0.9\linewidth}{@{\extracolsep{\fill}}lrr@{}}
\toprule
Task & Full data & Few-shot \\
\midrule
\multicolumn{3}{@{}l}{\textit{Pick-and-place (18 tasks)}} \\
Cutting board $\rightarrow$ Basket & 98.0 & 88.0 \\
Cutting board $\rightarrow$ Cardboard box & 62.0 & 52.0 \\
Cutting board $\rightarrow$ Pan & 90.0 & 84.0 \\
Cutting board $\rightarrow$ Pot & 90.0 & 90.0 \\
Cutting board $\rightarrow$ Tiered basket & 52.0 & 48.0 \\
Placemat $\rightarrow$ Basket & 88.0 & 78.0 \\
Placemat $\rightarrow$ Bowl & 66.0 & 52.0 \\
Placemat $\rightarrow$ Plate & 48.0 & 44.0 \\
Placemat $\rightarrow$ Tiered shelf & 44.0 & 28.0 \\
Plate $\rightarrow$ Bowl & 72.0 & 56.0 \\
Plate $\rightarrow$ Cardboard box & 54.0 & 42.0 \\
Plate $\rightarrow$ Pan & 50.0 & 48.0 \\
Plate $\rightarrow$ Plate & 90.0 & 80.0 \\
Tray $\rightarrow$ Cardboard box & 92.0 & 84.0 \\
Tray $\rightarrow$ Plate & 76.0 & 84.0 \\
Tray $\rightarrow$ Pot & 86.0 & 92.0 \\
Tray $\rightarrow$ Tiered basket & 86.0 & 70.0 \\
Tray $\rightarrow$ Tiered shelf & 50.0 & 46.0 \\
\addlinespace[2pt]
\textbf{Category average} & \textbf{71.9} & \textbf{64.8} \\
\midrule
\multicolumn{3}{@{}l}{\textit{Articulated objects (6 tasks)}} \\
Bottle $\rightarrow$ Cabinet + close & 92.0 & 80.0 \\
Can $\rightarrow$ Drawer + close & 96.0 & 84.0 \\
Cup $\rightarrow$ Drawer + close & 64.0 & 54.0 \\
Milk $\rightarrow$ Microwave + close & 70.0 & 70.0 \\
Potato $\rightarrow$ Microwave + close & 80.0 & 58.0 \\
Wine $\rightarrow$ Cabinet + close & 78.0 & 72.0 \\
\addlinespace[2pt]
\textbf{Category average} & \textbf{80.0} & \textbf{69.7} \\
\midrule
\textbf{Overall average} & \textbf{73.9} & \textbf{66.0} \\
\bottomrule
\end{tabular*}
\end{table}

\paragraph{Baseline variants and sources.}
The GR00T N1.6~\citep{nvidia2025gr00tn16} category scores, the Qwen2.5 variants, and the robot-only DIAL results are from Figures~7--8 of \citet{chen2026dial}. The FLARE row is DIAL's FLARE-style extension of the fine-tuned GR00T-Qwen2.5 baseline~\citep{chen2026dial,zheng2025flare}, rather than a different-backbone result. DIAL + EgoDex uses their additional human-data training and the in-distribution results in Figure~9. For the remaining category scores, we average the published per-task results: LDA-1B, Table~VI of \citet{lyu2026lda}; RLDX-1, Table~14 of \citet{kim2026rldx1}; and JoyAI-RA 0.1, Table~B of \citet{zhang2026joyaira}.

\paragraph{Evaluation configuration.}
We evaluate the EMA teacher checkpoint using 20 flow integration steps, generating 16 action steps and executing all 16 steps per prediction.

\subsection{Real-world dexterous manipulation}
\label{sec:real_world_protocol}

\begin{figure}[htbp]
\centering
\begingroup
\input{figures/morandi_colors}
\pgfmathsetmacro{\RWSetupUnit}{\linewidth/396pt}
\pgfmathsetlengthmacro{\RWSetupPhotoHeight}{190*\RWSetupUnit pt}
\resizebox{0.85\linewidth}{!}{%
\begin{tikzpicture}[x=\RWSetupUnit pt,y=\RWSetupUnit pt,
    every node/.style={inner sep=0pt,outer sep=0pt,text=ffInk},
    setup label/.style={align=left,font=\fontsize{7.5}{8.5}\selectfont,
        fill=white,inner sep=2pt,rounded corners=2pt},
    setup leader/.style={line width=0.7pt,line cap=round,line join=round}]
    \path[use as bounding box] (0,0) rectangle (396,234);
    \node[anchor=south west] at (8,22) {\includegraphics[
        height=\RWSetupPhotoHeight]{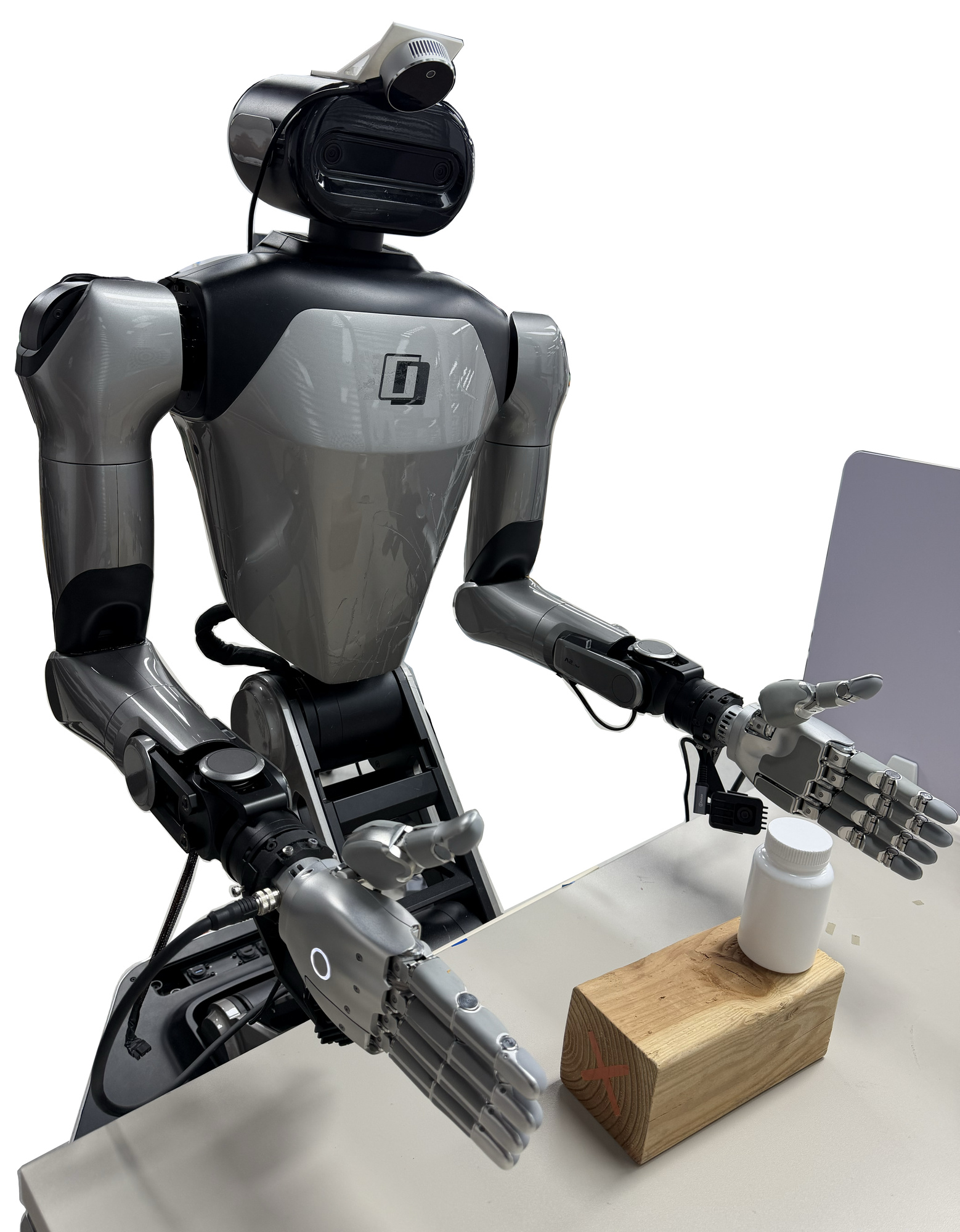}};
    \node[anchor=south west] at (218,22) {\includegraphics[
        height=\RWSetupPhotoHeight]{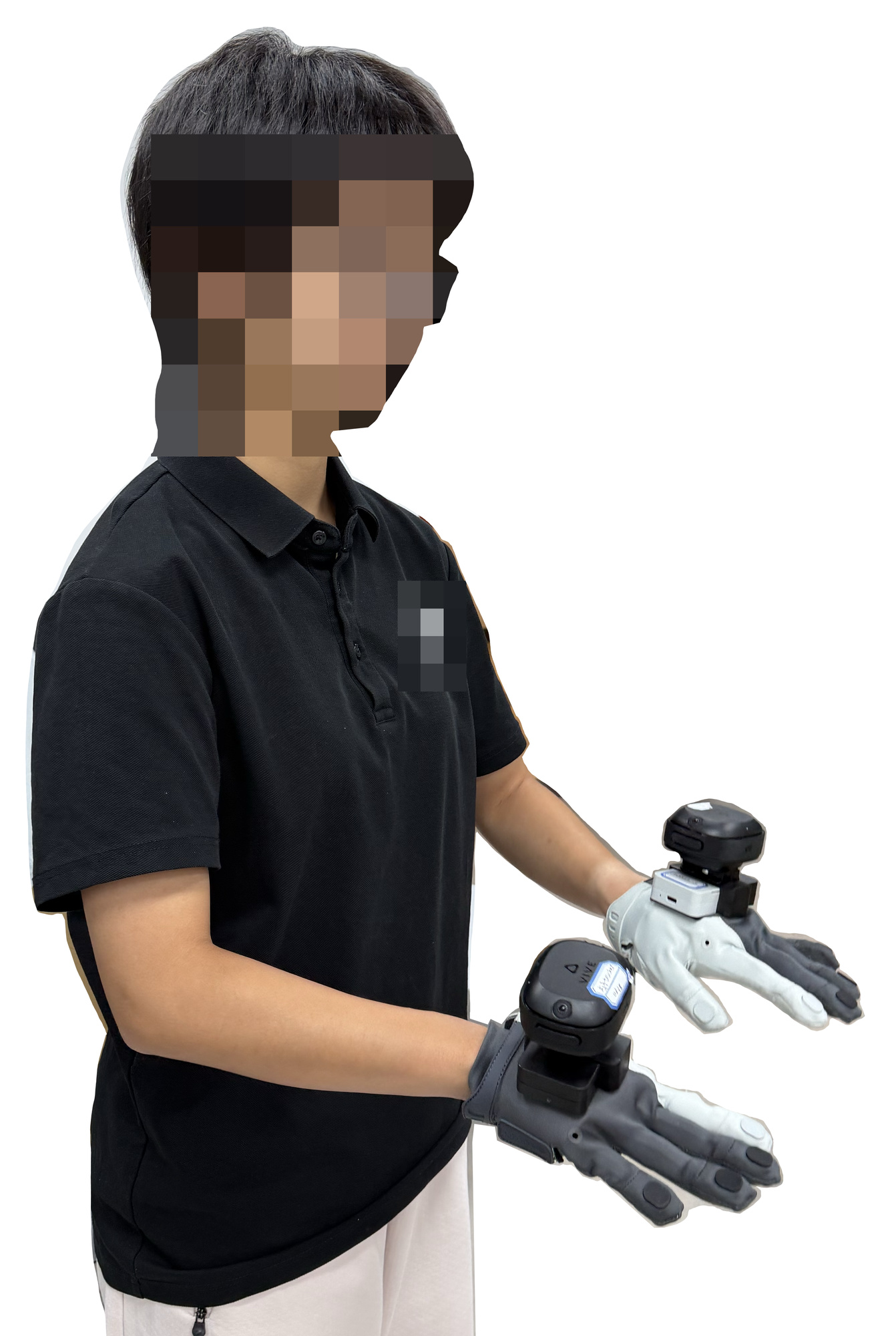}};
    \draw[ffGrid,line width=0.4pt] (198,20) -- (198,215);

    \node[anchor=west,font=\fontsize{9}{10}\selectfont\bfseries]
        at (3,228) {(a) Robot platform};
    \node[anchor=west,font=\fontsize{9}{10}\selectfont\bfseries]
        at (205,228) {(b) Teleoperation};

    \draw[setup leader,ffBlueDark] (140,197) -- (112,197) -- (73,198.5);
    \fill[ffBlueDark] (73,198.5) circle (1.6);
    \node[setup label,anchor=west] at (112,207)
        {Head camera\\\textbf{RealSense L515}};
    \draw[setup leader,ffBlueDark] (160,133) -- (160,106) -- (121,88);
    \fill[ffBlueDark] (121,88) circle (1.6);
    \node[setup label,anchor=center] at (158,145)
        {Wrist cameras\\\textbf{OAK-1-W}};
    \draw[setup leader,ffSageDark] (69,34) -- (69,48) -- (67,66);
    \fill[ffSageDark] (67,66) circle (1.6);
    \node[setup label,anchor=center] at (69,29) {\textbf{Sharpa Wave hands}};

    \draw[setup leader,ffClayDark] (365,140) -- (365,115) -- (320,94);
    \fill[ffClayDark] (320,94) circle (1.6);
    \node[setup label,anchor=center] at (361,153)
        {Back of hand\\\textbf{VIVE tracker}};
    \draw[setup leader,ffSageDark] (359,53) -- (334,53) -- (312,56);
    \fill[ffSageDark] (312,56) circle (1.6);
    \node[setup label,anchor=west] at (340,40)
        {Data gloves\\\textbf{MANUS}};

    \fill[ffBluebg,rounded corners=2pt] (0,0) rectangle (194,16);
    \fill[ffJoint,rounded corners=2pt] (202,0) rectangle (396,16);
    \node[font=\fontsize{8}{9}\selectfont\bfseries] at (97,8) {Galaxea R1 Pro};
    \node[font=\fontsize{8}{9}\selectfont\bfseries] at (299,8) {Human operator};
\end{tikzpicture}}\par
\setlength{\abovecaptionskip}{6pt}
\caption{\textbf{Real-world hardware and teleoperation setup.} (a) Galaxea R1 Pro with two Sharpa Wave hands, a RealSense L515 head camera, and OAK-1-W wrist cameras. (b) MANUS gloves with a VIVE tracker on the back of each hand. The operator's face and clothing logo are pixelated.}
\label{fig:real_world_setup}
\endgroup
\end{figure}

\paragraph{Hardware and recorded observations.}
Our real-world platform is a Galaxea R1 Pro equipped with two Sharpa Wave dexterous hands (Figure~\ref{fig:real_world_setup}). The teleoperation recordings include RGB observations from a head-mounted RealSense L515 and an OAK-1-W camera at each wrist. The converted training data store all three views at $1280\times720$ resolution and 10\,Hz. The recorded joint-state vector has 58 coordinates: seven joints for each arm and 22 joints for each hand. End-effector poses and hand-joint targets are also available in the recordings.

\paragraph{Teleoperation interface.}
The operator wears MANUS gloves, with a VIVE tracker mounted on the back of each hand (Figure~\ref{fig:real_world_setup}b).

\begin{figure}[!t]
\centering
\begingroup
\input{figures/morandi_colors}
\newcommand{\RWSequenceFrame}[2]{%
    \node[anchor=south west] at ({66.5*#1},14) {#2};%
}
\newcommand{\RWSequence}[2]{%
    \begingroup
    \pgfmathsetmacro{\RWSeqUnit}{\linewidth/396pt}%
    \pgfmathsetlengthmacro{\RWSeqPhotoWidth}{63.5*\RWSeqUnit pt}%
    \begin{tikzpicture}[x=\RWSeqUnit pt,y=\RWSeqUnit pt,
        every node/.style={inner sep=0pt,outer sep=0pt,text=ffInk}]
        \path[use as bounding box] (0,0) rectangle (396,49.72);
        #1%
        \fill[ffData,rounded corners=2.5pt] (0,0) rectangle (396,11);
        \node[font=\fontsize{8}{9}\selectfont\bfseries] at (198,5.5) {#2};
    \end{tikzpicture}%
    \endgroup
}
\RWSequence{%
    \RWSequenceFrame{0}{\includegraphics[width=\RWSeqPhotoWidth]{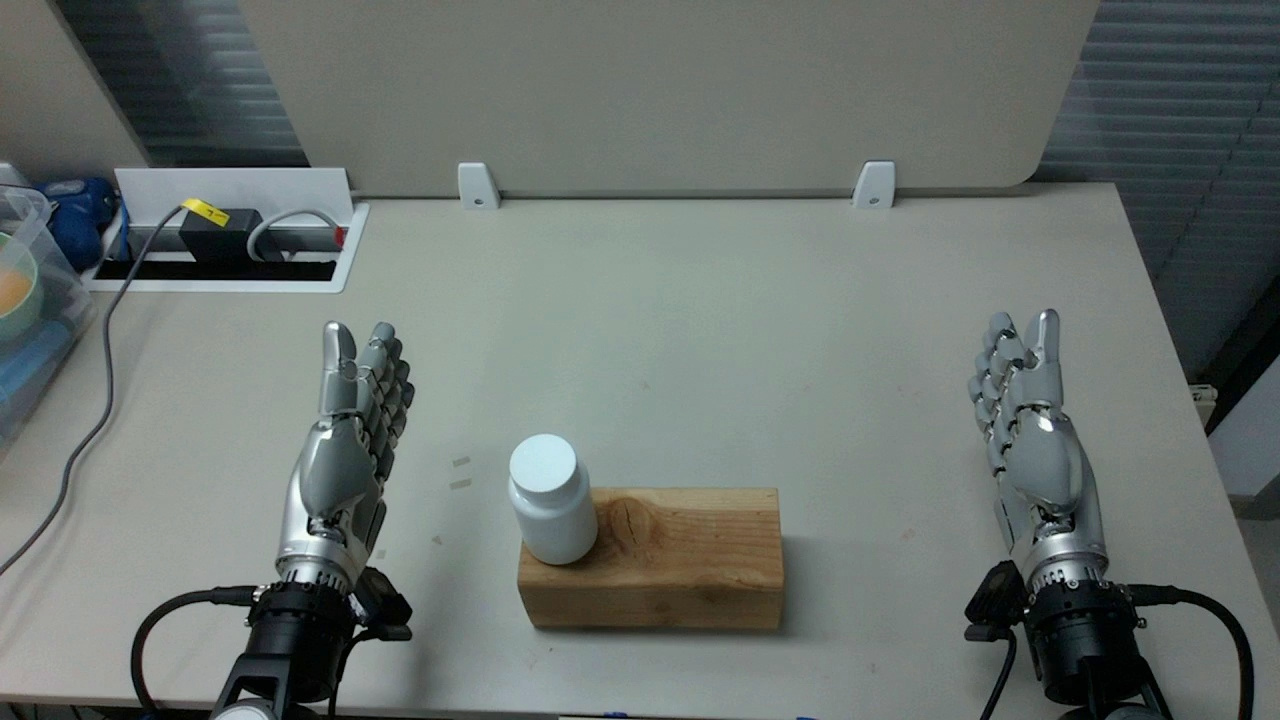}}%
    \RWSequenceFrame{1}{\includegraphics[width=\RWSeqPhotoWidth]{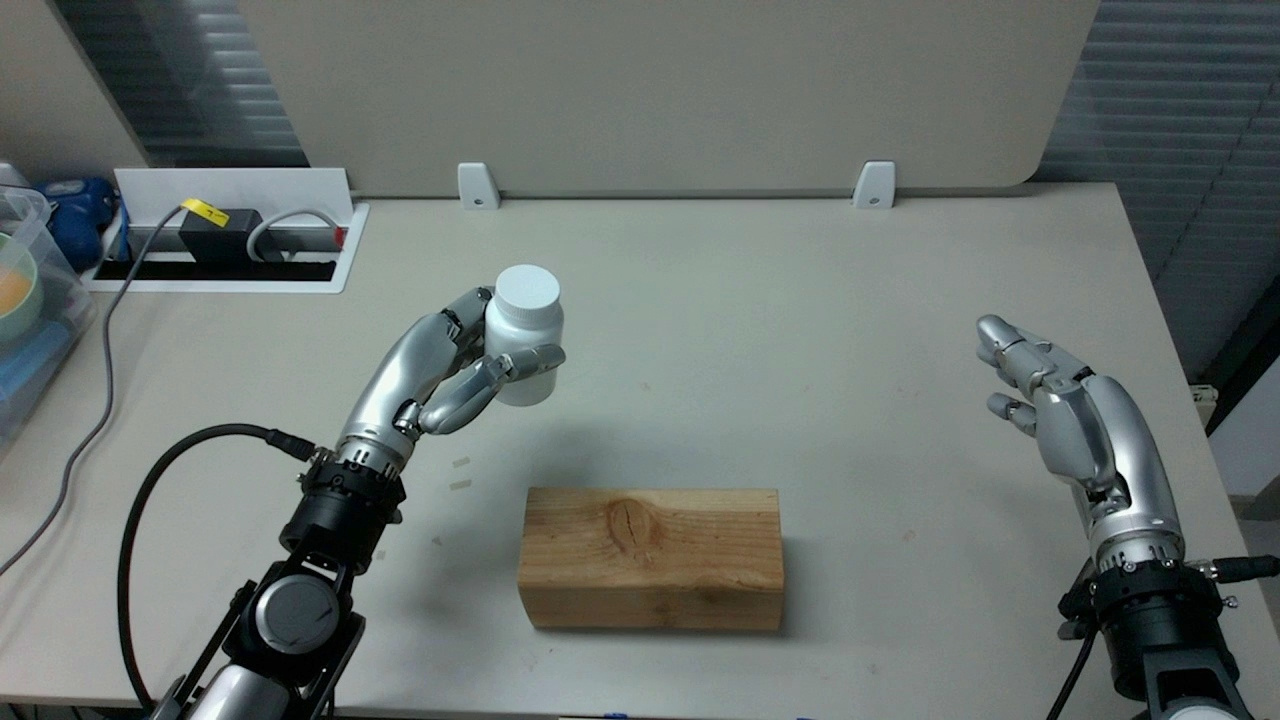}}%
    \RWSequenceFrame{2}{\includegraphics[width=\RWSeqPhotoWidth]{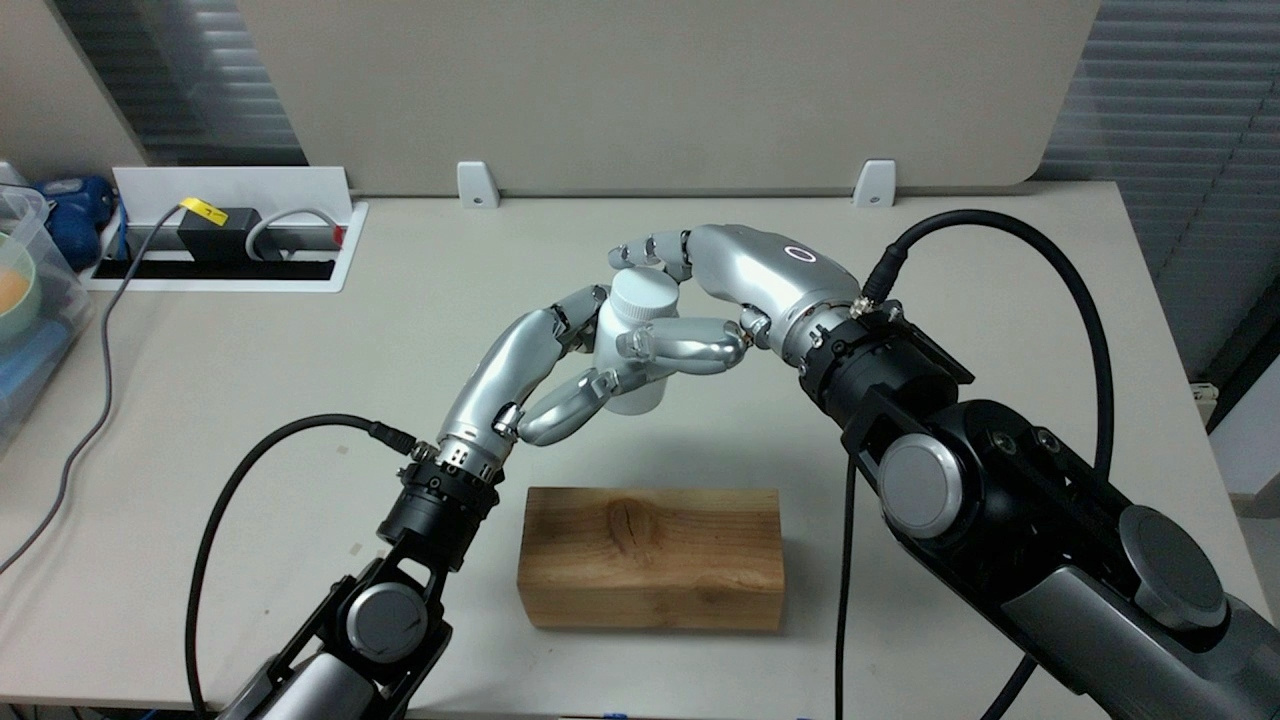}}%
    \RWSequenceFrame{3}{\includegraphics[width=\RWSeqPhotoWidth]{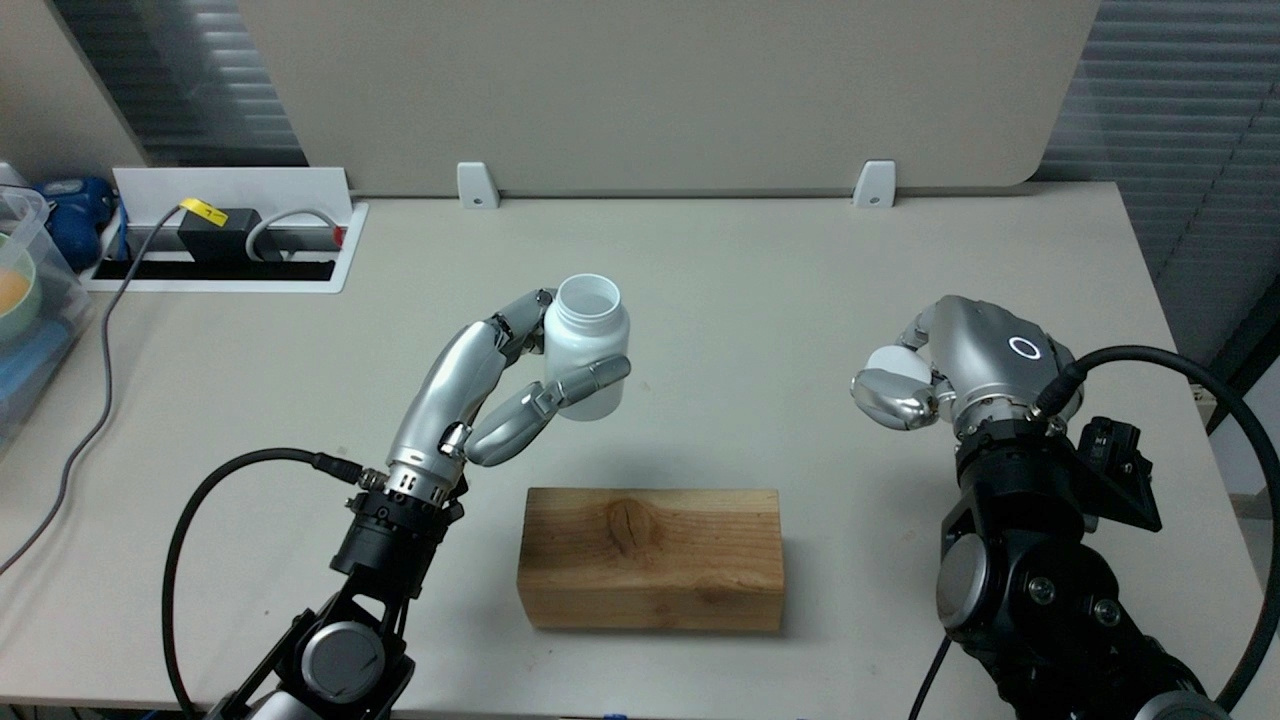}}%
    \RWSequenceFrame{4}{\includegraphics[width=\RWSeqPhotoWidth]{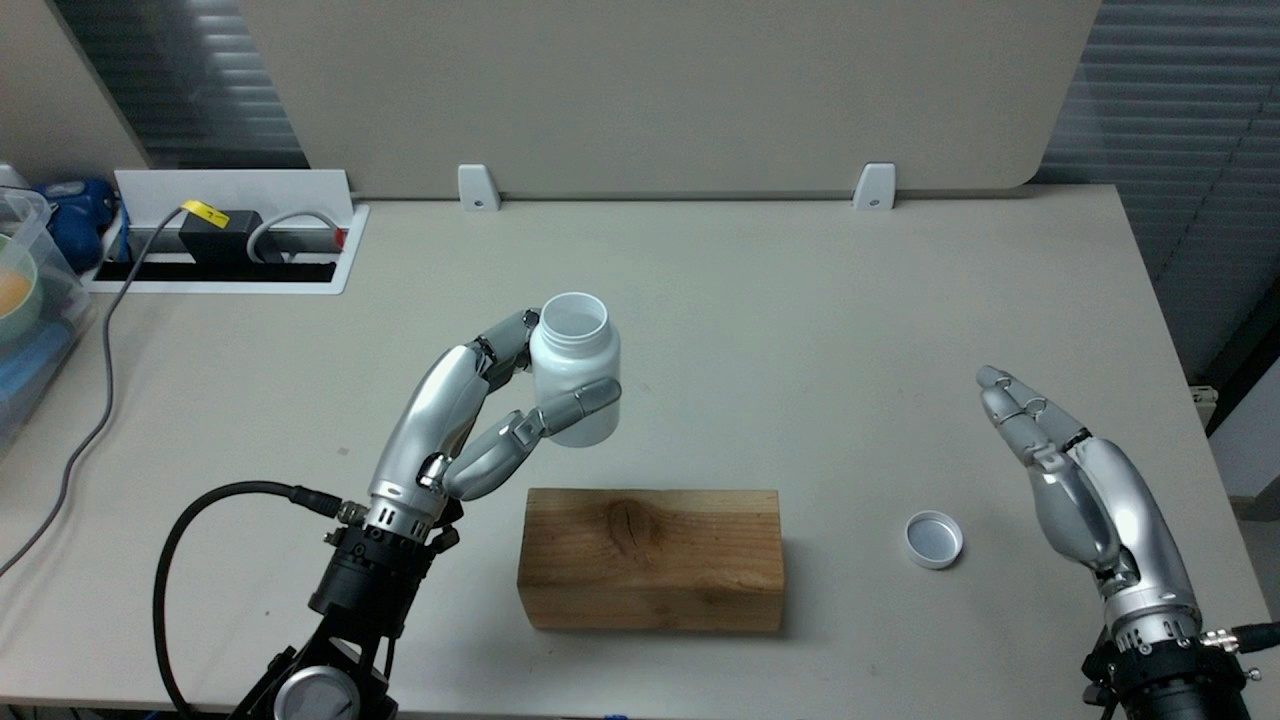}}%
    \RWSequenceFrame{5}{\includegraphics[width=\RWSeqPhotoWidth]{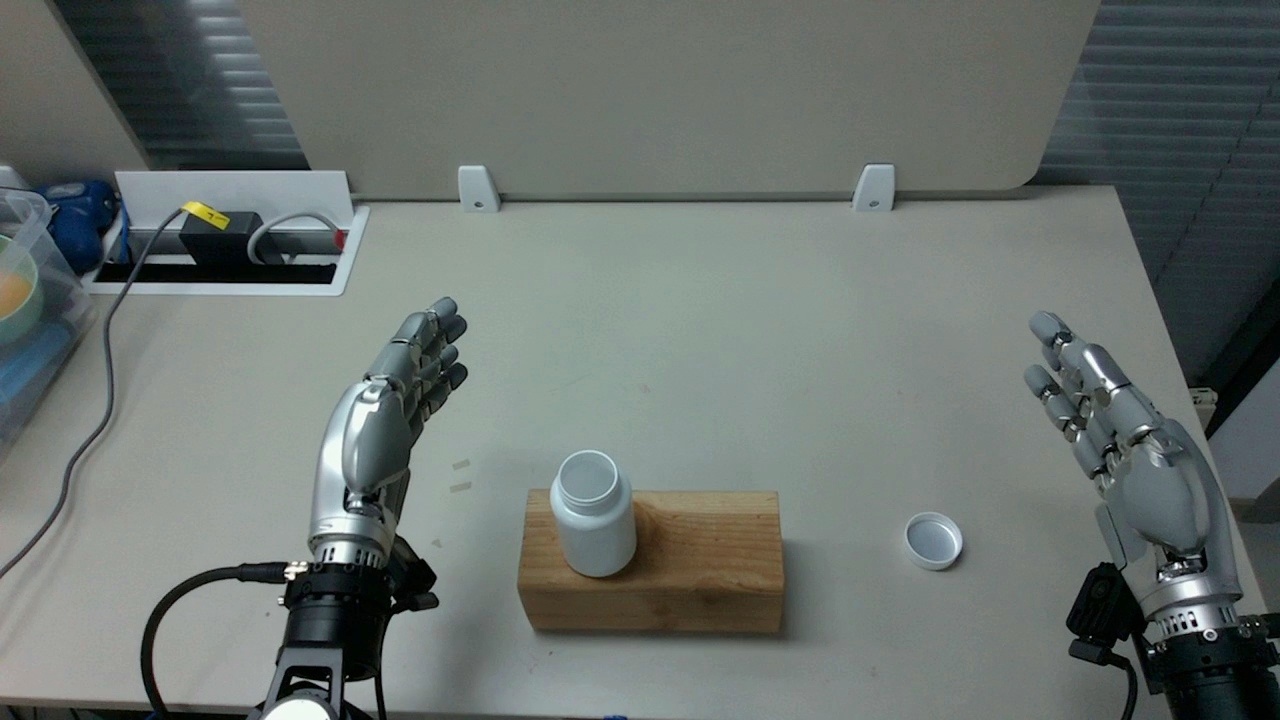}}%
}{(a) Bottle opening}\par\vspace{6pt}
\RWSequence{%
    \RWSequenceFrame{0}{\includegraphics[width=\RWSeqPhotoWidth]{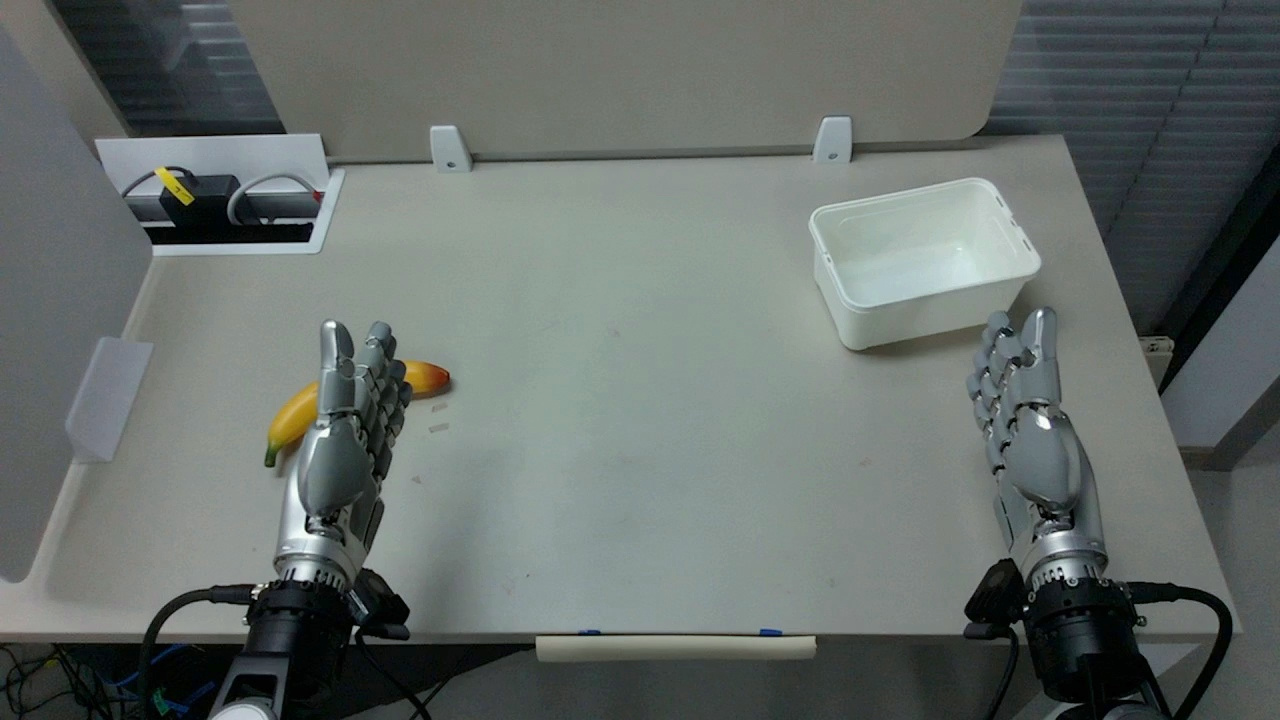}}%
    \RWSequenceFrame{1}{\includegraphics[width=\RWSeqPhotoWidth]{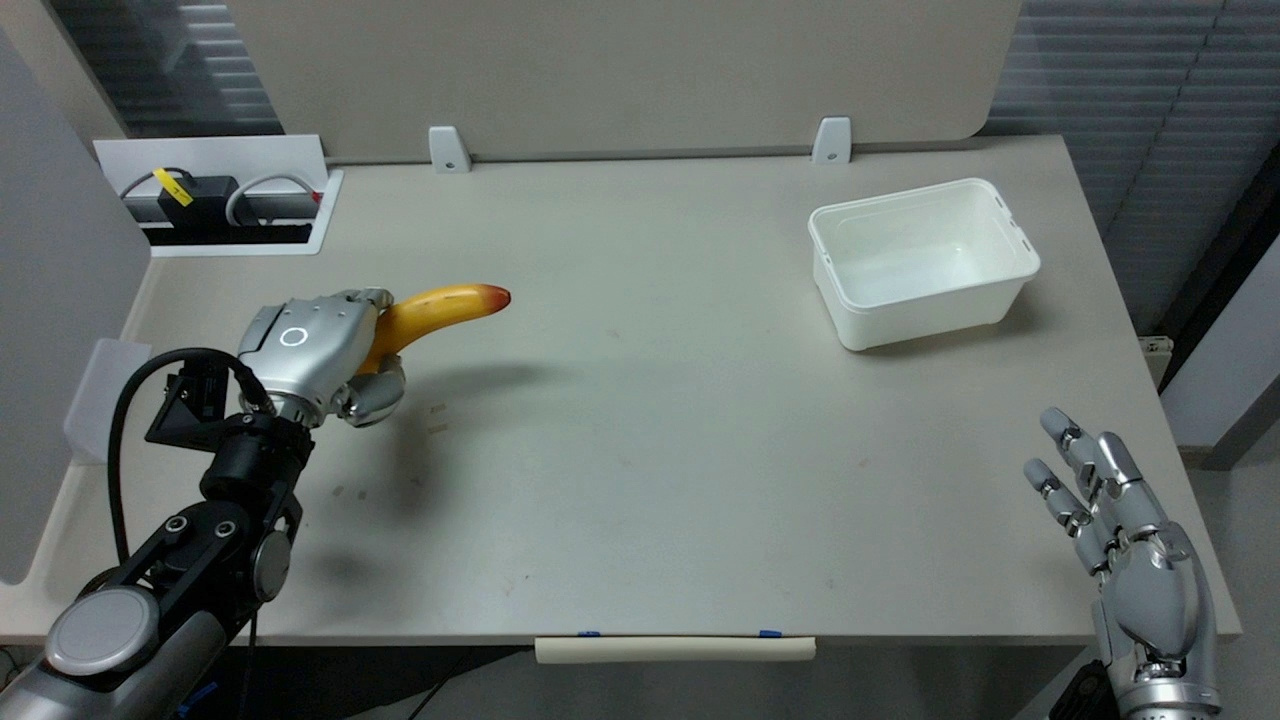}}%
    \RWSequenceFrame{2}{\includegraphics[width=\RWSeqPhotoWidth]{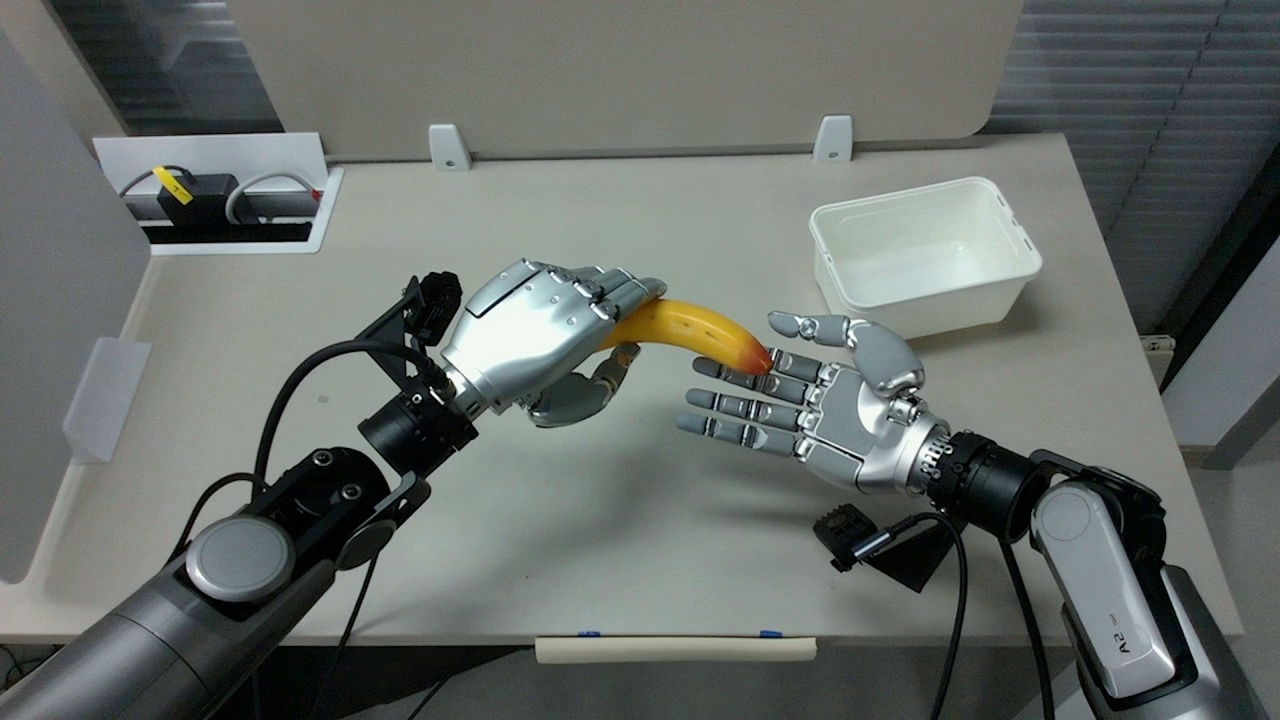}}%
    \RWSequenceFrame{3}{\includegraphics[width=\RWSeqPhotoWidth]{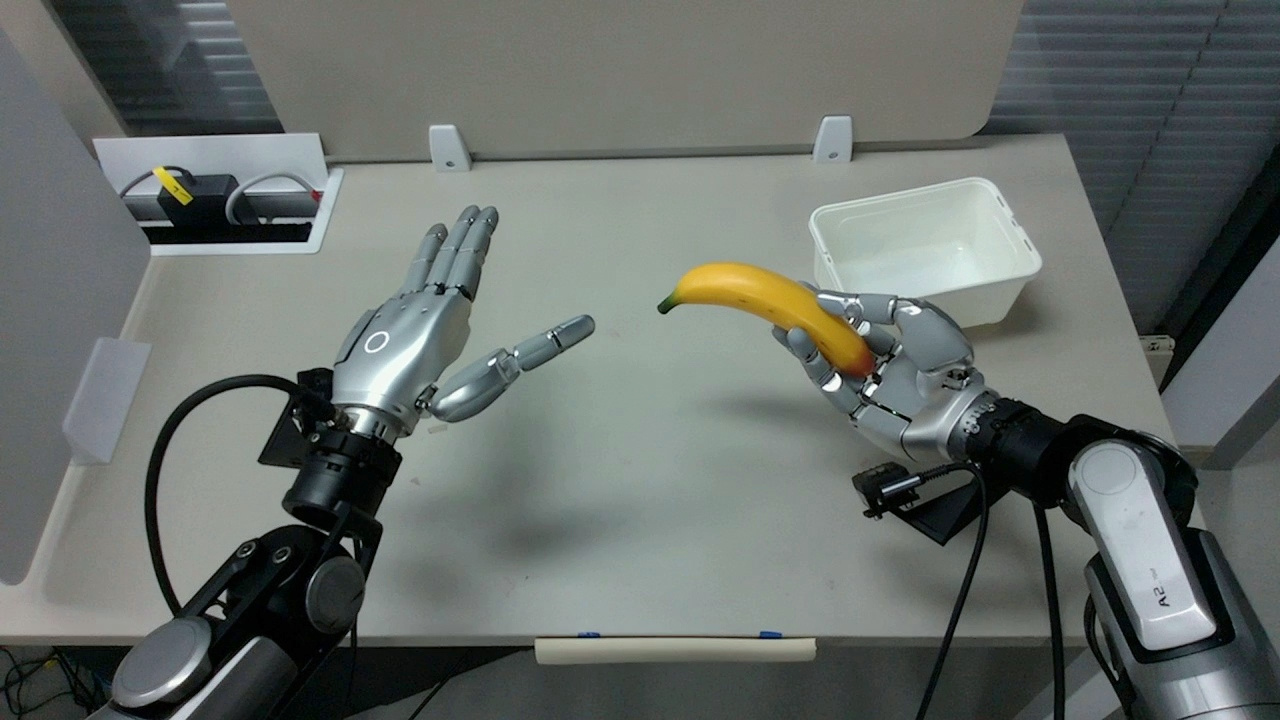}}%
    \RWSequenceFrame{4}{\includegraphics[width=\RWSeqPhotoWidth]{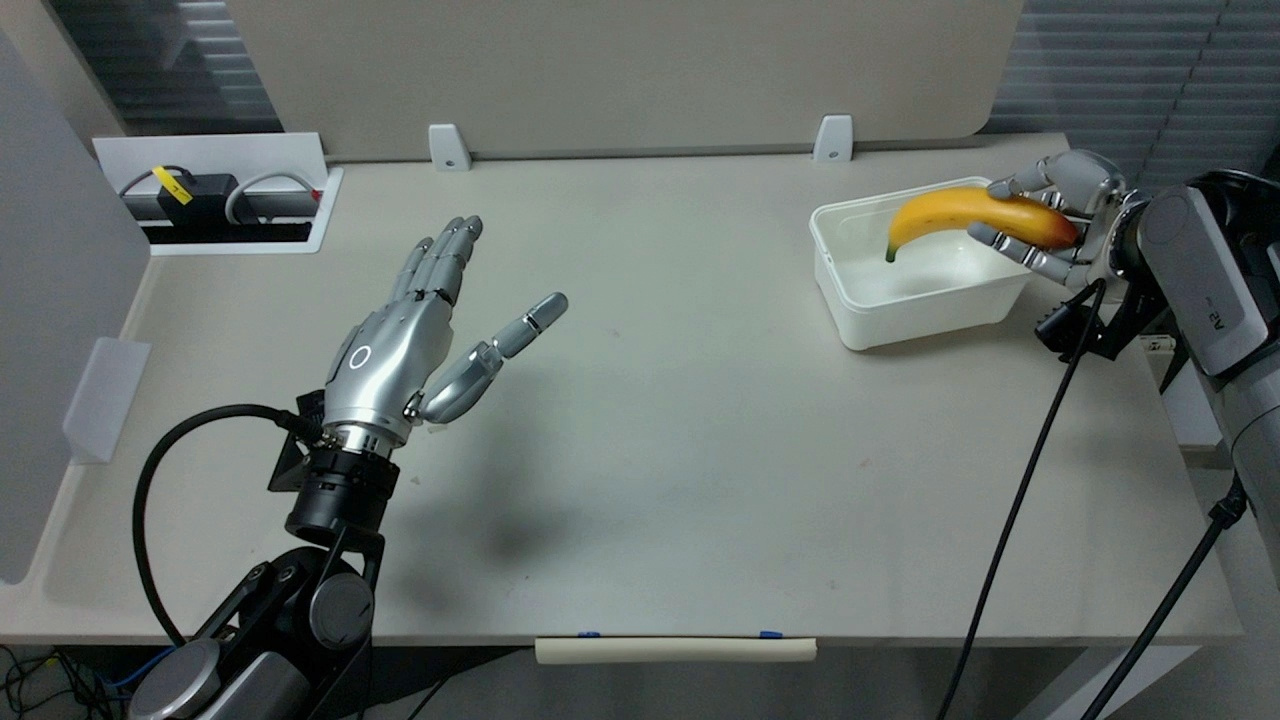}}%
    \RWSequenceFrame{5}{\includegraphics[width=\RWSeqPhotoWidth]{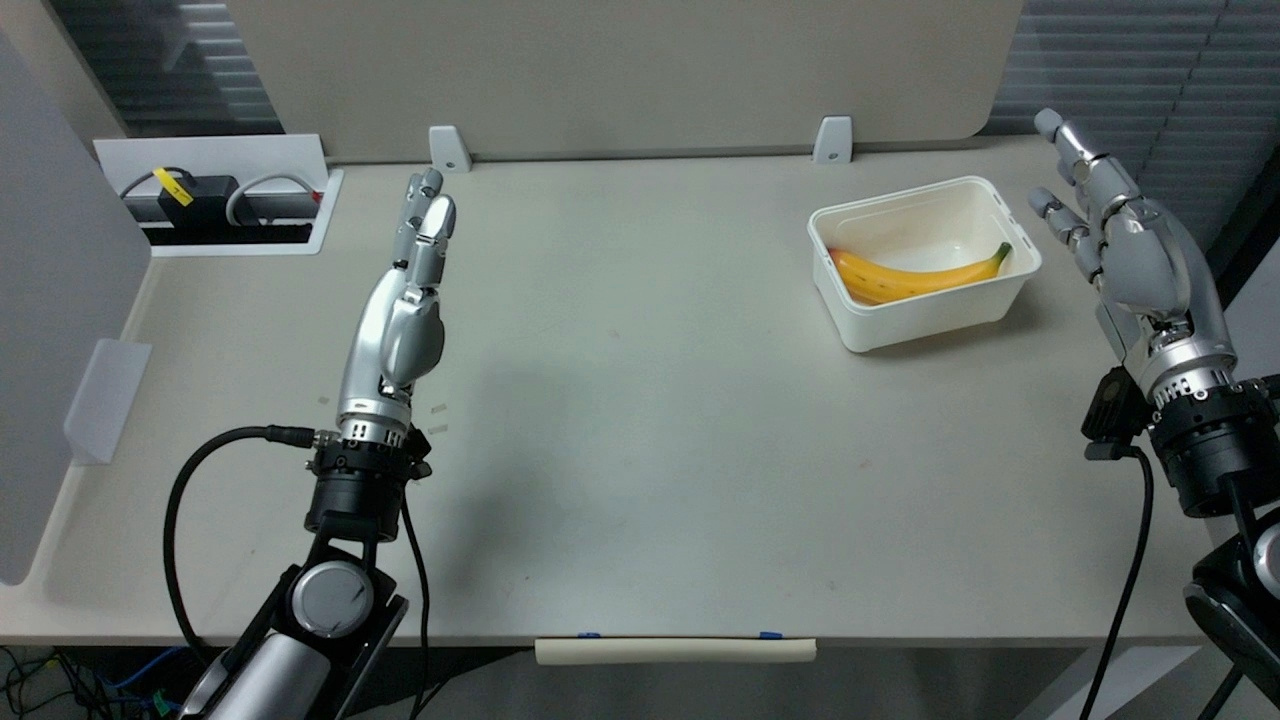}}%
}{(b) Handover}\par\vspace{6pt}
\RWSequence{%
    \RWSequenceFrame{0}{\includegraphics[width=\RWSeqPhotoWidth]{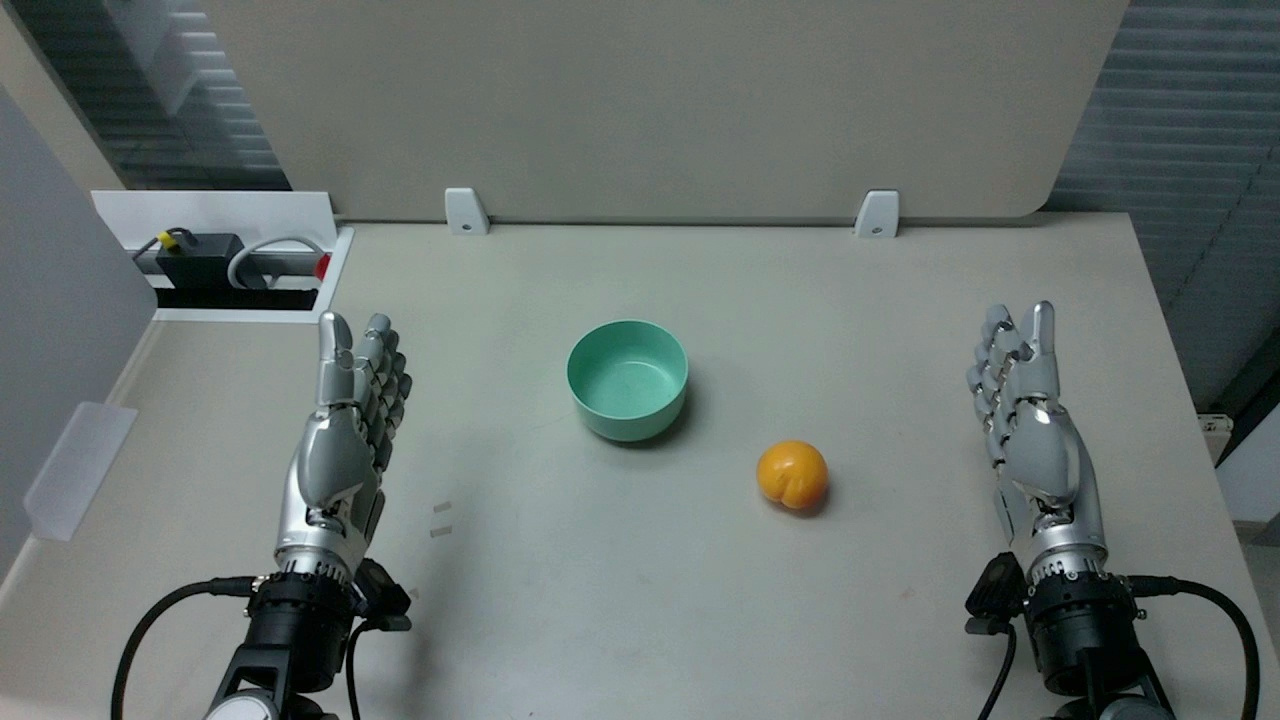}}%
    \RWSequenceFrame{1}{\includegraphics[width=\RWSeqPhotoWidth]{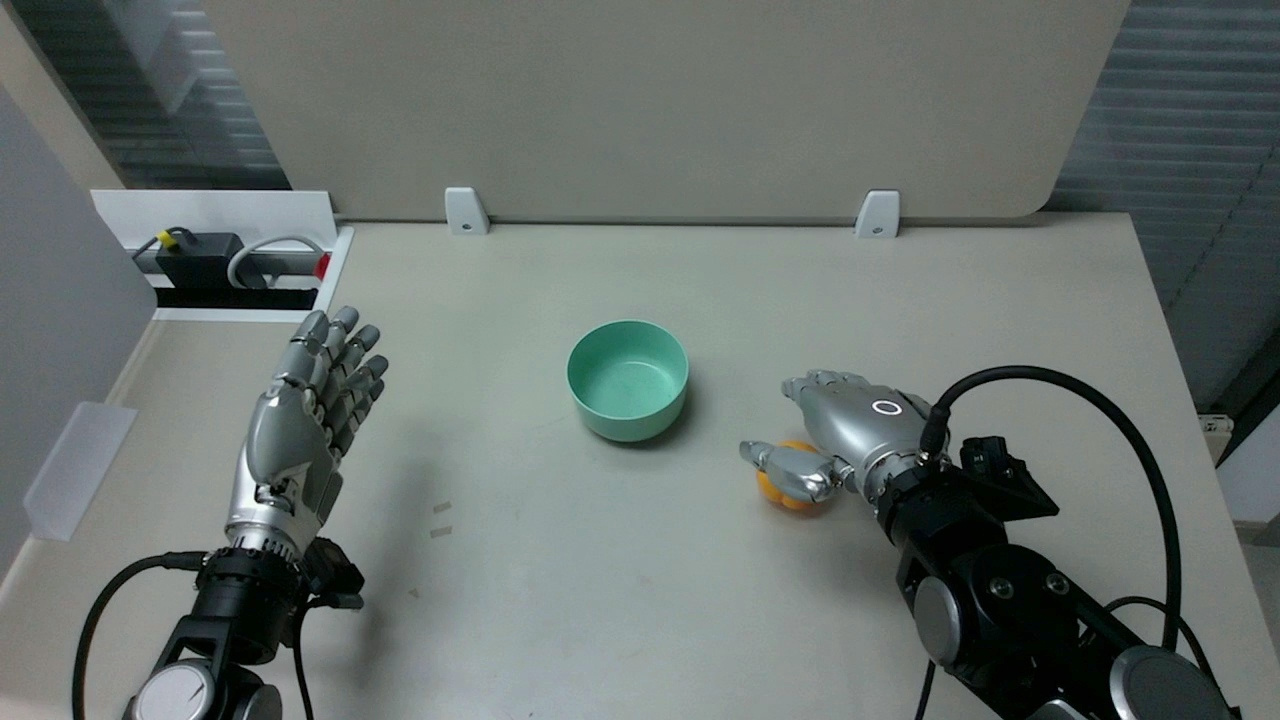}}%
    \RWSequenceFrame{2}{\includegraphics[width=\RWSeqPhotoWidth]{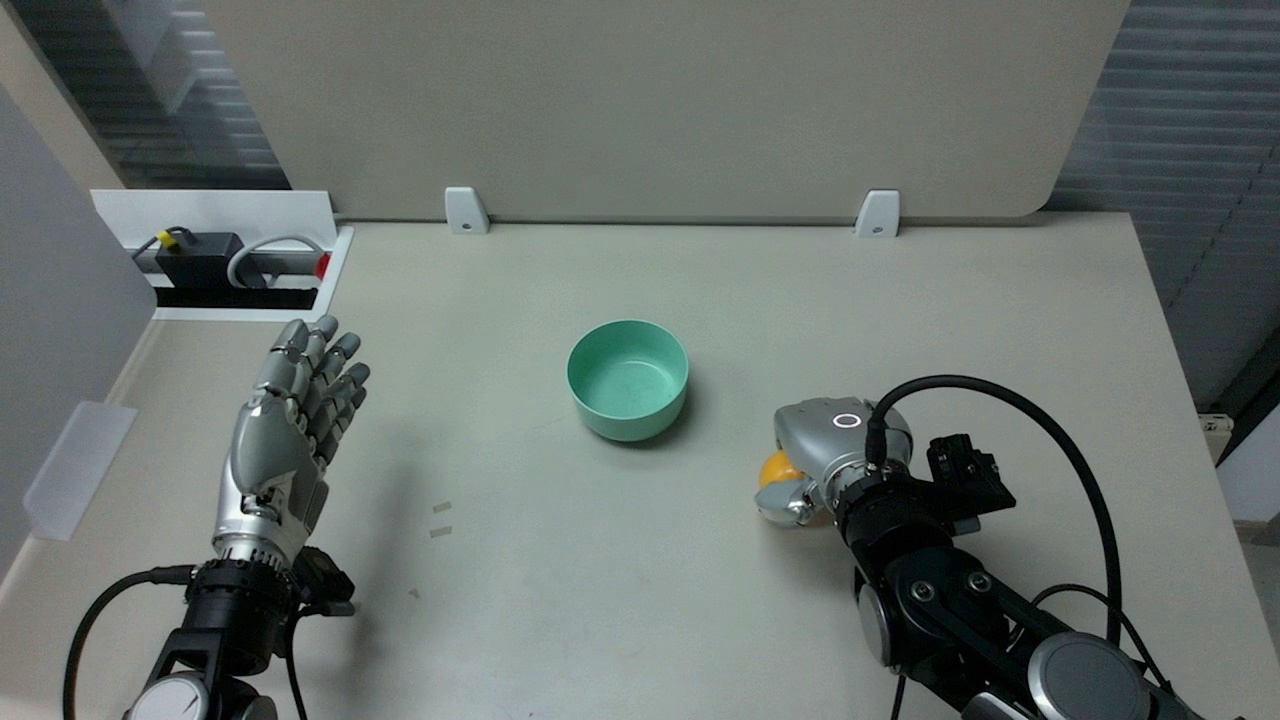}}%
    \RWSequenceFrame{3}{\includegraphics[width=\RWSeqPhotoWidth]{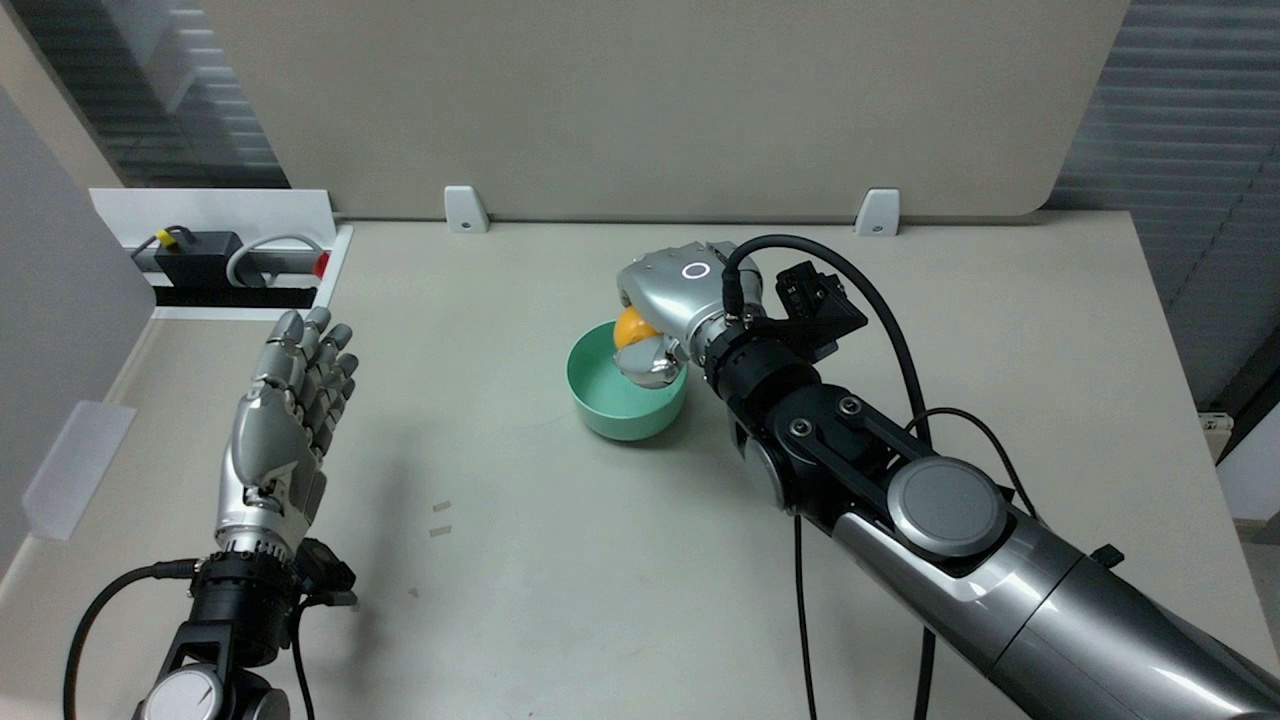}}%
    \RWSequenceFrame{4}{\includegraphics[width=\RWSeqPhotoWidth]{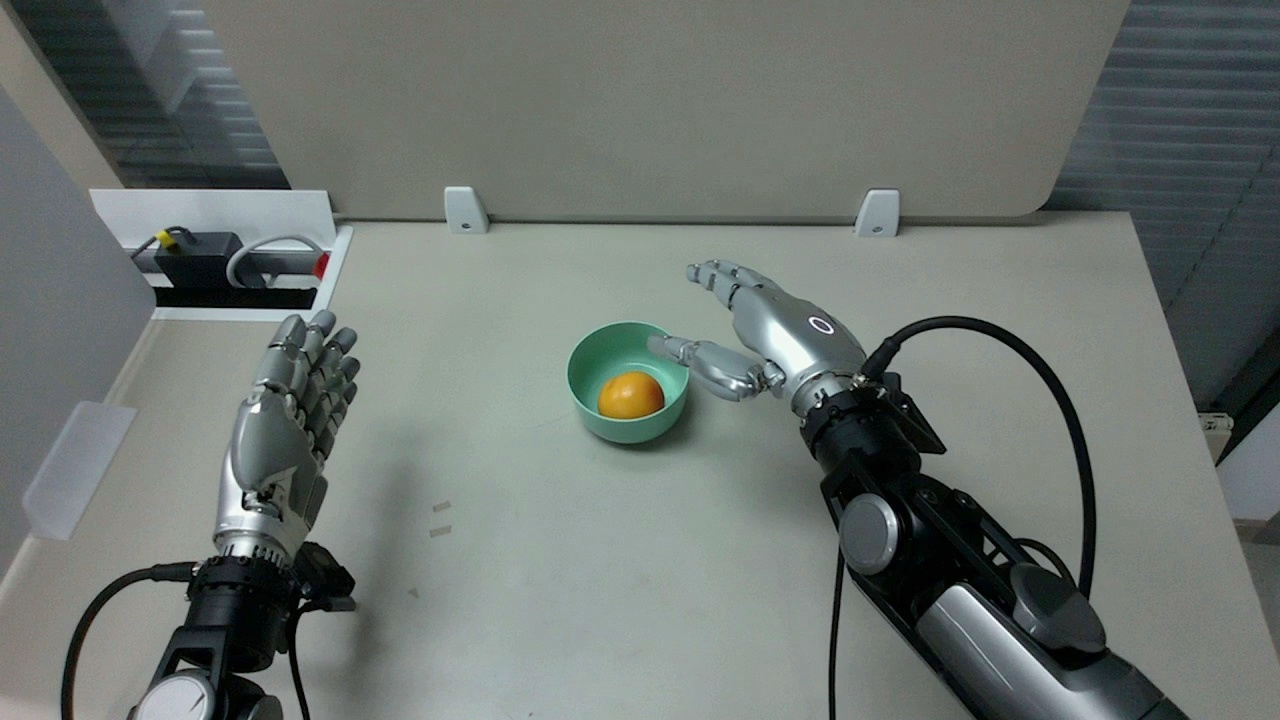}}%
    \RWSequenceFrame{5}{\includegraphics[width=\RWSeqPhotoWidth]{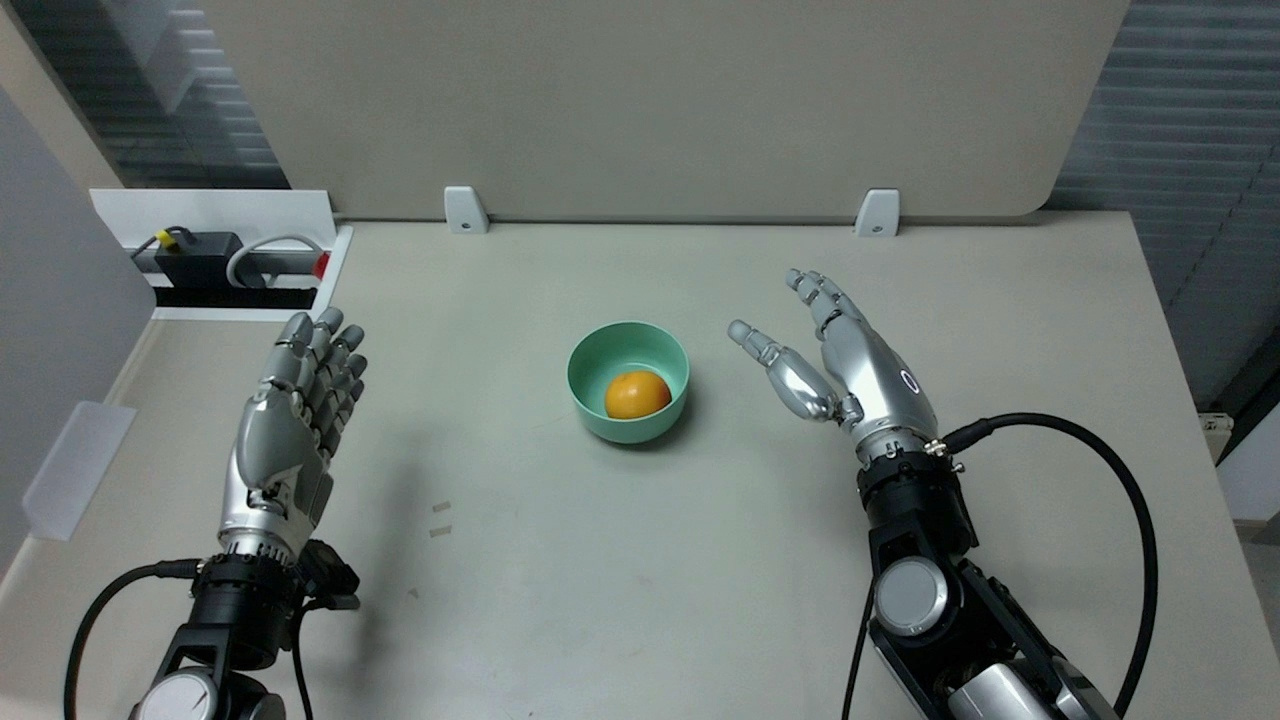}}%
}{(c) Pick and place}\par\vspace{6pt}
\RWSequence{%
    \RWSequenceFrame{0}{\includegraphics[width=\RWSeqPhotoWidth]{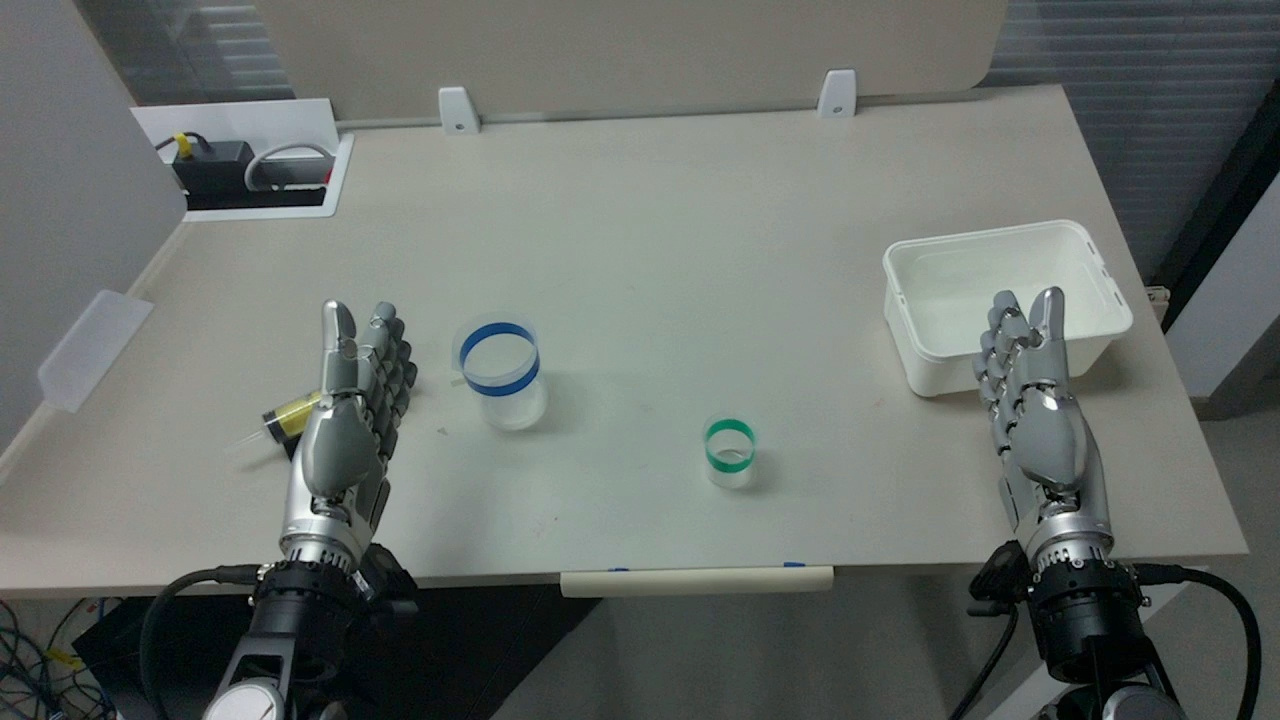}}%
    \RWSequenceFrame{1}{\includegraphics[width=\RWSeqPhotoWidth]{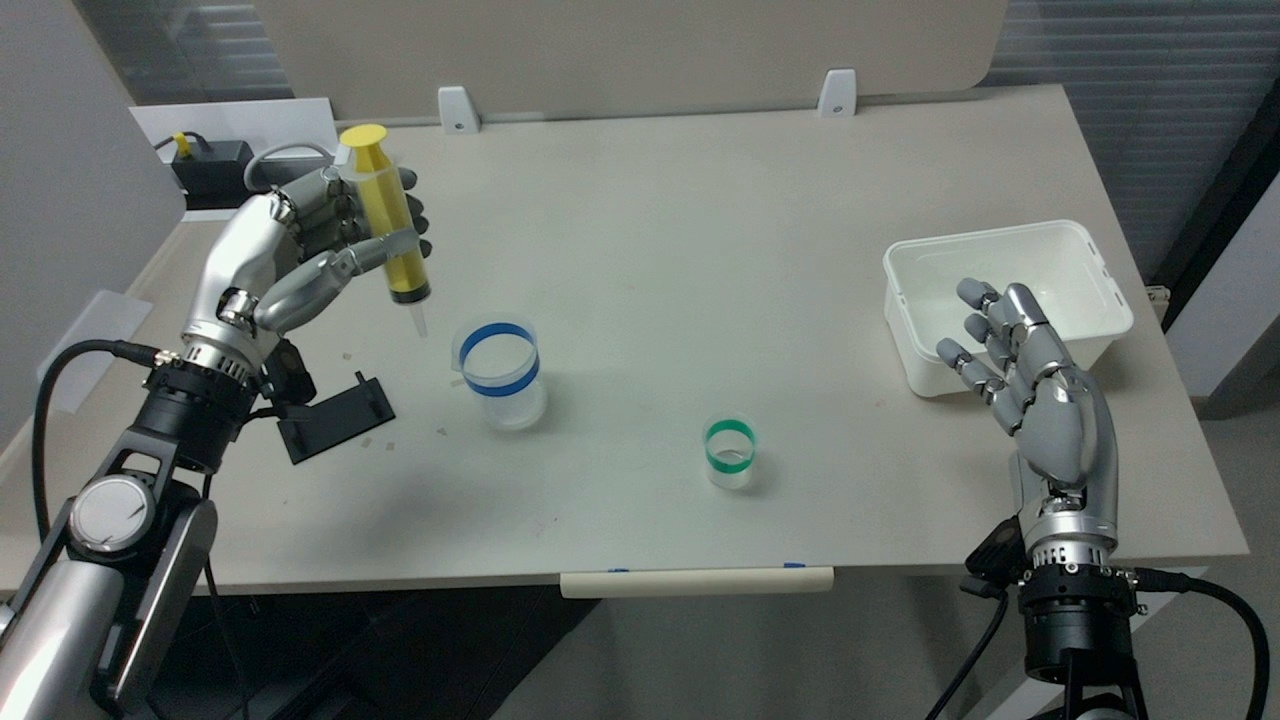}}%
    \RWSequenceFrame{2}{\includegraphics[width=\RWSeqPhotoWidth]{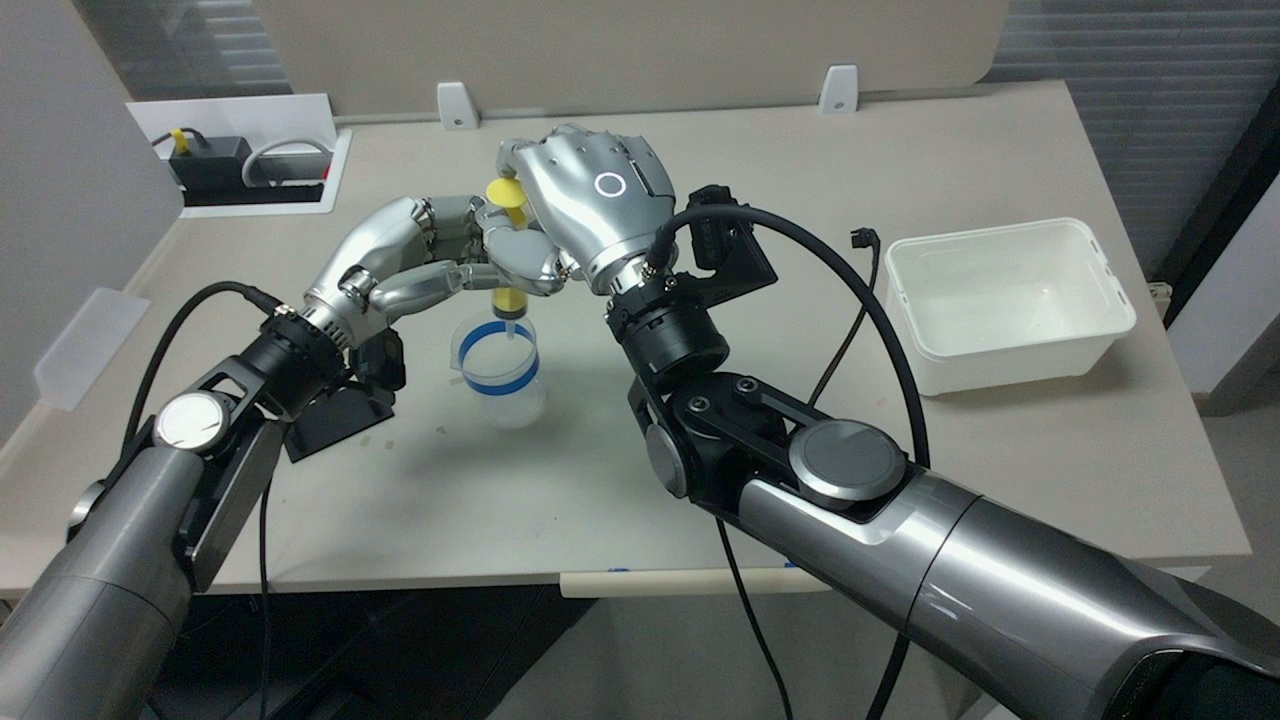}}%
    \RWSequenceFrame{3}{\includegraphics[width=\RWSeqPhotoWidth]{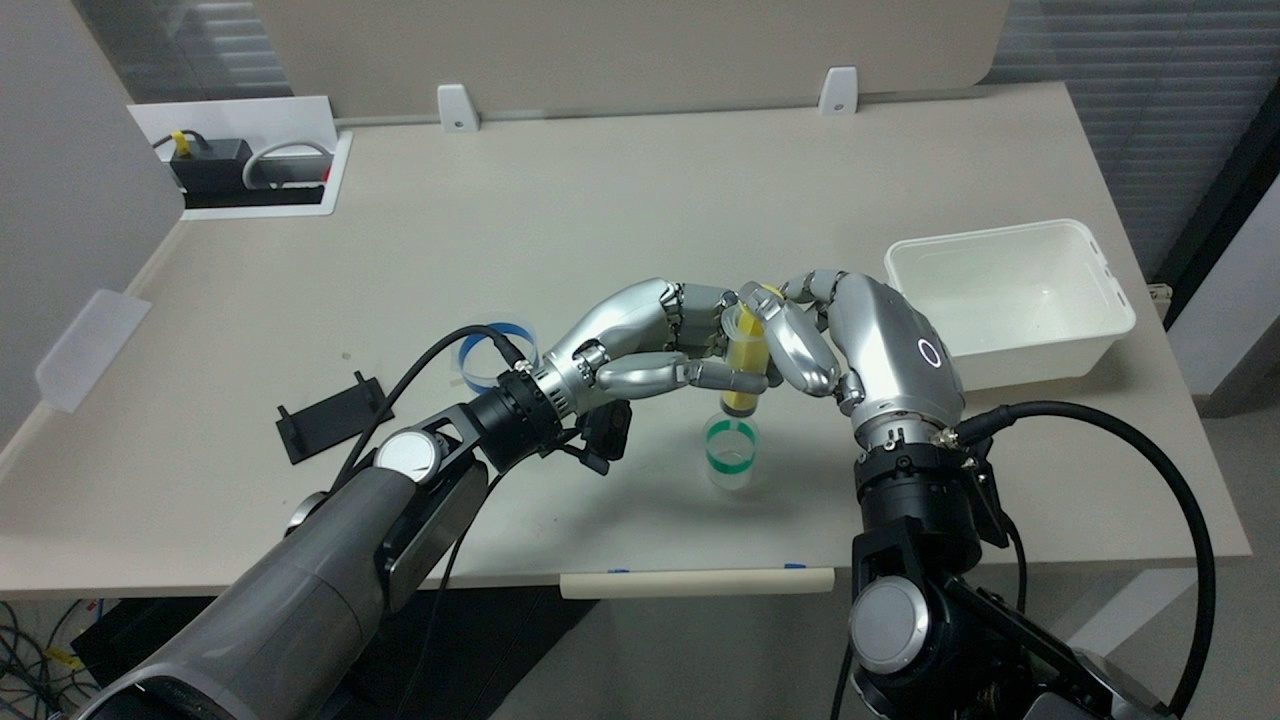}}%
    \RWSequenceFrame{4}{\includegraphics[width=\RWSeqPhotoWidth]{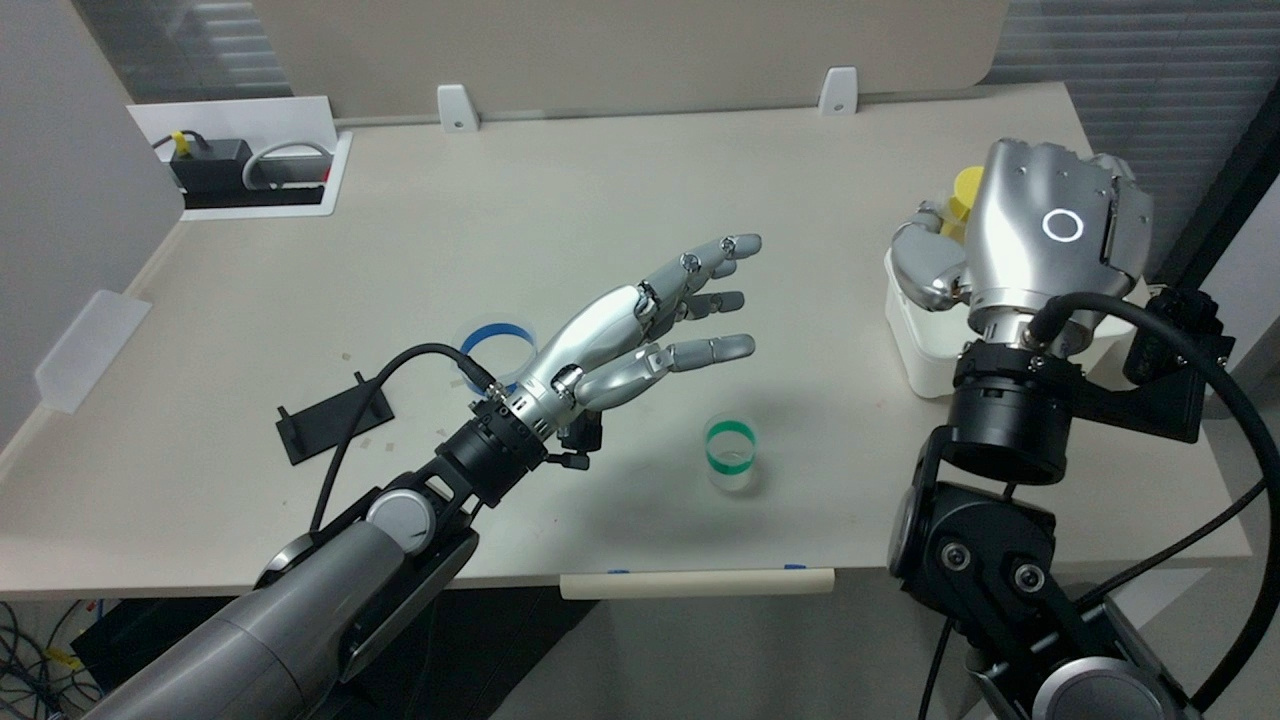}}%
    \RWSequenceFrame{5}{\includegraphics[width=\RWSeqPhotoWidth]{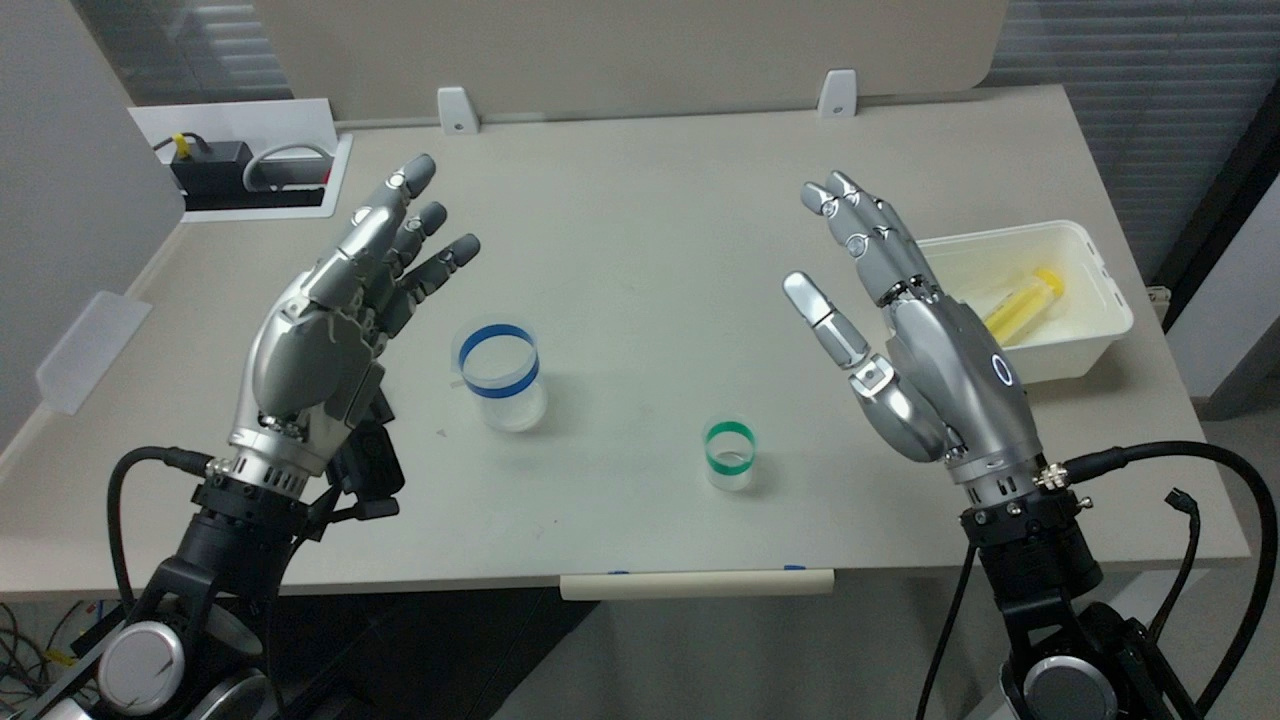}}%
}{(d) Syringe transfer}\par
\setlength{\abovecaptionskip}{6pt}
\caption{\textbf{Real-world task demonstrations.} Each row presents a task from its initial configuration to completion, ordered from left to right.}
\label{fig:real_world_workflows}
\endgroup
\end{figure}

\paragraph{Tasks, completion criteria, and progress scores.}
We use four tabletop tasks with language instructions from the demonstration metadata. A trial succeeds only when the complete task goal is achieved. Progress scores lie in $[0,1]$: bottle opening, handover, and pick and place receive the highest achieved milestone score (0 if none), whereas syringe transfer sums the credits for completed steps, each counted once:
\begin{enumerate}
    \item \textbf{Bottle opening:} unscrew the bottle lid, place the lid on the table, and return the bottle to the wooden block. Milestones: pick up the bottle (0.1), unscrew the cap with at least two rotation attempts (0.5), remove the cap (0.6), place the cap on the table (0.8), and return the bottle to the block (1.0).
    \item \textbf{Handover:} pick up the banana with the left hand, transfer it to the right hand, and release it into the white box. Milestones: grasp the banana (0.2), lift it from the table (0.4), transfer it to the right hand (0.7), and place it into the box (1.0).
    \item \textbf{Pick and place:} use the right hand to pick up the fruit and place it into the bowl. Milestones: lift the fruit from the table (0.5) and place it into the bowl (1.0).
    \item \textbf{Syringe transfer:} draw material from the left beaker with the syringe, transfer it into the right beaker, and place the syringe into the white box. Additive credits: pick up the syringe (+0.1), aim at the left beaker (+0.1), pull the plunger to draw liquid (+0.2), aim at the right beaker (+0.1), push the plunger to inject liquid (+0.2), transfer the syringe to the right hand (+0.2), and place it into the box (+0.1).
\end{enumerate}
Reported progress scores are averaged over trials and scaled to 0--100. Figure~\ref{fig:real_world_tasks} illustrates the four tasks, and Figure~\ref{fig:real_world_workflows} visualizes each task from its initial configuration to completion.

\paragraph{Demonstration budget and evaluation.}
We use 100 teleoperated demonstrations per task for post-training, totaling 400 demonstrations. We compare \methodname\ with $\pi_{0.5}$~\citep{pi2025pi05} and GR00T N1.7~\citep{nvidia2026gr00tn17} under this per-task demonstration budget. For each method, a single model is jointly post-trained on the combined demonstrations from all four tasks. Each method is evaluated in 20 trials per task. Task-level success is the percentage of trials that achieve the complete task goal (Figure~\ref{fig:real_world_tasks}).

\paragraph{Post-training settings.}
All methods are post-trained for 20k optimization steps with a batch size of 128. For \methodname, we use Muon~\citep{jordan2024muon} with learning rates of $5\times10^{-5}$ for the coupled vision--action MoT module and $1\times10^{-5}$ for the VLM, and 100 learning-rate warmup steps. We use the LeRobot implementations~\citep{cadene2024lerobot} of $\pi_{0.5}$ and GR00T N1.7, retaining the recommended default optimizer settings for each baseline.

\paragraph{Action-space adaptation for $\pi_{0.5}$.}
The released $\pi_{0.5}$ configuration uses a 32-dimensional action space. To accommodate our 58-DoF setting, we replace its action input and output layers with newly initialized 58-dimensional layers before post-training. This adaptation changes the pretrained action interface and should be considered when interpreting its performance.

\subsection{Additional ablations}
\label{sec:ablation_protocol}

\paragraph{Training setting.}
The reduced pretraining set contains approximately 1k hours of S1 data and 2k hours of additional S2 egocentric data, with S1 replay retained during S2. The default recipe uses 50k S1 updates and 100k S2 updates at batch size 256. Downstream adaptation follows the RoboCasa few-shot protocol in Appendix~\ref{sec:robocasa_protocol}. Pretraining teacher variants share the same adaptation recipe; post-training teacher variants start from the same full-scale pretrained checkpoint.

\paragraph{Pretraining-data scaling protocol.}
The scaling comparison uses nested subsets containing $0$, $1/10$, $1/3$, or all of the pretraining data, with matched source proportions for the nonzero subsets. At each nonzero scale, we apply the same fraction to every S1 source and the additional S2 egocentric data, train both stages in order, and retain the selected S1 data for replay throughout S2. Baseline and \methodname\ share the same subset and optimization budget at each scale; the baseline disables reliability conditioning and privileged-future distillation. The zero-data control skips both pretraining stages. Each downstream benchmark keeps its adaptation data and budget fixed across scales. Real-world open-loop evaluation computes mean squared error between predicted action chunks and held-out teleoperated actions, averaged over timesteps and valid normalized action coordinates; these demonstrations are excluded from adaptation. RoboCasa evaluation follows the few-shot closed-loop protocol in Appendix~\ref{sec:robocasa_protocol}. Video hours count unique retained pretraining data, not repeated sampling or S1 replay in S2. We fit $\log_{10}(\mathrm{MSE})$ and success percentage by ordinary least squares against $\log_{10}(h)$ using only nonzero durations, and omit the zero-hour control from the figure. Thus, $m$ is the dimensionless log--log slope for MSE or success-rate percentage points per decade, respectively. The nonzero durations are 3k, 10k, and 30k hours. For full-scale real-world MSE, the baseline value is the lowest observed along its checkpoint trajectory (0.039), rather than the later 0.042 endpoint shown in Figure~\ref{fig:teaser}(b); this avoids attributing a gain to the baseline's late-stage overfitting. Figure~\ref{fig:pretraining_scaling} uses a logarithmic 0.02--0.10 MSE axis and a linear 50--70\% success axis; the success axis marks the omitted lower range with a break.

\paragraph{Reliability sensitivity and loss weighting.}
Table~\ref{tab:ablation_reliability} varies human-data reliability while keeping robot-data reliability at one. The $q_d=1$ control removes source-dependent corruption and downweighting. All three tested values below one ($q_d=0.8,0.5,0.2$) improve average success from 52.0\% to 54.0--54.3\%, a spread of only 0.3 percentage points. The gains come primarily from Art., which improves from 47.7\% to 54.9--58.0\%, while PnP remains near 53\%. This limited sensitivity supports the use of coarse source-level reliability without more complex task- or sample-specific estimates. At $q_d=0.2$, retaining reliability-conditioned corruption but using unit action-loss weights gives nearly identical results (54.2\% versus 54.3\% average success). Thus, the benefit persists without loss downweighting; its additional contribution is small in this setting.

\begin{table}[!htbp]
\setlength{\belowcaptionskip}{7pt}
\caption{\textbf{Ablation of reliability and loss weighting.} RoboCasa few-shot success rates (\%) after reduced-scale pretraining. The last row changes only the loss weight. Dark and light green shading mark the best and second-best scores per column, including ties.}
\label{tab:ablation_reliability}
\centering
\small
\setlength{\tabcolsep}{7pt}
\renewcommand{\arraystretch}{1.20}
\ablationValues[ffSagel]{reliability}{ffSageDark}
  {53.4, 53.7, 53.0, 53.1, 52.9}
  {47.7, 55.0, 57.7, 58.0, 58.0}
  {52.0, 54.0, 54.2, 54.3, 54.2}
\begin{tabular*}{0.80\linewidth}{@{\extracolsep{\fill}}cc|rr|r@{}}
\toprule
\textbf{Human $\boldsymbol{q}_d$} & \textbf{Loss weight} & \textbf{PnP} & \textbf{Art.} & \textbf{Avg.} \\
\midrule
1.0 & 1.0 & \ablationResults{reliability}{1} \\
0.8 & 0.8 & \ablationResults{reliability}{2} \\
0.5 & 0.5 & \ablationResults{reliability}{3} \\
0.2 & 0.2 & \ablationResults{reliability}{4} \\
\midrule
0.2 & 1.0 & \ablationResults{reliability}{5} \\
\bottomrule
\end{tabular*}
\end{table}

\section{Qualitative future prediction and IDM outputs}
\label{sec:qualitative_outputs}

\begin{figure}[!ht]
    \centering
    \includegraphics[width=\linewidth]{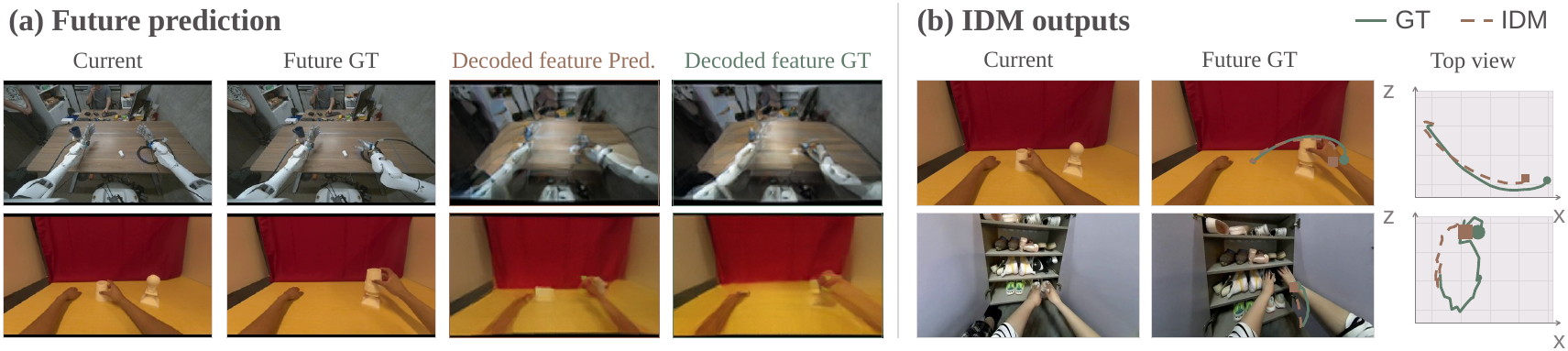}
    \caption{\textbf{Future prediction and IDM outputs.}
    \textbf{(a)} Current/future RGB and RAE reconstructions of pooled GT versus predicted DINO features, with identical decoding and cropping.
    \textbf{(b)} Relatively smooth IDM outputs (dashed) and GT (solid).}
    \label{fig:qualitative_forecasts_idm}
\end{figure}

Figure~\ref{fig:qualitative_forecasts_idm} shows current-only future prediction and IDM outputs conditioned on current and GT future observations.

Pooled GT and predicted DINOv3-B/16 features~\citep{simeoni2025dinov3} share a $12\!\times\!10\!\times\!768$ grid and the same frozen RAEv2 decoder~\citep{singh2026raev2}. Only decoder positional embeddings are resized for the rectangular grid. These lossy reconstructions and selected pretraining clips are qualitative illustrations, not pixel-accurate futures or held-out evaluations.

\section{Limitations}

The analysis in Appendix~\ref{sec:population_interpretation} assumes conditionally independent, zero-mean Gaussian annotation noise. Systematic retargeting bias can shift conditional action means, leaving flow-field discrepancies even at maximal corruption; source weighting alone does not correct these shifts. Dataset-level reliability misses within-source variation. Future work could explore finer-grained estimates at the example, time-step, or action-coordinate level. Dependence on action annotations motivates self-supervised learning from unannotated videos. The tested scales leave compositional generalization to unseen object--skill--scene combinations unverified, motivating larger-scale studies of supervision quality and diversity. Predicted futures and EMA teachers may supply biased or action-irrelevant supervision, while adding computational cost.

\section{Additional Related Work}
\label{sec:additional_related}

\subsection{Embodied Data Pyramid}

Robot learning draws on a data pyramid shaped by coverage, collection cost, and action availability. Human videos offer broad interaction coverage~\citep{grauman2022ego4d,hoque2025egodex}, whereas robot demonstrations tie observations to executable controls~\citep{walke2023bridge,oxe2023,khazatsky2024droid,agibot2025}. Simulation expands trajectory coverage through data generation~\citep{mandlekar2023mimicgen}, and handheld interfaces make transferable demonstrations easier to collect~\citep{chi2024umi}. Exploiting this diversity requires connecting visual experience to action supervision. Early use of human video focused on learning visual representations for control~\citep{nair2022r3m,karamcheti2023voltron}. Later approaches make the connection more explicit by inferring latent actions from visual transitions~\citep{ye2024lapa,bu2025univla,zhang2026clap}. EgoMimic and EgoScale convert estimated or tracked human hand motion into robot supervision~\citep{kareer2024egomimic,zheng2026egoscale}, whereas X-Diffusion transfers action distributions across embodiments~\citep{pace2025xdiffusion}. These routes broaden the available training signal, but produce mixtures in which robot actions may be absent or noisy. We study joint future--action learning under this uneven supervision.

\subsection{Representation Learning for Generation}

Generative objectives have long provided a way to learn representations: denoising autoencoders extract useful features by reconstructing corrupted inputs~\citep{vincent2010dae}, while VAEs and latent diffusion learn the spaces in which generation takes place~\citep{kingma2014vae,rombach2022ldm}. REPA~\citep{yu2025repa} exploits the reverse connection, using pretrained clean-image representations to guide intermediate denoiser features. This makes the source and form of representation guidance central design choices. Existing VAE features can supply alignment targets~\citep{wang2026sra2}. REPA-E jointly tunes the VAE and latent diffusion transformer end to end~\citep{leng2025repae}, whereas RAE rebuilds the generative latent space around a pretrained representation encoder~\citep{zheng2025rae}. Guidance can also come from within the generator itself. SRA and LayerSync use representations from the model's own layers as alignment targets, reducing dependence on external encoders~\citep{jiang2026sra,haghighi2025layersync}. Self-Flow~\citep{chefer2026selfflow} develops this internal approach with an EMA teacher evaluated at a cleaner, noise-shifted input and aligns deeper teacher features to earlier student features; it also evaluates the mechanism on joint video--action prediction.

\subsection{Learning from Noisy Data}

Noisy supervision can bias optimization, motivating loss corrections based on assumptions about label corruption~\citep{natarajan2013noisy,patrini2017correction}. When trust is estimated from the data, meta-reweighting~\citep{ren2018reweight} and sample selection with peer learning or semi-supervision~\citep{han2018coteaching,li2020dividemix} adjust how examples contribute to training. For restoration and generation, the problem extends from unreliable labels to corruption of the observations themselves. Noise2Noise and Noise2Self establish that restoration remains possible without clean targets under suitable noise assumptions~\citep{lehtinen2018noise2noise,batson2019noise2self}, while AmbientGAN and Ambient Diffusion learn generative distributions from corrupted measurements~\citep{bora2018ambientgan,daras2023ambient}. The usefulness of lower-quality data can also depend on the denoising stage. Ambient Omni~\citep{daras2025ambientomni} studies noise- and timestep-dependent use or weighting of data with different quality, and Ambient Diffusion Policy~\citep{wei2026ambientpolicy} carries this idea into policy learning by exploiting coarse structure or local motion at different stages. We jointly adapt action corruption and loss weighting to source reliability.

\subsection*{Author Contributions}

\begingroup
\setlength{\parskip}{3pt}
\noindent\textbf{Data:} Chubin Zhang, Tianle Zhang.\par
\noindent\textbf{Pretraining:} Chubin Zhang.\par
\noindent\textbf{Post-training:} Chubin Zhang, Changyuan Wang, Yirui Wang, Ao Li.\par
\noindent\textbf{Deployment and demo:} Yirui Wang.\par
\noindent\textbf{Method design:} Chubin Zhang, Cai Zhou.\par
\noindent\textbf{Writing:} Chubin Zhang, Changyuan Wang, Shilin Ma, Yue Su.\par
\noindent\textbf{Guidance:} Xiuwei Xu, Tianle Zhang, Yansong Tang.\par
\noindent\textbf{Industry lead:} Tianle Zhang.\par
\noindent\textbf{Corresponding:} Yansong Tang.\par
\endgroup

\subsection*{AI use statement}

Generative AI tools were used to polish the manuscript's writing, create the paper teaser and illustrative icons in the pipeline diagram, assist with coding, and support literature search and review. AI-generated visual content is limited to these illustrative elements and does not include dataset example images. The authors are responsible for the accuracy and integrity of the manuscript, code, references, and reported results.

\subsection*{Ethics statement}

This work studies policy learning from egocentric human video and robot demonstrations. Such data can raise privacy and data-governance concerns; the internal egocentric data used for pretraining cannot be publicly released due to applicable data-access and disclosure policies. This restriction should not be interpreted as granting permission to redistribute or repurpose these data. Learned robot policies can fail in ways that damage objects or cause injury, and the reported results do not establish safety beyond the evaluated tasks and operating conditions. Any deployment requires task-specific safety assessment, appropriate safeguards, and human oversight.

\pagebreak[1]
\subsection*{Reproducibility statement}

The main paper and appendix describe the method, training objectives, and two-stage pretraining procedure, together with the post-training settings and evaluation protocols. Full reproduction of pretraining is limited by the internal egocentric data, which cannot be publicly released due to policy restrictions. In contrast, the post-training procedure is fully specified to support reproduction from the pretrained checkpoints. We will open-source the method implementation and release model checkpoints, enabling researchers to reproduce post-training and build on the method without access to the restricted internal egocentric data.

\end{document}